\documentclass[letterpaper, 11pt, reqno]{amsart}
\usepackage[utf8]{inputenc}
\usepackage[T1]{fontenc}
\usepackage[top = 1in, bottom = 1in, left = 1in, right = 1in]{geometry}
\usepackage{etoolbox}
	\makeatletter
		\patchcmd{\@adjustvertspacing}{plus.7\baselineskip}{minus.14\baselineskip}{}{}
	\makeatother
	\AtBeginDocument{
		\raggedbottom
		\predisplaypenalty = 100
		\allowdisplaybreaks[1]
		\setlength{\emergencystretch}{1em}
	}
\usepackage[bottom, hang, flushmargin]{footmisc}
	\newlength\tindent
	\newcommand\affiliation[2][*]{\begingroup
		\renewcommand\thefootnote{}\footnote{#1 #2}%
		\addtocounter{footnote}{-1}%
		\endgroup}
	\makeatletter
	\patchcmd{\@maketitle}
		{\dimen@34\p@}
		{\dimen@20\p@}
		{}{}
	\makeatother
\usepackage[skip = 6pt plus 1pt minus 1pt, indent = 0pt]{parskip}

\usepackage{amsmath}
\usepackage{amsthm}
	\usepackage{aliascnt}
\usepackage{amsfonts}
\usepackage{bm}
\usepackage{mathtools}
\usepackage{dsfont}

\usepackage{algorithmic}
\usepackage[ruled, vlined, linesnumbered]{algorithm2e}
	\DontPrintSemicolon{}
	\SetNlSty{texttt}{}{.}
	\SetAlCapHSkip{0ex}
	\SetAlgoNlRelativeSize{-1}
	\makeatletter
		\renewcommand*\nllabel[1]{%
			{\let\@currentlabel\algocf@currentlabel
			\let\@currentcounter\algocf@currentcounter
			\label{#1}}%
		}%
		\renewcommand*\algocf@nl@sethref[1]{%
		\renewcommand\theHAlgoLine{\thealgocfproc.#1}%
			\hyper@refstepcounter{AlgoLine}%
			\gdef\algocf@currentlabel{#1}%
			\gdef\algocf@currentcounter{AlgoLine}%
		}%
	\makeatother

\usepackage{booktabs}
\usepackage{placeins}
\usepackage{graphicx}
\usepackage{subcaption}
\usepackage{caption}
\usepackage{float} 
\usepackage{tikz}
\usepackage{pgfplots}
	\pgfplotsset{compat = newest}
	\usepgfplotslibrary{external}
	\usetikzlibrary{intersections}
\usepackage{xcolor}
	\definecolor{forestgreen}{RGB}{34, 139, 34}
	\definecolor{tomato}{RGB}{255, 99, 71}
	\definecolor{royalblue}{RGB}{65, 105, 225}

\usepackage{lipsum}
\usepackage{enumitem}
	
\usepackage{silence}
\usepackage{comment}
\usepackage{xparse}
\usepackage[normalem]{ulem}

\usepackage[colorlinks = true, linkcolor = gray, citecolor = gray, urlcolor = gray]{hyperref}
\usepackage[backref = true, style = alphabetic, sortcites = true, maxcitenames = 2, maxbibnames = 99, backend = biber, url = false]{biblatex}
	\AtBeginRefsection{\GenRefcontextData{sorting = ynt}}
	\AtEveryCite{\localrefcontext[sorting = ynt]}
	\makeatletter
		\def\ps@bibliography{%
			\ps@headings
			\def\leftmark{\MakeUppercase{\refname}}%
			\let\rightmark\leftmark
		}
	\makeatother
	\AtBeginBibliography{\thispagestyle{bibliography}}
	\renewbibmacro*{title}{%
		\iffieldundef{title}{}
		{\iffieldundef{url}
			{\printtext[title]{\printfield[titlecase]{title}}}%
			{\printtext[title]{\href{\thefield{url}}{\printfield[titlecase]{title}}}}
		}}
	\newcommand{\citet}[2][]{\citeauthor{#2} \cite[#1]{#2}}
\usepackage{doi}
\usepackage{cleveref}
	\crefname{theorem}{theorem}{theorems}
	\Crefname{theorem}{Theorem}{Theorems}
	\crefname{lemma}{lemma}{lemmas}
	\Crefname{lemma}{Lemma}{Lemmas}
	\crefname{corollary}{corollary}{corollaries}
	\Crefname{corollary}{Corollary}{Corollaries}
	\crefname{proposition}{proposition}{propositions}
	\Crefname{proposition}{Proposition}{Propositions}
	\crefname{definition}{definition}{definitions}
	\Crefname{definition}{Definition}{Definitions}
	\crefname{remark}{remark}{remarks}
	\Crefname{remark}{Remark}{Remarks}
	\crefname{example}{example}{examples}
	\Crefname{example}{Example}{Examples}
	\crefname{problem}{problem}{problems}
	\Crefname{problem}{Problem}{Problems}
	\crefname{table}{tab.}{tabs.}
	\Crefname{table}{Tab.}{Tabs.}
	\crefname{section}{section}{sections}
	\Crefname{section}{Section}{Sections}
	\crefname{figure}{fig.}{figs.}
	\Crefname{figure}{Fig.}{Figs.}
	\crefname{equation}{eq.}{eqs.}
	\Crefname{equation}{Eq.}{Eqs.}
	\crefname{line}{line}{lines}
	\Crefname{line}{Line}{Lines}
	\crefname{proof}{proof}{proofs}
	\Crefname{proof}{Proof}{Proofs}
	\crefname{assumption}{assumption}{assumptions}
	\Crefname{assumption}{Assumption}{Assumptions}
	\crefname{question}{question}{questions}
	\Crefname{question}{Question}{Questions}
	\crefname{appendix}{appendix}{appendices}
	\Crefname{appendix}{Appendix}{Appendices}
	\AddToHook{cmd/appendix/before}{%
		\crefalias{section}{appendix}%
		\crefalias{subsection}{appendix}%
		\crefalias{subsubsection}{appendix}%
	}

\theoremstyle{plain}
	\newtheorem{theorem}{Theorem}[section]
	\newaliascnt{lemma}{theorem}
	\newtheorem{lemma}[lemma]{Lemma}
	\aliascntresetthe{lemma}
	\newaliascnt{corollary}{theorem}
	\newtheorem{corollary}[corollary]{Corollary}
	\aliascntresetthe{corollary}
	\newaliascnt{proposition}{theorem}
	\newtheorem{proposition}[proposition]{Proposition}
	\aliascntresetthe{proposition}
	\newaliascnt{properties}{theorem}
	
	\aliascntresetthe{properties}
\theoremstyle{remark}
	\newaliascnt{remark}{theorem}
	\newtheorem{remark}[remark]{Remark}
	\aliascntresetthe{remark}
	\newaliascnt{todo}{theorem}
	
	\aliascntresetthe{todo}
	\newaliascnt{ideas}{theorem}
	
	\aliascntresetthe{ideas}
\theoremstyle{definition}
	\newaliascnt{definition}{theorem}
	\newtheorem{definition}[definition]{Definition}
	\aliascntresetthe{definition}
	\newaliascnt{assumption}{theorem}
	\newtheorem{assumption}[assumption]{Assumption}
	\aliascntresetthe{assumption}
	\newaliascnt{example}{theorem}
	
	\aliascntresetthe{example}
	\newaliascnt{problem}{theorem}
	
	\aliascntresetthe{problem}

\AddToHook{cmd/tableofcontents/before}{%
	\renewcommand{\leftmark}{\MakeUppercase{\contentsname}}
	\renewcommand{\rightmark}{\MakeUppercase{\contentsname}}
}
\makeatletter
	\def\@tocline#1#2#3#4#5#6#7{\relax
		\ifnum #1 > \c@tocdepth 
		\else
			\par \addpenalty\@secpenalty\addvspace{#2}%
			\begingroup \hyphenpenalty\@M
			\@ifempty{#4}{%
				\@tempdima\csname r@tocindent\number#1\endcsname\relax
			}{%
				\@tempdima#4\relax
			}%
			\parindent\z@ \leftskip#3\relax \advance\leftskip\@tempdima\relax
			\rightskip\@pnumwidth plus4em \parfillskip-\@pnumwidth
			#5\leavevmode\hskip-\@tempdima
				\ifcase #1
				\or\or \hskip 1em \or \hskip 2em \else \hskip 3em \fi%
				\ifnum #1 = 1 {\bfseries #6}\else #6\fi\nobreak\relax
			\dotfill\hbox to\@pnumwidth{\ifnum #1 = 1 \hss{\bfseries #7}\else \@tocpagenum{#7}\fi}\par
			\nobreak
			\endgroup
		\fi}
\makeatother

	\let\epsilon\varepsilon
	
	\let\oldforall\forall
	\renewcommand{\forall}{\oldforall\,}
	\let\oldexists\exists
	\renewcommand{\exists}{\oldexists\,}
	\DeclareMathSymbol{\gtrsim}{\mathrel}{AMSa}{"26}
	\DeclareMathSymbol{\lesssim}{\mathrel}{AMSa}{"2E}
	\DeclareMathSymbol{\lessgtr}{\mathrel}{AMSa}{"37}
	\makeatletter
		\DeclareRobustCommand\boldlessgtr{\mathrel{\mathpalette\boldlessgtr@{}}}
		\newcommand\boldlessgtr@[2]{\ooalign{\hidewidth\raise0.2pt%
			\hbox{$\m@th#1\lessgtr$}\hidewidth\cr$\m@th#1\lessgtr$\cr}}
	\makeatother

	\newcommand{\blankeq}{\kern14.65pt}

\newcommand{\real}{\mathbb{R}}
\newcommand{\integer}{\mathbb{Z}}
\renewcommand{\natural}{\mathbb{N}}

\newcommand{\sign}[1]{\operatorname{sign}\left(#1\right)}
\newcommand{\bigo}[1]{\mathcal{O}\left(#1\right)}
\newcommand{\smallo}[1]{o\left(#1\right)}
\newcommand{\norm}[1]{\left\lVert#1\right\rVert}
\newcommand{\dist}[2]{\operatorname{dist}\left(#1, #2\right)}
\newcommand{\innerproduct}[2]{\left\langle #1, #2 \right\rangle}
\newcommand{\set}[1]{\left\{#1\right\}}

\newcommand{\abs}[1]{\left|#1\right|}
\newcommand{\eval}[1]{\left.#1\right|}

\newcommand{\ceil}[1]{\left\lceil#1\right\rceil}

\renewcommand{\ker}[1]{\operatorname{ker}\left(#1\right)}
\newcommand{\rank}[1]{\operatorname{rank}\left(#1\right)}
\renewcommand{\dim}[1]{\operatorname{dim}\left(#1\right)}
\newcommand{\codim}[1]{\operatorname{codim}\left(#1\right)}

\renewcommand{\d}{\mathrm{d}}
\newcommand{\midbar}{\;\middle|\;}

\newcommand{\train}{\mathcal{D}}

\newif\ifshowcomment
\setbool{showcomment}{true}
\NewDocumentCommand{\caporali}{+o +m}{%
	\ifshowcomment
		\IfValueT{#1}{\sout{#1} }{\color{tomato}#2}%
	\else
		\IfValueT{#1}{#1}%
	\fi
}

\NewDocumentCommand{\hanin}{+o +m}{%
	\ifshowcomment
		\IfValueT{#1}{\sout{#1} }{\color{blue}#2}%
	\else
		\IfValueT{#1}{#1}%
	\fi
}

\providecommand{\note}[1]{}
\renewcommand{\note}[1]{{\textcolor{red}{[#1]}}}

\title[Critically damped Polyak stepsizes]{A new algorithmic approach for finding flat minima: critically damped Polyak stepsizes}
\author[F. Caporali]{Francesco Caporali}
\author[J. M. Altschuler]{Jason M. Altschuler}
\author[B. Hanin]{Boris Hanin}
\newcommand{\makeaffiliations}{
	\affiliation[]{
		Department of Operations Research and Financial Engineering, School of Engineering and Applied Science, Princeton University.
		\texttt{\{fcaporali,jaltschuler,bhanin\}@princeton.edu}.}
}

\bibliography{files/bibliography}

\begin{document}

\begin{abstract}
	We present a new adaptive stepsize rule that biases gradient descent toward flatter minima in both theory and practice, and establish the first uniform-in-time guarantees for both sharpness and loss reduction for a broad class of locally quadratic objectives.
	Our \textit{critically damped Polyak (CDP) stepsizes} multiply the classical Polyak rule by $4$ and a mild damping factor.
	The factor $4$ is critical in that it produces equioscillations around the manifold of minimizers $\mathcal{M}$ to leading order; damping provably drives convergence to $\mathcal{M}$. CDP thus operates at the edge of stability by adapting its stepsizes to local curvature. 
    Once the loss is small, its projected iterates follow gradient descent on $\log{\lambda_1(\nabla^2 \mathcal{L})}$ along $\mathcal{M}$ to leading order, paralleling the flatness bias of normalized gradient descent \cite{eos_arora_2022}.
    Unlike normalized gradient descent, CDP step lengths shrink with the distance to $\mathcal M$, enabling convergence to an interpolating solution that locally minimizes sharpness at finite learning rate.
    We establish our theory for a broad class of non-negative losses with locally quadratic Morse--Bott structure near $\mathcal{M}$, capturing regular interpolation regimes in smooth overparameterized neural networks.
    Empirically, CDP stepsizes produce a flattening effect similar to that of Sharpness-Aware Minimization (SAM) \cite{sam_foret_2021}, reducing sharpness even in settings beyond our theory, including nonlinear teacher-student networks and fine-tuning of pre-trained LLMs. 
\end{abstract}

\maketitle
\makeaffiliations
\enlargethispage{2pt}
\section{Introduction} \label{sec:introduction}

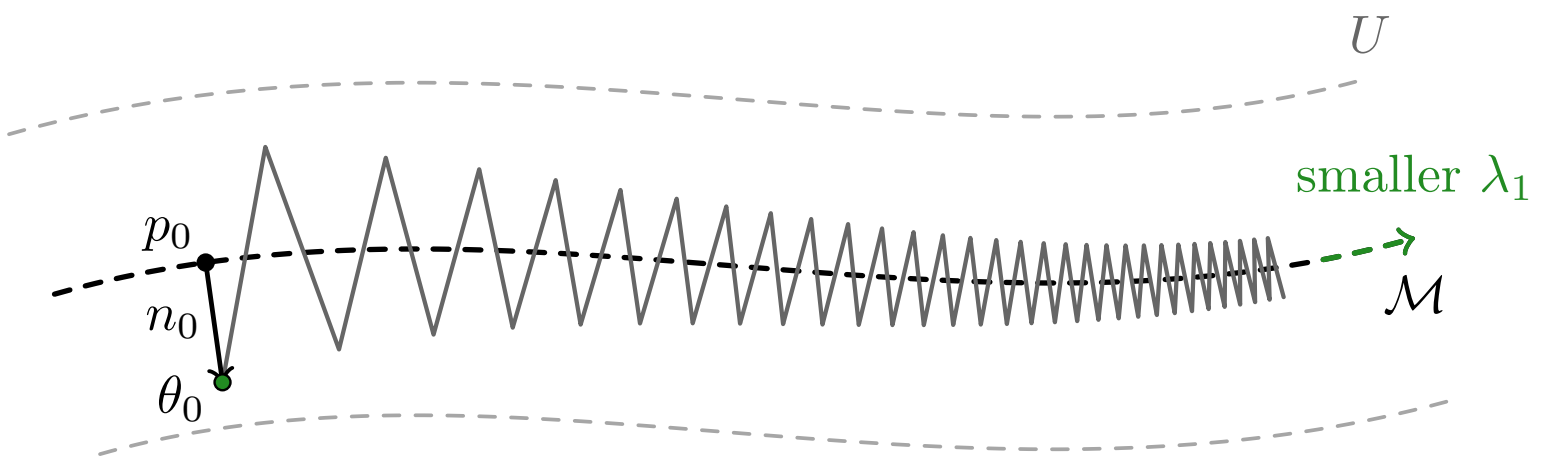
\begin{figure}
	\centering
	\tikzexternaldisable
	\begin{tikzpicture}[x = 1cm, y = 1cm, line cap = round, line join = round, every node/.style = {font = \small}]
	\pgfmathdeclarefunction{manifold}{1}{\pgfmathparse{0.006 * (#1 - 4.5)^3 - 0.08 * (#1 - 4.5)}}
	\pgfmathdeclarefunction{slope}{1}{\pgfmathparse{0.018 * (#1 - 4.5)^2 - 0.08}}
	\pgfmathdeclarefunction{normalx}{1}{\pgfmathparse{-slope(#1) / sqrt(1 + slope(#1)^2)}}
	\pgfmathdeclarefunction{normaly}{1}{\pgfmathparse{1 / sqrt(1 + slope(#1)^2)}}
	\pgfmathdeclarefunction{position}{1}{\pgfmathparse{1 + 7.1 * ln(1 + #1 / 6) / ln(11)}}
	\pgfmathdeclarefunction{amplitude}{1}{\pgfmathparse{(-1)^(#1 + 1) * 0.8 / sqrt(1 + #1 / 4)}}
	\foreach \s in {-1.1, 1.1} {
		\draw[black!35, dashed, line width = 0.7pt] plot[domain = 0:9, samples = 100] ({\x + \s * normalx(\x)}, {manifold(\x) + \s * normaly(\x)});
	}
	\draw[->, black, dashed, line width = 1pt] plot[domain = 0:9, samples = 100] (\x, {manifold(\x)});
	\begin{scope}
		\clip (8.34, -1) rectangle (9.2, 1);
		\draw[->, forestgreen, dashed, line width = 1pt] plot[domain = 0:9, samples = 100] (\x, {manifold(\x)});
	\end{scope}
	\node[below = 6pt, inner sep = 1pt] at (9, {manifold(9)}) {$\mathcal{M}$};
	\node[above = 6pt, inner sep = 1pt, forestgreen] at (9, {manifold(9)}) {smaller $\lambda_1$};
	\node[above = 2pt, black!60] at (8.7, {manifold(8.7) + 1.1}) {$U$};
	\coordinate (pzero) at (1, {manifold(1)});
	\coordinate (thetazero) at ({1 - 0.8 * normalx(1)}, {manifold(1) - 0.8 * normaly(1)});
	\draw[black!60, line width = 0.8pt] (thetazero)
		\foreach \k in {1,...,60} {
			-- ({position(\k) + amplitude(\k) * normalx(position(\k))}, {manifold(position(\k)) + amplitude(\k) * normaly(position(\k))})
		};
	\draw[->, black, line width = 0.9pt] (pzero) -- (thetazero) node[midway, left = 1pt, inner sep = 1pt] {$n_0$};
	\filldraw[fill = black, draw = black, line width = 0.45pt] (pzero) circle (1.5pt);
	\filldraw[fill = forestgreen, draw = black, line width = 0.45pt] (thetazero) circle (1.5pt);
	\node[above left = 1pt, inner sep = 1pt] at (pzero) {$p_0$};
	\node[left = 2pt, yshift = -3pt, inner sep = 1pt] at (thetazero) {$\theta_0$};
\end{tikzpicture}
	\tikzexternalenable
	\caption{CDP dynamics near the manifold of loss minima $\mathcal{M}$.
	Here $\theta_0$ is the initial iterate, $p_0$ its nearest point on $\mathcal{M}$, and $n_0 = \theta_0 - p_0$ the normal displacement.
	Oscillations bias the iterates toward flatter minima, but this drift slows as the loss decreases.
	Our stepsizes exploit this tension, damping the oscillations gradually to reduce the loss while sustaining motion along $\mathcal{M}$.}
	\label{fig:cdp_dynamics_schematic}
\end{figure}

Modern machine learning is predicated on efficient optimization in high dimensions.
A prototypical example is learning the parameters $\theta$ of a model such as a neural network by minimizing an empirical risk $\mathcal{L} (\theta)$ using a first-order method such as gradient descent (GD)
\begin{equation} \label{eq:gd_intro}
	\theta_{k + 1} = \theta_k - \eta \nabla \mathcal{L}(\theta_k),
\end{equation}
where the learning rate (a.k.a.\ stepsize) $\eta$ may depend adaptively on the values and gradients of the objective.
A hallmark of the loss functions that commonly appear in practice is that they are overparameterized, i.e., there are more model parameters than training data points.
In overparameterized models, the set of global loss minima can form large, continuously varying families, and empirical studies find low-loss paths connecting trained solutions \cite{landscape_cooper_2018,connectivity_garipov_2018}.

Although all global minima achieve the same training loss, their out-of-sample performance can differ substantially.
Obtaining high-quality models therefore depends on carefully setting stepsizes and other optimization hyperparameters, as it is the optimizer that determines which loss minimum is ultimately selected.
This is often called the implicit bias of the optimizer.

This paper presents a new approach for biasing GD toward flatter minima, i.e., minimizers with small sharpness, as measured by the top eigenvalue of the loss Hessian.
Specifically, we propose and analyze the \textit{critically damped Polyak (CDP) stepsize rule}
\begin{equation}
	\label{eq:cdp_step_size}
	\eta_\kappa = 4 e^{-\kappa \mathcal{L}(\theta)} \frac{\mathcal{L}(\theta)}{\norm{\nabla \mathcal{L}(\theta)}^2},
\end{equation}
with a single damping parameter $\kappa \geq 0$.
For losses with global minimum value $0$, as in many overparametrized regression problems, this rule multiplies the classical Polyak stepsize $\mathcal{L}(\theta) / \norm{\nabla\mathcal{L}(\theta)}^2$ \cite{optimization_polyak_1987} by $4$ and the damping factor $e^{-\kappa\mathcal{L}(\theta)}$.
In the discussion below, we refer to GD with this stepsize rule simply as CDP.
Our main result gives estimates that are uniform in time: for sufficiently smooth overparameterized losses that are locally quadratic near their minima, CDP drives the loss to zero while its projected iterates move toward flatter minima.
To state things succinctly, denote by $\lambda_1(p)$ the top eigenvalue of the loss Hessian at a point $p$.

\begin{theorem}[Informal] \label{thm:cdp_sharpness_dynamics_informal}
	Suppose $\mathcal{L}: \mathbb{R}^d \to \mathbb{R}_{\geq 0}$ is a smooth loss with global minimum value $0$.
	Suppose $\mathcal{L}$ is locally quadratic near $\mathcal{M} = \mathcal{L}^{-1}(0)$.
	If we initialize $\theta_0$ sufficiently close to $\mathcal{M}$, then for sufficiently large damping $\kappa$, CDP iterates $\theta_k$ with stepsizes from \cref{eq:cdp_step_size} satisfy
	\begin{equation}
		\begin{aligned}
			\norm{\theta_k - p_k} & \simeq \frac{\mathrm{const}}{\sqrt{k}}, & p_{k + 1} & = p_k - \norm{\theta_k - p_k}^2\nabla^{\mathcal{M}} \log{\lambda_1(p_k)} + \bigo{k^{-3/2}},
		\end{aligned}
		\label{eq:cdp_sharpness_dynamics_informal}
	\end{equation}
	where $p_k$ is the point on $\mathcal{M}$ nearest to $\theta_k$, and $\nabla^{\mathcal{M}}$ is the Riemannian gradient along $\mathcal{M}$ (see \Cref{fig:cdp_dynamics_schematic}).
\end{theorem}

See \Cref{thm:localized_cdp_dynamics} for the formal statement and \Cref{def:locally_quadratic_loss} for the precise definition of locally quadratic near $\mathcal{M}$.
Optimization schemes that prefer flatter minima are motivated by empirical and theoretical work linking flatness (as measured by spectral properties of the loss Hessian) to various notions of robustness \cite{sam_foret_2021,stability_wu_2018}.
Broadly speaking, there are two strands of work on first-order optimizers that seek flat minima.
The first considers optimizers such as Sharpness-Aware Minimization (SAM) \cite{sam_foret_2021} and Adaptive SAM \cite{asam_kwon_2021}, which replace the loss by its worst-case value in a neighborhood of the parameters and approximate gradient descent on this modified objective using two gradient evaluations per step.
The practical success of SAM-type algorithms has motivated theoretical analyses showing that, locally in time, these optimizers decrease various notions of sharpness, including the top Hessian eigenvalue \cite{sam_bartlett_2023,sam_wen_2023}.
The bounds in these analyses generally deteriorate as the time horizon $T$ grows\footnote{E.g., the results in \citet[Theorem 4.5]{sam_wen_2023} only concern $T / (\eta \rho^2)$ steps of learning, with learning rate $\eta$ and perturbation radius $\rho$, where $T$ is a fixed constant.}, whereas our results are uniform in $T$.

\begin{figure}[tb]
	\centering
	\includegraphics[width = \linewidth]{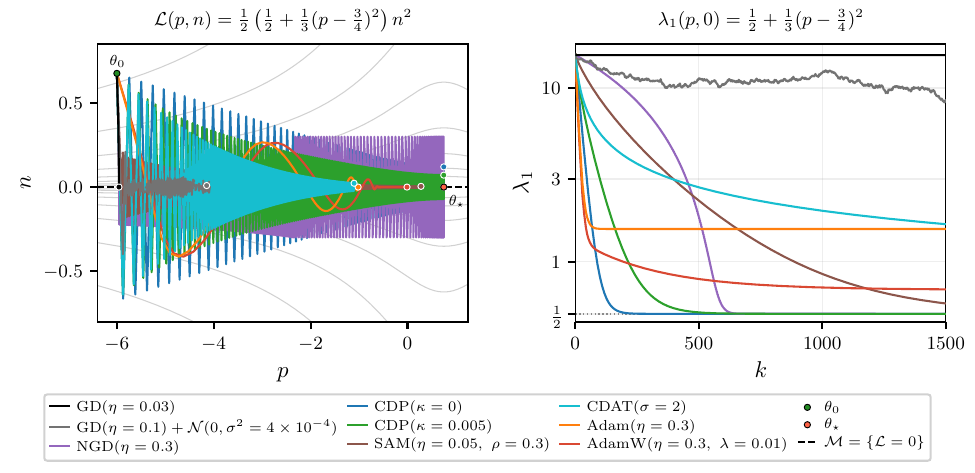}
	\caption{Optimizer trajectories in the $(p, n)$ plane (left) and sharpness at their projections $(p_k, 0)$ onto the manifold of minimizers $\mathcal{M} = \{n = 0\}$ (right), starting from $\theta_0 = (-6, 0.675)$ for $1500$ updates.
	CDP stepsizes sustain motion toward flatter minima as the loss decreases.}
	\label{fig:toy_optimizer_comparison}
\end{figure}

The second strand considers normalized gradient descent (NGD) \cite{eos_arora_2022,ngd_josz_2026} and Curvature Dynamics Aware Tuning (CDAT) \cite{cdat_roulet_2024}.
A core theme in these articles, illustrated in \Cref{fig:cdp_dynamics_schematic,fig:toy_optimizer_comparison}, also provides the starting point for our analysis.
To explain it, consider an overparameterized loss $\mathcal{L}$ with a positive-dimensional manifold of global minima $\mathcal{M}$.
For a first-order optimizer to drift toward flatter minima, it must avoid converging too quickly to $\mathcal{M}$.
Since the negative gradient of $\mathcal{L}$, regardless of normalization, essentially points toward $\mathcal{M}$, a natural option is to choose adaptive stepsizes that cause the iterates $\theta_k$ of \cref{eq:gd_intro} to oscillate around $\mathcal{M}$.
This produces \textit{fast dynamics} perpendicular to $\mathcal{M}$, in which iterates typically align with the top eigenvector of the loss Hessian, and \textit{slow dynamics} parallel to $\mathcal{M}$.
\citet[Theorem 4.4]{eos_arora_2022} show that the limiting dynamics decrease the top eigenvalue of the loss Hessian locally in time, while we obtain guarantees uniform in time.
In contrast, \citet{ngd_josz_2026} can give guarantees uniform in time and treats nonsmooth objectives, but does not fully cover the class of locally quadratic losses considered here.
See \Cref{sec:prior_work} for a more detailed discussion.

Before reporting on experiments with CDP stepsizes, we briefly comment on several aspects of \Cref{thm:cdp_sharpness_dynamics_informal}:
\begin{itemize}[label = {--}, listparindent = 0pt]
	\item \textit{Balancing loss and sharpness reduction.}
	The second relation in \cref{eq:cdp_sharpness_dynamics_informal} describes the implicit bias: to leading order, the projected CDP iterates follow GD on $\log\lambda_1$ along $\mathcal{M}$, thereby moving toward flatter minima.
	The first relation establishes convergence to $\mathcal{M}$.
	These effects must be balanced: approaching $\mathcal{M}$ reduces the loss but also shrinks the effective stepsize $\norm{\theta_k - p_k}^2$ for sharpness reduction.
	Converging too quickly can therefore limit progress toward flatter minima.
	This motivates hybrid schedules that combine a CDP phase for sharpness reduction with GD or Adam phases for loss reduction, as in \Cref{fig:hyperbola_optimizer_trajectories,fig:qwen_sharpness_perplexity_sequence_512}.
	\item \textit{Convergence along $\mathcal{M}$.}
	The effective stepsizes in \cref{eq:cdp_sharpness_dynamics_informal} are of order $1/k$, so their sum diverges while their squares are summable.
	This enables continued motion along $\mathcal{M}$: although initialization must be sufficiently close to $\mathcal{M}$, the projected iterates need not remain close to $p_0$.
	Under additional assumptions, \Cref{thm:morse_bott_cdp_convergence} establishes convergence of the full iterates to an interpolating solution that locally minimizes sharpness.

	\item \textit{Adaptive edge of stability scaling.}
	The edge of stability (EoS) is a regime in which full-batch GD with a fixed stepsize $\eta$ exhibits oscillations while the sharpness $\lambda_1(\nabla^2\mathcal{L})$ stabilizes near $2/\eta$ \cite{stability_wu_2018,stabilization_damian_2023,eos_cohen_2021}.
	The CDP trajectories exhibit similar oscillations, as illustrated in \Cref{fig:hyperbola_optimizer_trajectories,fig:ratio_circle_optimizer_trajectories}, but for a different reason: here the stepsize adapts to the local sharpness rather than the other way around.
	As explained in \Cref{sec:cdp_proof_outline}, for CDP iterates the normal displacement $n_k = \theta_k - p_k$ rapidly aligns with the top Hessian eigenvector $v_1(p_k)$.
	In this aligned regime, the local quadratic approximation gives
	\begin{equation*}
		\mathcal{L}(\theta_k) \approx \frac{\lambda_1(p_k)}{2}\norm{n_k}^2.
	\end{equation*}
	Since the damping factor $e^{-\kappa\mathcal{L}(\theta_k)}$ approaches $1$ as the loss vanishes, CDP stepsizes scale as
	\begin{equation*}
		\eta_\kappa(\theta_k) = 4 e^{-\kappa \mathcal{L}(\theta_k)} \frac{\mathcal{L}(\theta_k)}{\norm{\nabla \mathcal{L}(\theta_k)}^2} \approx \frac{2}{\lambda_1(p_k)}.
	\end{equation*}
	Thus CDP dynamically adapts its stepsize to the local EoS scale, with the factor $4$ giving the precise leading-order coefficient.

	\item \textit{Contrast with normalized GD.}
	A closely related result was obtained by \citet{eos_arora_2022} for Normalized Gradient Descent (NGD)
	\begin{equation*}
		x_{k + 1} = x_k - \eta\frac{\nabla\mathcal{L}(x_k)}{\norm{\nabla\mathcal{L}(x_k)}}.
	\end{equation*}
	They show that, under an additional initialization condition and with vanishingly small added perturbations, the projected NGD iterates track gradient flow on $\log\lambda_1$ over fixed finite rescaled time intervals as $\eta \to 0$, while remaining within an $\bigo{\eta}$ neighborhood of $\mathcal{M}$.

	The key difference is that NGD here takes steps of fixed length (additive updates), whereas CDP steps shrink with the distance $\norm{n_k}$ to $\mathcal{M}$ (multiplicative updates).
	This enables CDP iterates to converge to interpolating solutions and yields global-in-time convergence to a local sharpness minimum under the assumptions of \Cref{thm:morse_bott_cdp_convergence}.
	The distinction is already apparent for the one-dimensional quadratic loss $q(n) = \frac{\lambda}{2} n^2$, where $\mathcal{M} = \{0\}$.
	For this loss, the NGD and CDP updates are
	\begin{equation*}
		\begin{aligned}
			\text{NGD: } & n_{k + 1} = n_k - \eta \, \mathrm{sign}(n_k), \\
			\text{CDP: } & n_{k + 1} = \big(1 - 2e^{-\kappa q(n_k)}\big) n_k.
		\end{aligned}
	\end{equation*}
	Thus, for fixed $\eta$, NGD cannot generally converge to $\mathcal{M}$, whereas CDP converges for any $\kappa > 0$ in this example.
	For empirical comparisons of NGD and CDP on a toy model and in LLM fine-tuning, see \Cref{fig:toy_optimizer_comparison} and \Cref{tab:llm_sharpness_comparison}.

	\item \textit{Role of $\kappa$.}
	In the scalar quadratic example above, $\kappa = 0$ gives $n_{k + 1} = -n_k$, preserving the distance to $\mathcal{M}$, whereas any $\kappa > 0$ contracts the normal oscillations.
	For general losses, higher-order terms also affect the oscillation amplitude.
	Our convergence theorem requires sufficiently large $\kappa$ to ensure that damping dominates these effects.

	\item \textit{Normal quadratic structure in smooth neural networks.}
	Our normal quadratic assumption arises naturally for smooth neural networks trained with mean-squared error when zero training loss is attainable (common in overparametrized regimes).
	If the empirical tangent kernel (i.e., the Gram matrix of the model's parameter gradients on the training data) is positive definite at nearby interpolating solutions, then $\mathcal{M}$ is locally a smooth manifold and the Hessian is positive definite in normal directions.
	This gives the Morse--Bott structure underlying our analysis.
	See \Cref{sec:nonlinear_morse_bott_structure} for details.
\end{itemize}

\subsection{Examples} \label{subsec:optimization_examples}

We now illustrate the effect of CDP stepsizes on visualizable two-dimensional toy losses (\Cref{subsubsec:toy_loss_examples}), student-teacher networks (\Cref{subsubsec:student_teacher_experiments}), and fine-tuning of pre-trained LLMs where we also compare with other optimizers (\Cref{subsubsec:pretrained_llm_experiments}).
Throughout the neural network and LLM experiments, we estimate sharpness using the Lanczos method for tractability, retaining the notation $\lambda_1$ for these estimates.

\subsubsection{Two-dimensional toy models} \label{subsubsec:toy_loss_examples}

\begin{figure}[tbp]
	\centering
	\includegraphics[width = 0.9\linewidth]{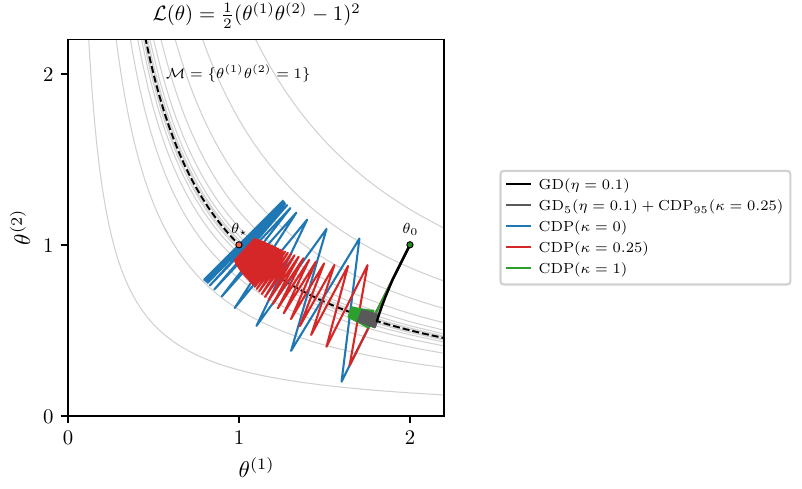}
	\caption{GD, CDP, and hybrid trajectories for $\mathcal{L}(\theta) = \frac{1}{2}(\theta^{(1)}\theta^{(2)} - 1)^2$, starting from $\theta_0 = (2, 1)$ for $100$ steps.
	The hybrid schedule uses $5$ GD steps followed by $95$ CDP steps.
	CDP induces drift along the manifold of minima $\mathcal{M}$ toward $\theta_\star = (1, 1)$, where sharpness is minimized.}
	\label{fig:hyperbola_optimizer_trajectories}
\end{figure}

Consider first the two-dimensional loss
\begin{equation}
	\mathcal{L}(\theta^{(1)}, \theta^{(2)}) = \frac{1}{2}(\theta^{(1)}\theta^{(2)} - 1)^2,
	\label{eq:toy_hyperbola_loss}
\end{equation}
which can be thought of as the mean-squared error over a dataset of one datapoint $(x, y) = (1, 1)$ for a toy linear network $f_\theta(x) = \theta^{(1)}\theta^{(2)}x$.
The set of global minima for $\mathcal{L}$ is the hyperbola $\mathcal{M} = \{\theta^{(1)}\theta^{(2)} = 1\}$, and the flattest minima occur on the diagonal $\theta^{(1)} = \theta^{(2)} \in \{\pm 1\}$.
In \Cref{fig:hyperbola_optimizer_trajectories}, we compare trajectories with a common starting point and a fixed number of steps using CDP with various $\kappa$.
Two phenomena are immediately apparent:
\begin{itemize}[label = {--}, listparindent = 0pt]
	\item \textit{Convergence.}
	The displayed trajectories with $\kappa > 0$ move toward one of the flattest minima, whereas the trajectory with $\kappa = 0$ approaches a period-$2$ orbit around a flattest minimum.
	\item \textit{Damping.}
	Increasing the damping parameter $\kappa > 0$ produces stronger contraction toward $\mathcal{M}$, but less drift along it.
	These trajectories illustrate the tradeoff suggested by the effective $p_k$ stepsize in \cref{eq:cdp_sharpness_dynamics_informal}.
\end{itemize}

\begin{figure}[htb!]
	\centering
	\includegraphics[width = 0.49\linewidth]{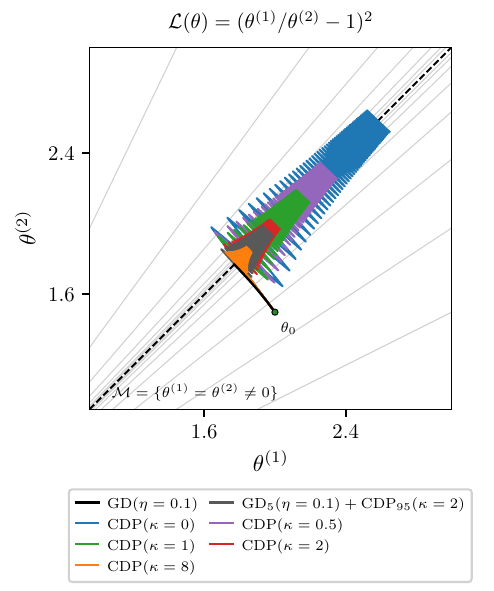}
	\hfill
	\includegraphics[width = 0.49\linewidth]{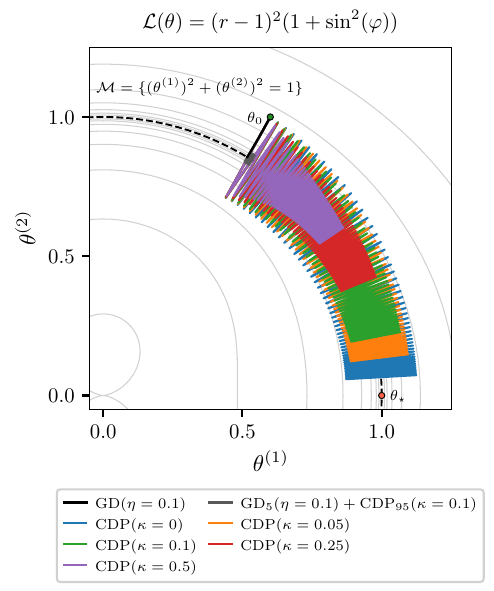}
	\caption{Analog of \Cref{fig:hyperbola_optimizer_trajectories} for two other losses.
	CDP induces drift toward lower sharpness along the manifold of minima $\mathcal{M}$.
	Left: $\mathcal{M}$ is the diagonal, with sharpness decreasing outward along it and no finite flattest minimum.
	Right: $\mathcal{M}$ is the circle, with a flattest minimum at $\theta_\star = (1, 0)$.}
	\label{fig:ratio_circle_optimizer_trajectories}
\end{figure}

The same tendency toward flatter minima is illustrated in \Cref{fig:ratio_circle_optimizer_trajectories} for two other losses.
The first is
\begin{equation*}
	\mathcal{L}(\theta) = (\theta^{(1)} / \theta^{(2)} - 1)^2,
\end{equation*}
whose minimum set on $\theta^{(2)} \neq 0$ is the diagonal $\theta^{(1)} = \theta^{(2)}$, with sharpness $4 / (\theta^{(1)})^2$ and no finite flattest minimum (see \Cref{fig:ratio_circle_optimizer_trajectories} left).
The second, expressed in polar coordinates $(r, \varphi)$, is
\begin{equation*}
	\mathcal{L}(\theta) = (r - 1)^2(1 + \sin^2\varphi),
\end{equation*}
whose minimum set is the unit circle $r = 1$, with flattest minima at $\varphi = 0, \pi$ (see \Cref{fig:ratio_circle_optimizer_trajectories} right).\footnote{These two losses are not globally smooth, but they have the smooth Morse--Bott structure of \Cref{def:locally_quadratic_loss} near the displayed portions of $\mathcal{M}$, which is sufficient for the local analysis.}

\begin{figure}[htb!]
	\centering
	\captionsetup{skip = 5pt}
	\includegraphics[width = \linewidth]{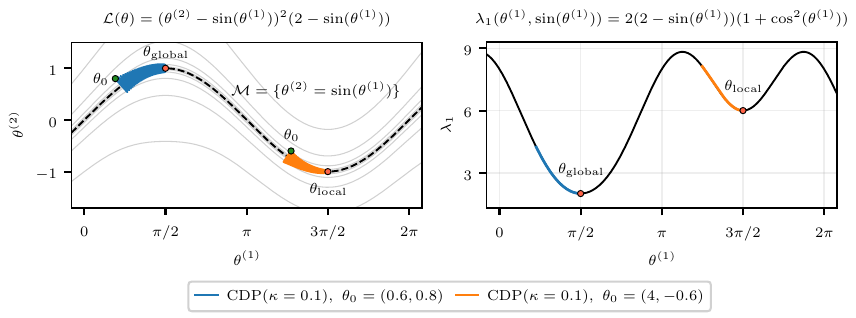}
	\caption{CDP trajectories with $\kappa = 0.1$ for $2000$ steps on \cref{eq:toy_sine_loss}, from two different initializations.
	Left: trajectories over loss contours, with the minimizer manifold $\mathcal{M}$ shown dashed.
	Right: sharpness $\lambda_1$ along $\mathcal{M}$, parametrized by $\theta = (\theta^{(1)}, \sin(\theta^{(1)}))$.
	Depending on initialization, CDP approaches either a global or a strict local minimum of sharpness.}
	\label{fig:sine_sharpness_basins}
\end{figure}

\begin{remark}[Local versus global sharpness minima] \label{rem:sine_sharpness_minima}
	The dynamics in \Cref{thm:cdp_sharpness_dynamics_informal} need not approach a globally flattest minimum.
	For example, the loss
	\begin{equation}
		\mathcal{L}(\theta^{(1)}, \theta^{(2)}) = (\theta^{(2)} - \sin(\theta^{(1)}))^2(2 - \sin(\theta^{(1)})),
		\label{eq:toy_sine_loss}
	\end{equation}
	has minimum set $\mathcal{M} = \{\theta^{(2)} = \sin(\theta^{(1)})\}$.
	Its sharpness has a global minimum of $2$ at $(\pi / 2, 1)$ and a strict local minimum of $6$ at $(3\pi / 2, -1)$.
	CDP trajectories can be attracted to either, see \Cref{fig:sine_sharpness_basins}.
\end{remark}

\subsubsection{Student-teacher models} \label{subsubsec:student_teacher_experiments}

We next test CDP stepsizes on neural networks in the student-teacher setting.
A fixed teacher network generates labels, and we train a wider student with the same depth, input and output dimensions, and activation to fit them by minimizing mean-squared error.
This construction makes zero training loss attainable (see \Cref{subsec:student_teacher_details} for setup details).

\begin{figure}[tbp]
	\centering
	\begin{subfigure}[t]{0.49\linewidth}
		\centering
		\includegraphics[width = \linewidth]{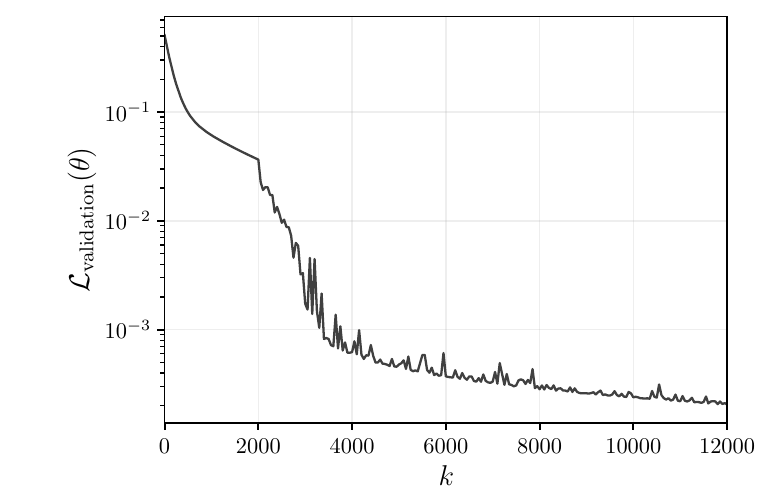}
		\phantomsubcaption
		\label{fig:student_teacher_loss}
	\end{subfigure}
	\hfill
	\begin{subfigure}[t]{0.49\linewidth}
		\centering
		\includegraphics[width = \linewidth]{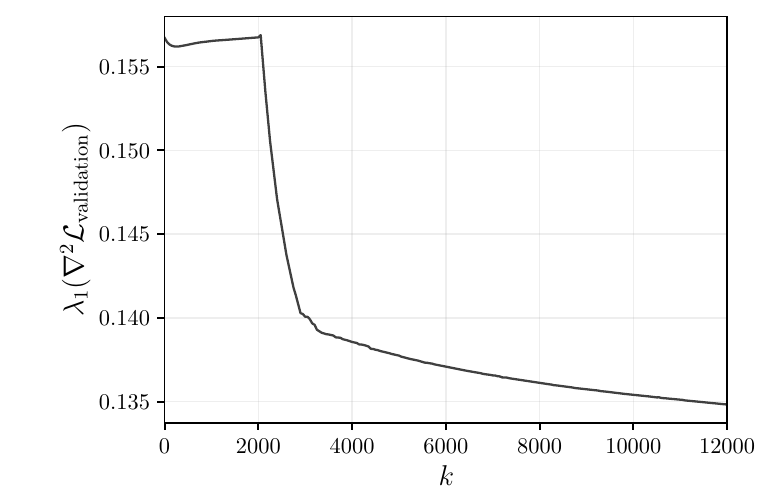}
		\phantomsubcaption
		\label{fig:student_teacher_sharpness}
	\end{subfigure}
	\caption{Validation loss (left) and top eigenvalue of the validation-loss Hessian (right) for a fully connected ReLU student-teacher network with $N_0 = 8$, $N_2 = 1$, $N_1^{\mathrm{teacher}} = 8$, $N_1^{\mathrm{student}} = 512$, and $4096$ training examples.
	Training uses SGD with batch size $256$ and $\eta = 0.06$ for $2000$ steps, followed by $10^4$ steps with CDP stepsizes ($\kappa = 20$) and the same batch size.
	The validation loss decreases in both phases, whereas sharpness decreases only after switching to CDP.}
	\label{fig:student_teacher_loss_and_sharpness}
\end{figure}

In \Cref{fig:student_teacher_loss_and_sharpness}, we first train a fully connected ReLU network with fixed-stepsize SGD and then switch to CDP stepsizes.
The validation loss decreases during both phases, whereas the top validation-loss Hessian eigenvalue remains approximately constant during SGD and decreases only after switching to CDP.
The same qualitative phenomenon holds for deeper residual ReLU networks (see \Cref{fig:residual_student_teacher_loss_and_sharpness}).
These experiments illustrate CDP's flattening effect for ReLU networks beyond the smoothness assumptions of our theory.

\begin{figure}[tbp]
	\centering
	\includegraphics[width = 0.49\textwidth]{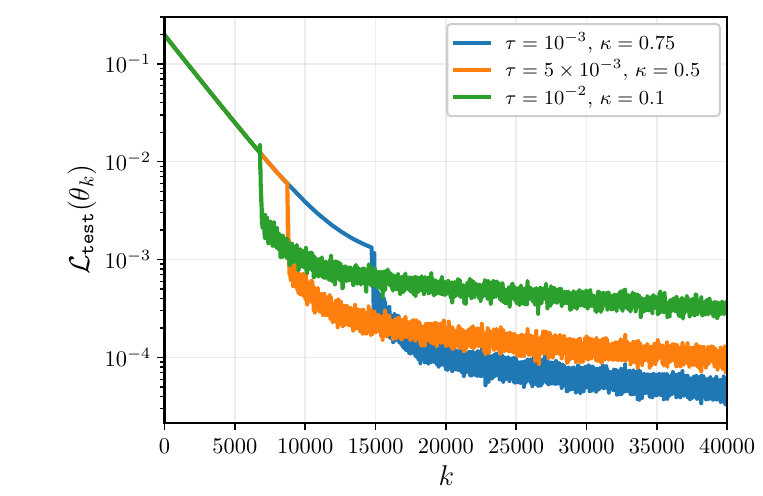}
	\includegraphics[width = 0.49\textwidth]{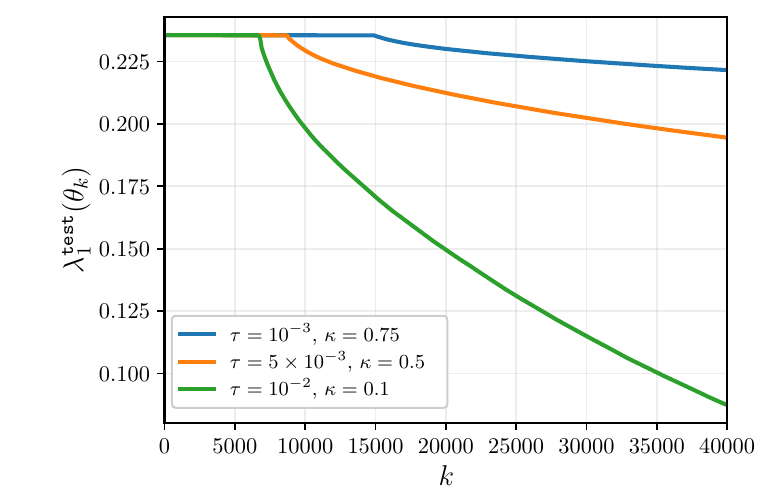}
	\caption{Test loss (left) and top eigenvalue of the test-loss Hessian (right) for smooth $\tanh$ student-teacher networks, with varying CDP switching thresholds $\tau$ and damping parameters $\kappa$.
	Earlier switching with weaker damping slows loss decreases, but accelerates sharpness decrease.}
	\label{fig:deep_network_polyak_threshold_comparison}
\end{figure}

For smooth $\tanh$ networks, \Cref{fig:deep_network_polyak_threshold_comparison} shows that earlier switching with weaker damping slows loss reduction but produces a larger decrease in sharpness.

All the displayed student-teacher experiments use the neural tangent kernel (NTK) parameterization \cite{ntk_jacot_2018}.
With a mean-field parameterization, however, we consistently observe that CDP increases sharpness.
We conjecture that this difference reflects the greater spectral degeneracy of the Hessian in mean-field networks and leave a careful investigation to future work.

\subsubsection{Pre-trained LLMs} \label{subsubsec:pretrained_llm_experiments}

We next investigate CDP stepsizes for fine-tuning pre-trained LLMs on \texttt{WikiText-2} using cross-entropy loss.
Our theoretical results do not apply directly to this setting, since cross-entropy loss need not attain its infimum at finite parameters.
Nevertheless we still ask: can CDP reduce sharpness relative to standard optimizers?

\begin{figure}[htbp!]
	\centering
	\includegraphics[width = 0.49\textwidth]{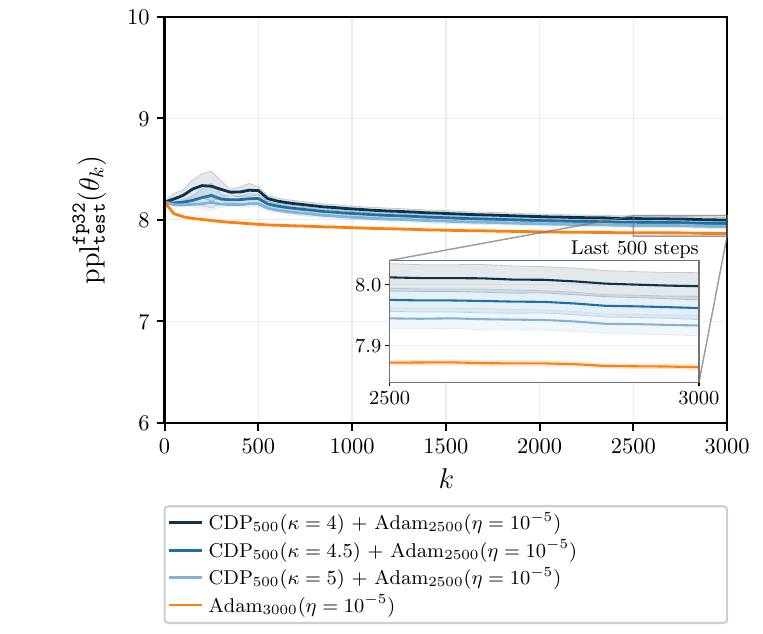}
	\includegraphics[width = 0.49\textwidth]{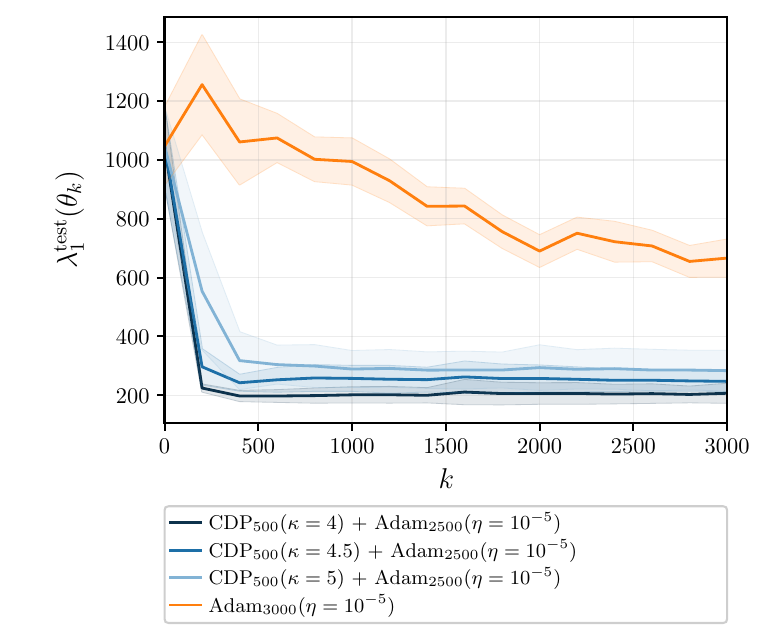}
	\caption{
		Fine-tuning \texttt{Qwen2.5-1.5B} on \texttt{WikiText-2} (training sequence length $512$, sliding-window test evaluation, $10$ seeds).
		Left: test perplexity, with a zoom on the final $500$ steps.
		Right: top test-loss Hessian eigenvalue $\lambda_1$, estimated by Lanczos.
		Adam($\eta = 10^{-5}$) is compared with $500$ CDP steps ($\kappa \in \{4, 4.5, 5\}$) followed by $2500$ Adam steps.
		The CDP phase lowers $\lambda_1$ substantially and the gap persists throughout the Adam phase, at the cost of $\approx 1\%$ higher final test perplexity.
	}
	\label{fig:qwen_sharpness_perplexity_sequence_512}
\end{figure}

For the pre-trained LLM \texttt{Qwen2.5-1.5B} \cite{qwen_yang_2024}, \Cref{fig:qwen_sharpness_perplexity_sequence_512} compares Adam with a hybrid schedule designed to reduce sharpness using CDP before switching to Adam to improve the loss.
The initial CDP phase drives the top Hessian eigenvalue $\lambda_1$ substantially below its value under Adam alone.
Subsequent Adam updates reduce the loss without undoing the flatter region found using CDP.
The price is a slightly higher final test perplexity than under Adam alone.

For the pre-trained LLM \texttt{GPT-2}, \Cref{tab:llm_sharpness_comparison} compares AdamW \cite{adamw_loshchilov_2019}, normalized gradient descent (NGD) \cite{eos_arora_2022}, and CDP.
CDP reduces $\lambda_1$ from $1475$ at initialization to $27$ after $400$ fine-tuning steps, nearly $20$ times smaller than AdamW's final value, at a cost of $0.1$ nats in validation loss.
In contrast, NGD does not reduce $\lambda_1$ at all.
The failure of NGD to reduce $\lambda_1$ appears to be a quirk of \texttt{GPT-2}, and we found that NGD can be competitive or even best for sharpness reduction in other settings (see \Cref{tab:qwen_small_sharpness}).
The step cap appears helpful for CDP: with $\eta_{\max} = 0.1$, sharpness decreases across all three seeds, whereas with larger caps of $0.3$ or $1$, one of three seeds moves away from the pre-trained weights and exhibits increasing sharpness (details in \Cref{subsec:llm_fine_tuning_details}).
This sensitivity is consistent with the local nature of \Cref{thm:cdp_sharpness_dynamics_informal}.

Full details for these experiments are provided in \Cref{subsec:llm_fine_tuning_details}, with additional fine-tuning and post-training quantization results in \Cref{subsec:qwen_small_quantization}.
We also report comparisons with competing methods after substantial parameter tuning (see \Cref{sec:experiments}).

\begin{table}[htbp!]
	\centering\footnotesize
	\caption{Top Hessian eigenvalue $\lambda_1$ and validation loss when fine-tuning \texttt{GPT-2} on \texttt{WikiText-2} for $400$ steps at sequence length $256$, reporting mean over $3$ seeds.
	At initialization, $\lambda_1 = 1475$ and $\mathcal{L}_{\text{validation}} = 3.73$.
	$\lambda_1$ is reported after partial fine-tuning ($k = 50$) and completion ($k = 400$).
	The step cap $\eta_{\max}$ bounds the per-step parameter movement under CDP.}
	\label{tab:llm_sharpness_comparison}
	\setlength{\tabcolsep}{10pt}
	\begin{tabular}{l cc c}
		\toprule
		\texttt{optimizer} & $\lambda_1$ ($k = 50$) & $\lambda_1$ ($k = 400$) & $\mathcal{L}_{\text{validation}}$ \\
		\midrule
		AdamW($\eta = 5 \times 10^{-5}$)       & $1302$ & $486$ & $\mathbf{3.18}$ \\
		NGD($\eta = 10^{-2}$)                  & $1819$ & $1977$ & $3.46$ \\
		CDP($\kappa = 1$, $\eta_{\max} = 0.1$)   & $732$ & $\mathbf{27}$ & $3.28$ \\
		CDP($\kappa = 0.5$, $\eta_{\max} = 0.3$) & $\mathbf{570}$ & $110$ & $3.57$ \\
		\bottomrule
	\end{tabular}
\end{table}

\subsection*{Outline.}
We review prior work in \Cref{sec:prior_work}, state our main result in \Cref{sec:localized_cdp_dynamics}, and outline its proof in \Cref{sec:cdp_proof_outline}.
\Cref{sec:morse_bott_dynamics} proves the main results and establishes convergence to a local minimum of sharpness under additional assumptions.
\Cref{sec:quadratic_morse_bott_dynamics} studies the exact quadratic model, and \Cref{sec:nonlinear_morse_bott_structure} develops the local geometry of regular interpolation manifolds for neural network losses.
\Cref{sec:experiments} collects experimental details and additional results.

\section{Prior Work} \label{sec:prior_work}

\textit{Flatness and generalization}. Our work studies how the choice of stepsize can bias gradient descent toward flatter interpolating minima.
The motivation comes in part from the empirical connection between large-batch training, sharper minima, and poorer generalization \cite{sharpness_keskar_2017}.
Flat minima are less sensitive to small weight perturbations. \citet{bayes_dziugaite_2017} use this robustness to obtain informative PAC-Bayesian generalization bounds for stochastic neural networks.

Flatness depends also on the parameterization; indeed \citet[Theorem 4]{sharpness_dinh_2017} show that rescaling a rectified network can make a minimum with nonzero Hessian arbitrarily sharp without changing the represented function.
Our sharpness statements concern the fixed parameterization in which the optimization is performed.

\textit{Noise and explicit regularization}. Noise can keep SGD iterates moving near a manifold of loss minima, affecting which minima the trajectory approaches.
In the small-learning-rate limit, \citet[Theorem 4.6]{noise_li_2022} describe this motion by a diffusion on the manifold, after rescaling time.
The loss geometry and noise covariance jointly determine this diffusion.
Under their label-noise covariance assumption, the limiting flow follows the negative tangential gradient of the Hessian trace \cite[Corollary 5.2]{noise_li_2022}.
The sampling scheme can introduce a further bias, as illustrated by the local regularization effect of sampling without replacement \cite{sgd_beneventano_2024}.
For CDP, the corresponding drift is deterministic and, under the assumptions of our results, follows the negative gradient of the logarithm of the top Hessian eigenvalue.

Flatness can also be encouraged explicitly by modifying the optimization objective.
Sharpness-Aware Minimization (SAM) \cite{sam_foret_2021} seeks to minimize the worst-case loss in a neighborhood of the parameters, approximating the gradient of this auxiliary objective using an additional gradient evaluation of the original loss at a perturbed point.
\citet[Theorem 21]{sam_bartlett_2023} show that a local expansion of the SAM update contains a drift toward lower top Hessian eigenvalues.
\citet[Theorem 4.5]{sam_wen_2023} establish tracking of a sharpness-decreasing flow over fixed rescaled time intervals, with constants and parameter restrictions depending on the time horizon.
In contrast to SAM (including ASAM \cite{asam_kwon_2021} and Polyak-type variants \cite{sam_polyak_oikonomou_2026}), CDP leaves the objective and gradient direction unchanged, inducing a bias toward flatter minima solely through an adaptive scalar stepsize computed from the current loss and gradient.

\textit{Stepsizes and regularization}. Even without noise, the finite stepsize of GD can act as an implicit regularizer.
\citet[Theorem 3.1]{regularization_barrett_2021} derive a gradient-flow approximation to GD with constant stepsize $\eta$.
To first order in $\eta$, this flow uses the modified loss $\mathcal{L} + \frac{\eta}{4}\norm{\nabla \mathcal{L}}^2$.
This expansion alone does not establish convergence to a minimum of sharpness.

The role of stepsizes in determining convergence rates is better established.
The Polyak rule provides adaptive convergence guarantees in convex optimization \cite{polyak_hazan_2019} and, in stochastic variants, under interpolation \cite{alig_berrada_2020,sps_loizou_2021}.
Nonconstant schedules can accelerate GD \cite{long_steps_grimmer_2023,pep_das_gupta_2024,hedging_altschuler_2023,silver_altschuler_2025,greed_altschuler_2018,arcsine_altschuler_2024,richardson_young_1953}, as surveyed in \cite{hedging_altschuler_2026}.
These results primarily concern convergence rates, rather than which minimizer is selected when the minimum set is nonisolated.
Stepsize choice can also influence feature learning: even the first stepsize can affect the features learned by high-dimensional two-layer networks \cite{features_ba_2022}.
Experiments with adaptive methods show that entering flat saddle regions too early can hinder feature learning and generalization \cite{flatland_galli_2026}.

\textit{Edge of stability}. The connection between stepsize and sharpness becomes particularly direct at EoS.
\citet{eos_cohen_2021} observe that full-batch GD with fixed stepsize $\eta$ often operates in this regime, with sharpness hovering just above $2/\eta$ and loss oscillating over individual steps.
\citet{stabilization_damian_2023} explain this behavior by analyzing a self-stabilization mechanism through which these oscillations oppose further sharpness growth.
Continuous-time descriptions of EoS dynamics include Central Flow \cite{central_cohen_2025}, Rod Flow \cite{rod_regis_2026}, and Edge Flow \cite{edge_marion_2026}.
Related work extends the EoS picture to adaptive methods using the preconditioned Hessian \cite{adaptive_cohen_2022}, reports similar behavior experimentally with capped Polyak stepsizes \cite{sharpening_roulet_2023}, studies deep matrix factorization \cite{linear_ghosh_2025}, and shows that GD can converge to flatter minima than gradient flow for shallow linear networks, revealing a tradeoff between convergence speed and implicit regularization \cite{linear_beneventano_2025}.
Local convergence to flat minima has also been established for least-squares models at a critical constant stepsize \cite{eos_macdonald_2025,flat_minima_macdonald_2026}.

The interaction between stepsize and curvature is also central to analyses of learning rate warmup \cite{warmup_kalra_2024} and curvature-aware learning rate tuning \cite{cdat_roulet_2024,salr_yue_2024}.
In contrast to fixed-stepsize GD, whose sharpness can stabilize near $2/\eta$, CDP adapts its stepsize to the local sharpness, with $\eta_k \approx 2/\lambda_1(p_k)$ near the manifold of minimizers after alignment with the top eigendirection.

\textit{Normalized gradient descent}. The closest theoretical comparison is the analysis of normalized GD by \citet[Theorems 4.3 and 4.4]{eos_arora_2022}.
Under regularity, spectral separation, and additional initialization conditions, they show that a slightly perturbed version of normalized GD exhibits oscillations near the minimizer manifold that align on average with the top Hessian eigendirection.
The projected iterates track gradient flow on $\log\lambda_1$ along the manifold over fixed finite rescaled time intervals as the step parameter $\eta$ tends to zero.
The normal oscillations remain of order $\eta$, reflecting the fixed update length of normalized GD.
For NGD with prescribed diminishing stepsizes, \citet{ngd_josz_2026} establishes convergence to flat minima under assumptions involving an auxiliary Lyapunov function.
In particular, the results in \citet[Theorem 3.32]{ngd_josz_2026} do not cover locally quadratic losses of the form
\begin{equation*}
	\mathcal{L}(n, p) = \frac{1}{2}\lambda_1(p)n^2
\end{equation*}
when the sharpness is positive, $C^5$, and nonsemialgebraic.
For example, $\lambda_1(p) = 1 + \tanh^4(p)$ has a degenerate minimum at $p = 0$.
With sufficiently large damping, CDP converges locally to this minimum.

In contrast, CDP takes steps whose lengths shrink adaptively with the distance to $\mathcal{M}$, allowing the iterates to approach $\mathcal{M}$ while their projections continue moving toward flatter minima.
Their normal distance decays as $k^{-1/2}$, while their effective tangential stepsizes are of order $1/k$ and have divergent sum.
These estimates are uniform in time and imply convergence of CDP iterates to an interpolating solution that locally minimizes sharpness under the assumptions of \Cref{thm:morse_bott_cdp_convergence}.

\section{Main result} \label{sec:localized_cdp_dynamics}
In this section we present our main result, \Cref{thm:localized_cdp_dynamics}.
For a twice differentiable loss $\mathcal{L}: \real^d \rightarrow \real$ with a smooth manifold $\mathcal{M}$ of minima of co-dimension $m$, write
\begin{equation*}
	H(p) \coloneqq \nabla^2 \mathcal{L}(p)|_{N_p\mathcal{M}}: N_p\mathcal{M} \to N_p\mathcal{M}
\end{equation*}
for the Hessian of $\mathcal{L}$ restricted to the normal space $N_p\mathcal{M}$ of $\mathcal{M}$ at $p \in \mathcal{M}$.
We write
\begin{equation*}
	\begin{aligned}
		H(p) = \sum_{i = 1}^m \lambda_i(p)v_i(p)v_i(p)^T, \ \lambda_1(p) \geq \cdots \geq \lambda_m(p)
	\end{aligned}
\end{equation*}
with $(v_i(p))_{i = 1}^m$ an orthonormal eigenbasis of $N_p\mathcal{M}$.

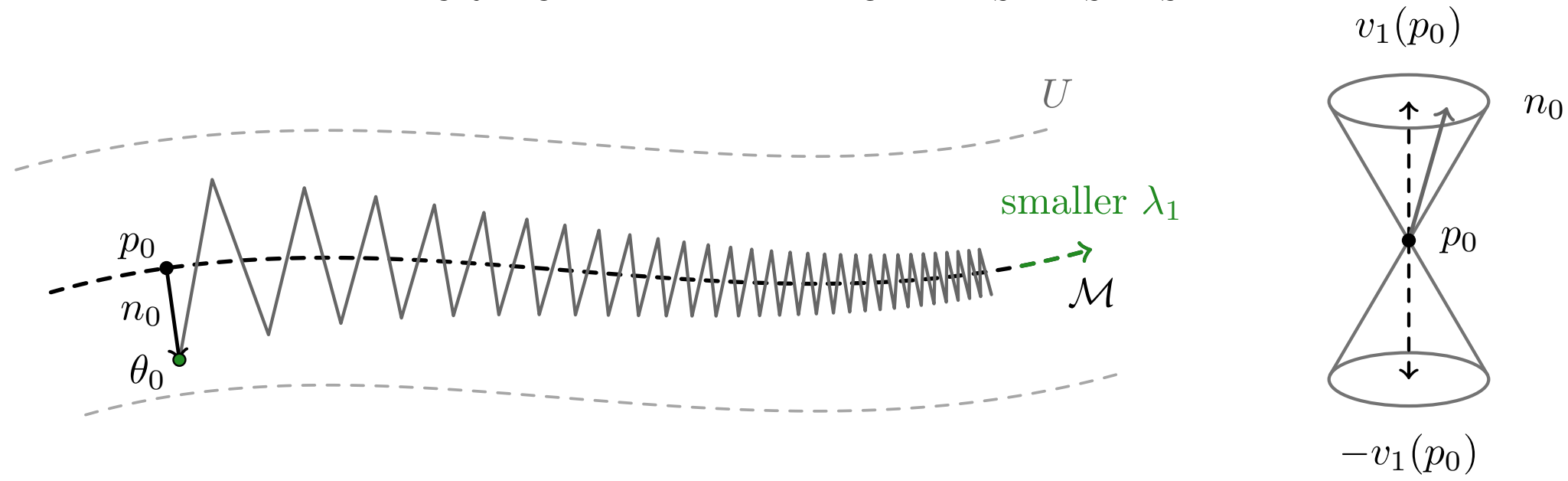
\begin{figure}[tb]
	\vspace{-16pt}
	\centering
	\captionsetup{skip = 4pt}
	\tikzexternaldisable
	\raisebox{-0.5\height}{\begin{tikzpicture}[x = 1cm, y = 1cm, line cap = round, line join = round, every node/.style = {font = \small}]
	\pgfmathdeclarefunction{manifold}{1}{\pgfmathparse{0.006 * (#1 - 4.5)^3 - 0.08 * (#1 - 4.5)}}
	\pgfmathdeclarefunction{slope}{1}{\pgfmathparse{0.018 * (#1 - 4.5)^2 - 0.08}}
	\pgfmathdeclarefunction{normalx}{1}{\pgfmathparse{-slope(#1) / sqrt(1 + slope(#1)^2)}}
	\pgfmathdeclarefunction{normaly}{1}{\pgfmathparse{1 / sqrt(1 + slope(#1)^2)}}
	\pgfmathdeclarefunction{position}{1}{\pgfmathparse{1 + 7.1 * ln(1 + #1 / 6) / ln(11)}}
	\pgfmathdeclarefunction{amplitude}{1}{\pgfmathparse{(-1)^(#1 + 1) * 0.8 / sqrt(1 + #1 / 4)}}
	\foreach \s in {-1.1, 1.1} {
		\draw[black!35, dashed, line width = 0.7pt] plot[domain = 0:9, samples = 100] ({\x + \s * normalx(\x)}, {manifold(\x) + \s * normaly(\x)});
	}
	\draw[->, black, dashed, line width = 1pt] plot[domain = 0:9, samples = 100] (\x, {manifold(\x)});
	\begin{scope}
		\clip (8.34, -1) rectangle (9.2, 1);
		\draw[->, forestgreen, dashed, line width = 1pt] plot[domain = 0:9, samples = 100] (\x, {manifold(\x)});
	\end{scope}
	\node[below = 6pt, inner sep = 1pt] at (9, {manifold(9)}) {$\mathcal{M}$};
	\node[above = 6pt, inner sep = 1pt, forestgreen] at (9, {manifold(9)}) {smaller $\lambda_1$};
	\node[above = 2pt, black!60] at (8.7, {manifold(8.7) + 1.1}) {$U$};
	\coordinate (pzero) at (1, {manifold(1)});
	\coordinate (thetazero) at ({1 - 0.8 * normalx(1)}, {manifold(1) - 0.8 * normaly(1)});
	\draw[black!60, line width = 0.8pt] (thetazero)
		\foreach \k in {1,...,60} {
			-- ({position(\k) + amplitude(\k) * normalx(position(\k))}, {manifold(position(\k)) + amplitude(\k) * normaly(position(\k))})
		};
	\draw[->, black, line width = 0.9pt] (pzero) -- (thetazero) node[midway, left = 1pt, inner sep = 1pt] {$n_0$};
	\filldraw[fill = black, draw = black, line width = 0.45pt] (pzero) circle (1.5pt);
	\filldraw[fill = forestgreen, draw = black, line width = 0.45pt] (thetazero) circle (1.5pt);
	\node[above left = 1pt, inner sep = 1pt] at (pzero) {$p_0$};
	\node[left = 2pt, yshift = -3pt, inner sep = 1pt] at (thetazero) {$\theta_0$};
\end{tikzpicture}}
	\hspace{0.8cm}
	\raisebox{-0.5\height}{\begin{tikzpicture}[x = 1cm, y = 1cm, line cap = round, line join = round, every node/.style = {font = \small}]
	\foreach \sign in {-1, 1} {
		\draw[black!55, line width = 0.8pt] (0, 0) -- (-0.68, {\sign * 1.16});
		\draw[black!55, line width = 0.8pt] (0, 0) -- (0.68, {\sign * 1.16});
		\draw[black!55, line width = 0.8pt] (0, {\sign * 1.2}) ellipse [x radius = 0.69cm, y radius = 0.23cm];
		\draw[->, black, dashed, line width = 0.8pt] (0, 0) -- (0, {\sign * 1.2});
	}
	\draw[->, black!60, line width = 1pt] (0, 0) -- (0.33, 1.15);
	\node[right = 3pt] at (0.75, 1.15) {$n_0$};
	\filldraw[fill = black, draw = black, line width = 0.45pt] (0, 0) circle (1.5pt);
	\node[right = 4pt] at (0, 0) {$p_0$};
	\node[above = 3pt] at (0, 1.43) {$v_1(p_0)$};
	\node[below = 3pt] at (0, -1.43) {$-v_1(p_0)$};
\end{tikzpicture}}
	\tikzexternalenable
	\caption{Schematic CDP dynamics near $\mathcal{M}$. Left: $\theta_k = p_k + n_k$ oscillates across the manifold, with decreasing normal distance and tangential drift toward lower sharpness. Right: the initial alignment condition selects cones about $\pm v_1(p_0)$.}
	\label{fig:nearest_point_decomposition}
	\vspace{-6pt}
\end{figure}

\begin{definition}\label{def:locally_quadratic_loss}
	We say $\mathcal{L}$ is \textit{locally quadratic near its minima} if
	\begin{itemize}[label = {--}]
		\item $\min \mathcal{L} = 0$ and $\mathcal{M} = \mathcal{L}^{-1}(0)$ is a smooth manifold of codimension $m > 0$;
		\item $\mathcal{L}$ is $5$ times continuously differentiable;
		\item $H(p)$ is positive definite for every $p \in \mathcal{M}$;
		\item At each point $p \in \mathcal{M}$ the top Hessian eigenvalue $\lambda_1(p)$ is distinct from the second eigenvalue $\lambda_2(p)$ (in case $m = 1$ this condition is vacuous).
	\end{itemize}
\end{definition}
Since $\mathcal{M}$ is a smooth submanifold of $\mathbb{R}^d$ (by the first bullet of \Cref{def:locally_quadratic_loss}), the Tubular Neighborhood Theorem \cite[Theorem 6.24]{smooth_lee_2013} implies that there exists an open neighborhood $U$ of $\mathcal{M}$ such that every $\theta \in U$ admits a unique nearest-point decomposition
\begin{equation}
	\begin{aligned}
		\theta & = p + n, & p &\in \mathcal{M}, & n &\in N_p\mathcal{M},
	\end{aligned}
	\label{eq:nearest_point_decomposition}
\end{equation}
The interpretation is that $p$ is the closest point of $\mathcal{M}$ to $\theta$ and $\norm{n}$ is the distance from $\theta$ to $\mathcal{M}$, as illustrated in \Cref{fig:nearest_point_decomposition}. Our main result is the following.

\begin{theorem} \label{thm:localized_cdp_dynamics}
	Suppose $\mathcal{L}$ satisfies \Cref{def:locally_quadratic_loss} above.
	For any compact set $K \subseteq \real^d$ and every $\delta > 0$, there exist $\kappa_0 \geq 0$, $C > 0$, and $k_0 \geq 1$ such that for any fixed $\kappa \geq \kappa_0$, there exists $\rho_0 > 0$ with the following property.
	Assume that the GD iterates defined below satisfy $\theta_k \in K$ for every $k \geq 0$.
	Moreover, let $\theta_0 = p_0 + n_0 \in K$ have tubular coordinates as in \cref{eq:nearest_point_decomposition} and satisfy
	\begin{equation*}
		\begin{aligned}
			0 < \norm{n_0} &\leq \rho_0, & \abs{\innerproduct{\frac{n_0}{\norm{n_0}}}{v_1(p_0)}} &\geq \delta
		\end{aligned}
	\end{equation*}
	and define
	\begin{equation*}
		\theta_{k + 1}
		= \theta_k - \frac{4e^{-\kappa \mathcal{L}(\theta_k)}\mathcal{L}(\theta_k)}{\norm{\nabla \mathcal{L}(\theta_k)}^2}\nabla \mathcal{L}(\theta_k).
	\end{equation*}
	Then the iterates admit tubular coordinates
	\begin{equation*}
		\begin{aligned}
			\theta_k & = p_k + n_k, & p_k &\in \mathcal{M}, & n_k &\in N_{p_k}\mathcal{M}
		\end{aligned}
	\end{equation*}
	as in \cref{eq:nearest_point_decomposition}, and for $k \geq k_0$ we have
	\begin{equation*}
		\norm{n_k} \leq \frac{C}{\sqrt{k}} \text{ and } p_{k + 1} = p_k - \norm{n_k}^2 \nabla^\mathcal{M} \log{\lambda_1(p_k)} + \bigo{k^{-3/2}},
	\end{equation*}
	where $\nabla^\mathcal{M}$ is the Riemannian gradient along $\mathcal{M}$.
\end{theorem}

The proof is provided in \Cref{subsec:morse_bott_cdp_convergence_proof}.
\Cref{thm:morse_bott_cdp_convergence} in the appendix strengthens \Cref{thm:localized_cdp_dynamics} to convergence to a single minimum $p_{\star}$ under additional assumptions: the base points remain in a compact region where $\lambda_1$ has a unique critical point $p_\star$, which is a strict local minimum.
One can extract sharper convergence rates under additional growth conditions on $\lambda_1$, e.g., polynomial sharpness rates under an additional local Polyak-\L{}ojasiewicz condition on $\lambda_1$ along $\mathcal{M}$ (\Cref{cor:morse_bott_sharpness_rate}).

In \Cref{sec:cdp_proof_outline}, we give an informal overview of the proof of \Cref{thm:localized_cdp_dynamics}.
Before doing this, we make several remarks:
\begin{itemize}[label = {--}]
	\item The constants $C, k_0$ and the implicit constant in the $\bigo{k^{-3/2}}$ error term can be chosen independently of $\kappa$ and may depend on $\delta$, the norm of $\mathcal{L}$ in $C^5(K, \real)$ and on
	\begin{equation*}
		\begin{aligned}
			\sup_{\theta = n + p \in K \cap U} \lambda_1(p), && \inf_{\theta = n + p \in K \cap U} \bigl(\lambda_1(p) - \lambda_2(p)\bigr), && \inf_{\theta = n + p \in K \cap U} \lambda_m(p)
		\end{aligned}
	\end{equation*}
	which are all strictly positive and finite by \Cref{def:locally_quadratic_loss} and where $U$ is the neighborhood of $\mathcal{M}$ on which we may apply the Tubular Neighborhood Theorem \cite[Theorem 6.24]{smooth_lee_2013}\footnote{For $m = 1$, we set $\lambda_2 = 0$.}.
	\item The initial alignment condition $\abs{\innerproduct{n_0}{v_1(p_0)}} > 0$ is necessary in general.
	This is analogous to how the power method requires the initialization to have non-zero correlation with the leading eigendirection.
	To see this, consider a setting where
	\begin{equation*}
		\theta
		= (\theta^{(1)}, \theta^{(2)}, \theta^{(3)}) \in \real^3 \text{ and } \mathcal{L}(\theta)
		= \frac{1}{2}\lambda_1(\theta^{(3)})(\theta^{(1)})^2 + \frac{1}{2}\lambda_2(\theta^{(3)})(\theta^{(2)})^2
	\end{equation*}
	with $\lambda_1(\theta^{(3)}) > \lambda_2(\theta^{(3)}) > 0.$ Then the set of minima is the line $\theta^{(1)} = \theta^{(2)} = 0$ and the top eigenvector of the Hessian at every minimum is $(1,0,0)$. However, writing the CDP iterates $\theta_k = p_k + n_k = (0,0,\theta_k^{(3)}) + (\theta_k^{(1)},\theta_k^{(2)},0)$, a direct computation shows
	\begin{equation*}
		\innerproduct{n_0}{(1, 0, 0)} = 0 \Rightarrow \innerproduct{n_k}{(1, 0, 0)} = 0 \forall k \geq 0.
	\end{equation*}
	Thus, the dynamics reduces to two-dimensional dynamics in the $(\theta^{(2)},\theta^{(3)})$-plane and CDP iterates will tend to minimize $\lambda_2$ instead of $\lambda_1$.
\end{itemize}

\section{Proof outline} \label{sec:cdp_proof_outline}

Our goal in this section is to provide an intuitive guide to the proof of \Cref{thm:localized_cdp_dynamics}. To do this, we remove the damping by taking $\kappa = 0$ and analyze the CDP updates
\begin{equation*}
	\begin{aligned}
		\theta_{k + 1} & = \theta_k - \frac{4\mathcal{L}(\theta_k)}{\norm{\nabla \mathcal{L}(\theta_k)}^2}\nabla \mathcal{L}(\theta_k), & \theta_k & = p_k + n_k.
	\end{aligned}
\end{equation*}

We begin with an important special case in which we have a global decomposition $\theta = n + p$ and
\begin{equation}
	\begin{aligned}
		\mathcal{L}(n, p) & = \frac{1}{2}n^T Hn, & H & = \sum_{i = 1}^m \lambda_i v_i v_i^T
	\end{aligned}
	\label{eq:frozen_hessian_quadratic_loss}
\end{equation}
for a Hessian that is independent of $p$. We have
\begin{equation}
	\mathcal{M} = \mathcal{L}^{-1}(0) = \set{\theta = p + n \in \real^d \mid n = 0}.
	\label{eq:quadratic_minimum_manifold}
\end{equation}
Since $\mathcal{L}$ is independent of $p$, we have $p_k = p_0$ and also
\begin{equation*}
	n_{k + 1} = n_k - \frac{2n_k^T Hn_k}{\norm{Hn_k}^2}Hn_k.
\end{equation*}
A direct computation (by squaring both sides) then shows
\begin{equation*}
	\norm{n_{k + 1}} = \norm{n_k}.
\end{equation*}
Note that $\norm{n_k}$ is the distance from $\theta_k$ to $\mathcal{M}$, so the preceding equation means that CDP stepsizes precisely preserve the distance to $\mathcal{M}$ in this quadratic model.
This is the motivation for choosing the constant $4$ in the definition of CDP stepsizes.
While the norm of $n_k$ is invariant under the CDP dynamics, the direction of $n_k$ aligns with $v_1$, the top eigendirection of $H$.
To see this, write
\begin{equation*}
	n_k = \sum_{i = 1}^m n_k^i v_i
\end{equation*}
and note that
\begin{equation*}
	n_{k + 1}^{i} = \left(1 - \frac{2\lambda_i}{\lambda_{\mathrm{effective}}(n_k)}\right)n_k^i,
\end{equation*}
where the effective eigenvalue is
\begin{equation*}
	\lambda_{\mathrm{effective}}(n) = \frac{\norm{Hn}^2}{n^T Hn} = \frac{\sum_{i = 1}^m \lambda_i^2 (n^i)^2}{\sum_{i = 1}^m \lambda_i (n^i)^2} \in [\lambda_m, \lambda_1].
\end{equation*}
If $m \geq 2$, $\lambda_1 > \lambda_2$, and $n_k^1 \neq 0$, then for every $i \geq 2$,
\begin{equation*}
	\abs{\frac{n_{k + 1}^i}{n_{k + 1}^1}} = \frac{\abs{2\lambda_i - \lambda_{\mathrm{effective}}(n_k)}}{2\lambda_1 - \lambda_{\mathrm{effective}}(n_k)} \abs{\frac{n_k^i}{n_k^1}} \leq q \abs{\frac{n_k^i}{n_k^1}}
\end{equation*}
for some $q < 1$ depending only on the spectrum\footnote{For example, simple algebra shows that one can take $q^2 = 1 - \frac{4(\lambda_1 - \lambda_2) \lambda_m}{(2\lambda_1 - \lambda_m)^2}$.}.
For $n \neq 0$, define $\rho_1(n) \coloneqq \lambda_1(n^1)^2/(n^T Hn)$, the fraction of the quadratic energy in the top eigendirection. For a simple top eigenvalue, the component update and the ratio estimate above give
\begin{equation*}
	\rho_1(n_{k + 1}) \geq \rho_1(n_k),
\end{equation*}
with strict inequality whenever $\rho_1(n_k) \in (0, 1)$. This is the core mechanism that drives alignment of the iterates $n_k$ with the top eigendirection.
When $\rho_1(n_0) > 0$, this alignment is exponentially fast, so for fixed $n_0$, we have $\rho_1(n_k) \geq 1 - \delta$ after $\bigo{\log{(1 / \delta)}}$ steps.
In summary, the frozen-$p$ model in \cref{eq:frozen_hessian_quadratic_loss} preserves the normal distance while sharpening spectral alignment.

Let us now consider what happens when our model is still quadratic but we allow the Hessian to depend on $p$:
\begin{equation}
	\begin{aligned}
		\mathcal{L}(n, p) & = \frac{1}{2}n^T H(p)n, & H(p) & = \sum_{i = 1}^m \lambda_i(p)v_i(p)v_i^T(p).
	\end{aligned}
	\label{eq:moving_hessian_quadratic_loss}
\end{equation}
In this model, the manifold of minima $\mathcal{M}$ is still given by \cref{eq:quadratic_minimum_manifold}, but now there are nontrivial dynamics in the $p$ variables along $\mathcal{M}$.
The crucial point is that these dynamics are slow once $\theta_k$ is close to $\mathcal{M}$. Indeed,
\begin{equation*}
	\norm{p_{k + 1} - p_k}
	= \frac{4\mathcal{L}(n_k, p_k)}{\norm{\nabla \mathcal{L}(n_k, p_k)}^2}\norm{\nabla_p \mathcal{L}(n_k, p_k)}
	= \bigo{\norm{n_k}^2}.
\end{equation*}
This ensures that the top Hessian eigenvector $v_1(p_k)$ only changes $\bigo{\norm{n_k}^2}$ per iteration and that
\begin{equation*}
	n_{k + 1} = n_k - \frac{2n_k^T H(p_k)n_k}{\norm{H(p_k)n_k}^2}H(p_k)n_k + \bigo{\norm{n_k}^3}.
\end{equation*}
This ensures that the $n_k$ iterates will rapidly align with the slowly varying top Hessian eigendirection, $v_1(p_k)$, as in the first toy model we considered.
The precise entry estimate used in the general proof is in \Cref{prop:morse_bott_top_cone_entry}. Assuming that
\begin{equation*}
	\frac{n_k}{\norm{n_k}} = \pm v_1(p_k) + \bigo{\norm{n_k}}
\end{equation*}
we find
\begin{equation*}
	\frac{n_k^T H(p_k)n_k}{\norm{H(p_k)n_k}^2}
	= \frac{1}{\lambda_1(p_k)} + \bigo{\norm{n_k}^2} \Longrightarrow p_{k + 1}
	= p_k - \norm{n_k}^2\nabla_p \log{\lambda_1(p_k)} + \bigo{\norm{n_k}^3}.
\end{equation*}
Moreover,
\begin{equation*}
	\norm{n_{k + 1}}^2
	= \norm{n_k}^2 - \frac{4(n_k^T H(p_k)n_k)^2\norm{\nabla_p \mathcal{L}(n_k, p_k)}^2}{\left(\norm{H(p_k)n_k}^2 + \norm{\nabla_p \mathcal{L}(n_k, p_k)}^2\right)^2}
	= \norm{n_k}^2 + \bigo{\norm{n_k}^4}.
\end{equation*}
After alignment, sufficiently strong damping produces a two-step contraction of the normal distance while preserving the leading tangential drift.
The extension to general losses is proved in \Cref{subsec:morse_bott_cdp_convergence_proof}.
Further results and examples for the exact quadratic model, including eigenvalue switching, are given in \Cref{sec:quadratic_morse_bott_dynamics}, with proofs in \Cref{subsec:quadratic_morse_bott_proofs}.

\subsection*{Acknowledgments}

We thank Sanjeev Arora and Atish Agarwala for helpful discussions and feedback.
JMA was partially supported by a Sloan Research Fellowship and a Seed Grant Award from Apple.
BH was partially supported by a 2024 Sloan Fellowship in Mathematics, NSF CAREER grant DMS-2143754, and NSF grant DMS-2133806, and DARPA AIQ grant (HR001124S0029).

We used AI tools to assist with manuscript editing, code development, and checks of mathematical arguments. We take full responsibility for the correctness of the final content.

\clearpage
\nocite{*}
{\sloppy\printbibliography}
\clearpage

\hypersetup{colorlinks = true, linkcolor = black}
\begingroup
\setlength{\parskip}{0pt}
\raggedbottom
\tableofcontents
\markboth{\MakeUppercase{\shortauthors}}{\MakeUppercase{\shorttitle}}
\clearpage
\endgroup
\hypersetup{colorlinks = true, linkcolor = gray}

\appendix

\section{General Morse-Bott dynamics with CDP steps} \label{sec:morse_bott_dynamics}

In this appendix, we state a local convergence theorem and prove it together with \Cref{thm:localized_cdp_dynamics}.
Let us start by recalling the proposed CDP stepsize from \cref{eq:cdp_step_size}, which we denote by $\mathrm{CDP}_{\rm{exp}}$:
\begin{equation*}
	\eta_\kappa(\theta) = \frac{4\mathcal{L}(\theta)}{\norm{\nabla \mathcal{L}(\theta)}^2} e^{-\kappa\mathcal{L}(\theta)}.
\end{equation*}
For the computations, we use its rational approximation, the $\mathrm{CDP}_{\rm{rat}}$ update
\begin{equation} \label{eq:rational_cdp_update}
	\theta_{k + 1}
	= \theta_k - \beta_\kappa(\theta_k) \nabla \mathcal{L}(\theta_k), \text{ where } \beta_\kappa(\theta) \coloneqq \frac{4 \mathcal{L}(\theta)}{\norm{\nabla \mathcal{L}(\theta)}^2} \frac{1}{1 + \kappa \mathcal{L}(\theta)}.
\end{equation}
Whenever a point is written in tubular coordinates as $\theta = p + n$ near $\mathcal{M}$, we also write
\begin{equation*}
	\beta_\kappa(p, n) \coloneqq \beta_\kappa(\theta) = \beta_\kappa(p + n).
\end{equation*}
For fixed $\kappa$, both damping factors have the same first-order Taylor expansion, $1 - \kappa\mathcal{L} + \bigo{\mathcal{L}^2}$, as $\mathcal{L} \to 0$.
Thus damping is still a higher-order perturbation near the interpolation manifold, since we are assuming $\mathcal{L}(\theta) = \bigo{\dist{\theta}{\mathcal{M}}^2}$.
The damping replaces the exact period-two oscillation of the quadratic model by a controlled contraction every two steps.

The conclusions below also hold for $\mathrm{CDP}_{\rm{exp}}$ in \cref{eq:cdp_step_size}, with the same $\kappa$, by \Cref{lem:damping_comparison} (note that the admissible radii and constants in the estimates may change).

\subsection{Assumptions and main result}

We write $\operatorname{exp}_p$ for the Riemannian exponential map on $\mathcal{M}$ induced by the ambient Euclidean metric, and $d_{\mathcal{M}}$ for the induced intrinsic Riemannian distance.
All gradients on $\mathcal{M}$ are Riemannian gradients for the metric induced by the ambient Euclidean structure \cite[Chapter 2]{riemannian_lee_2018}.
Thus, for any smooth $h: \mathcal{M} \to \real$, $p \in \mathcal{M}$, $\nabla_{\mathcal{M}} h(p) \in T_p\mathcal{M}$ is defined by
\begin{equation*}
	\eval{\frac{\d}{\d t}}_{t = 0} h(\gamma(t))
	= \nabla_{\mathcal{M}} h(p)^T \gamma'(0), \text{ for every smooth curve } \gamma: (-\epsilon, \epsilon) \to \mathcal{M} \text{ with } \gamma(0)
	= p.
\end{equation*}
Or, if $\widetilde h$ is a smooth extension of $h$ to a neighborhood of $p$ in $\real^\tau$, then $\nabla_{\mathcal{M}} h(p) = \Pi_{T_p\mathcal{M}} \nabla \widetilde h(p)$.
\begin{assumption}[Morse-Bott structure] \label{ass:morse_bott_structure}
	Let $\tau \in \natural$, let $\mathcal{M} \subset \real^\tau$ be a connected $C^5$ submanifold of positive codimension.
	Assume that $\mathcal{L} \in C^5(\real^\tau, \real)$, $\mathcal{L} \geq 0$, $\mathcal{L}|_{\mathcal{M}} \equiv 0$, and $\nabla \mathcal{L}|_{\mathcal{M}} \equiv 0$.
	For each $p \in \mathcal{M}$, let
	\begin{equation*}
		H(p) \coloneqq \nabla^2 \mathcal{L}(p)|_{N_p\mathcal{M}}: N_p\mathcal{M} \to N_p\mathcal{M}.
	\end{equation*}
	We assume that $H(p)$ is positive definite for every $p \in \mathcal{M}$.
\end{assumption}
Let $m \coloneqq \codim{\mathcal{M}} \geq 1$.
For $p \in \mathcal{M}$, let $\lambda_1(p) \geq \cdots \geq \lambda_m(p) > 0$ denote the eigenvalues of $H(p)$, and let $V_1(p): N_p\mathcal{M} \to N_p\mathcal{M}$ be the orthogonal projector onto the eigenspace of $\lambda_1(p)$.
When $m = 1$, we use the notational convention $\lambda_2 \equiv 0$.
Whenever $\lambda_1(p)$ is simple and $v_1(p)$ is a unit top eigenvector, this is exactly $V_1(p) = v_1(p) v_1(p)^T$.
\begin{assumption}[Spectral conditions and localization on $\mathcal{K}$] \label{ass:localized_sharpness_spectrum}
	We fix a compact set $\mathcal{K} \subset \mathcal{M}$ to localize the projected iterates and make the estimates uniform. We assume:
	\begin{enumerate}
		\item the top eigenvalue is uniformly separated, i.e.,
			\begin{equation*}
				\delta \coloneqq \inf_{p \in \mathcal{K}}\bigl(\lambda_1(p) - \lambda_2(p)\bigr) > 0;
			\end{equation*}
		\item the function $p \mapsto \lambda_1(p)$ has a unique critical point $p_\star$ in $\mathcal{K}$, and $p_\star$ is a strict local minimum of $\lambda_1$ on $\mathcal{M}$;
		\item the trajectories $(\theta_k)_{k \geq 0}$ are localized over $\mathcal{K}$ in the following sense: for every $k$, whenever $\theta_k$ admits tubular coordinates, i.e., whenever $\theta_k = p_k + n_k$, with $n_k \in N_{p_k}\mathcal{M} \setminus \{0\}$, its base point satisfies $p_k \in \mathcal{K}$.
	\end{enumerate}
\end{assumption}
For $p \in \mathcal{M}$ and $n \in N_p\mathcal{M} \setminus \{0\}$, define the alignment factor
\begin{equation*}
	\alpha_p(n) \coloneqq \frac{\lambda_1(p) n^T V_1(p) n}{n^T H(p) n} \in [0, 1].
\end{equation*}
This quantity measures how much of the quadratic form $n^T H(p) n$ lies in the top eigendirection.
It equals $1$ on the top eigenspace and $0$ on the orthogonal complement.
\begin{theorem}[Convergence of CDP near a Morse-Bott manifold] \label{thm:morse_bott_cdp_convergence}
	Under \Cref{ass:morse_bott_structure,ass:localized_sharpness_spectrum}, there exist a damping threshold $\bar{\kappa} > 0$ and a tubular radius $\bar{r} > 0$ such that for every $\kappa \geq \bar{\kappa}$ and every minimum initial alignment $\nu \in (0, 1)$, there exists $r_0 = r_0(\kappa, \nu) \in (0, \bar{r})$ with the following property.
	Let $(\theta_k)_{k \geq 0}$ be a GD trajectory with parameter $\kappa$ and either $\mathrm{CDP}_{\rm{rat}}$ stepsizes as in \cref{eq:rational_cdp_update} or $\mathrm{CDP}_{\rm{exp}}$ stepsizes as in \cref{eq:cdp_step_size}, localized as in \Cref{ass:localized_sharpness_spectrum}.
	Assume that the initial condition has tubular coordinates $\theta_0 = p_0 + n_0$ with $p_0 \in \mathcal{K}$, $n_0 \in N_{p_0}\mathcal{M}$, and
	\begin{equation*}
		0 < \norm{n_0} \leq r_0, \text{ and } \alpha_{p_0}(n_0) \geq \nu.
	\end{equation*}
	Then every iterate lies in the fixed tubular neighborhood and has tubular coordinates $\theta_k = p_k + n_k$ with $p_k \in \mathcal{K}$ and $0 < \norm{n_k} < \bar{r}$, and
	\begin{equation*}
		n_k \xrightarrow{k \to \infty} 0, \text{ and } p_k \xrightarrow{k \to \infty} p_\star.
	\end{equation*}
	Moreover, if $v_1$ is a local unit top eigenvector field defined near $p_\star$, then for all sufficiently large $k$,
	\begin{equation*}
		\begin{gathered}
			\norm{n_{k + 2}}^2 = \norm{n_k}^2 - \Theta(\norm{n_k}^4), \ \dist{\frac{n_k}{\norm{n_k}}}{\{\pm v_1(p_k)\}} \to 0, \ \norm{n_k} \asymp \frac{1}{\sqrt{k}}, \\
			p_{k + 1} = \operatorname{exp}_{p_k}\bigl(-\norm{n_k}^2 \nabla_{\mathcal{M}} \log{\lambda_1}(p_k) + \epsilon_k\bigr), \ \epsilon_k \in T_{p_k}\mathcal{M}, \ \norm{\epsilon_k} = \bigo{\norm{n_k}^3}, \\
			d_{\mathcal{M}}(p_{k + 1}, p_k) = \bigo{\frac{1}{k}}.
		\end{gathered}
	\end{equation*}
\end{theorem}
\begin{proof}
	See \hyperlink{proof:morse_bott_cdp_convergence}{proof of \Cref{thm:morse_bott_cdp_convergence}} in \Cref{subsec:morse_bott_cdp_convergence_proof}.
\end{proof}
\begin{remark}
	Note that it is possible to obtain a polynomial convergence rate for the sharpness under an additional local Polyak-\L{}ojasiewicz condition on $\lambda_1$ along $\mathcal{M}$ (see \Cref{subsec:sharpness_convergence_rate}).
\end{remark}

\begin{figure}[htb]
	\centering
	\tikzexternaldisable
	\begin{tikzpicture}[
	x = 1cm, y = 1cm, line cap = round, line join = round,
	proof_step/.style = {draw = black!45, fill = black!2, rounded corners = 2pt, text width = 4.7cm, minimum height = 1.35cm, inner sep = 5pt, align = center, font = \footnotesize},
	proof_result/.style = {proof_step, draw = royalblue!55!black, fill = royalblue!5},
	proof_arrow/.style = {->, black!70, line width = 0.75pt, shorten >= 2pt, shorten <= 2pt}
]
	\hypersetup{linkcolor = black!75}
	\node[proof_step] (expansions) at (0, 0) {\textbf{Tubular expansions}\\\Cref{lem:morse_bott_tubular_expansions}};
	\node[proof_step] (alignment) at (5.6, 0) {\textbf{Alignment and entry}\\\Cref{lem:morse_bott_spectral_alignment}, \Cref{prop:morse_bott_top_cone_entry}};
	\node[proof_step] (cone_dynamics) at (11.2, 0) {\textbf{Contraction and drift}\\\Cref{prop:morse_bott_top_cone_dynamics}, \Cref{cor:morse_bott_normal_decay}};
	\node[proof_result] (localized_dynamics) at (0, -2.15) {\textbf{Localized dynamics}\\\Cref{thm:localized_cdp_dynamics}};
	\node[proof_result] (convergence) at (5.6, -2.15) {\textbf{Convergence to $p_\star$}\\\Cref{thm:morse_bott_cdp_convergence}};
	\node[proof_result, draw = forestgreen!55!black, fill = forestgreen!5] (sharpness_rate) at (11.2, -2.15) {\textbf{Sharpness rate (local PL)}\\\Cref{cor:morse_bott_sharpness_rate}};
	\draw[proof_arrow] (expansions.east) -- (alignment.west);
	\draw[proof_arrow] (alignment.east) -- (cone_dynamics.west);
	\draw[proof_arrow, rounded corners = 3pt] (cone_dynamics.south) -- (11.2, -1.075) -- (0, -1.075) -- (localized_dynamics.north);
	\draw[proof_arrow, shorten <= 0pt] (5.6, -1.075) -- (convergence.north);
	\draw[proof_arrow] (convergence.east) -- (sharpness_rate.west);
\end{tikzpicture}
	\tikzexternalenable
	\caption{
		Proof structure of \Cref{sec:morse_bott_dynamics}.
		In addition, \Cref{lem:damping_comparison} transfers estimates between damping rules.
	}
	\label{fig:appendix_a_structure}
\end{figure}
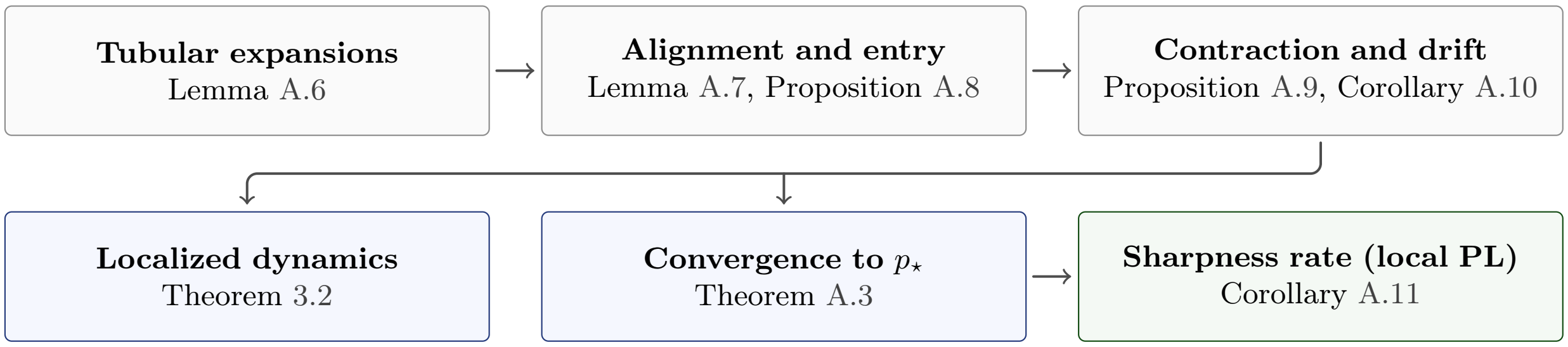

\subsection{Proof of the general Morse-Bott theorem} \label{subsec:morse_bott_cdp_convergence_proof}

Throughout the proof, \Cref{ass:morse_bott_structure,ass:localized_sharpness_spectrum} and the notation of \Cref{sec:morse_bott_dynamics} are applied.
All $\bigo{\cdot}$ constants are uniform for base points in $\mathcal{K}$ and unit normal directions over $\mathcal{K}$.
After a value of $\kappa$ is fixed, these constants may depend on $\kappa$, unless uniformity in $\kappa$ is explicitly stated.
The tubular radius $\bar{r}$ is fixed independently of $\kappa$. The threshold $\bar{\kappa}$ is chosen later.
For each fixed $\kappa \geq \bar{\kappa}$, the smaller radii used below may depend on $\kappa$.
Choose $\bar{r} > 0$ small enough that the tubular parametrization is unique on a fixed neighborhood of the zero section over an open neighborhood of $\mathcal{K} \subset \mathcal{M}$, for normal size less than $\bar{r}$.
By \Cref{ass:localized_sharpness_spectrum},
\begin{equation} \label{eq:morse_bott_spectral_gap}
	\delta \coloneqq \inf_{p \in \mathcal{K}} \bigl(\lambda_1(p) - \lambda_2(p)\bigr) > 0.
\end{equation}
By continuity and compactness, after passing to a fixed open neighborhood of $\mathcal{K} \subset \mathcal{M}$, the top eigenvalue remains simple there, say $\lambda_1 - \lambda_2 \geq \delta/2$.
All local top eigenvector fields and all $C^1$ bounds for $\lambda_1$ and $V_1$ below are taken in a finite trivialization family over this neighborhood.
For Taylor expansions, we use $p \in \mathcal{K}$, unit $u \in N_p\mathcal{M}$, and $0 \leq s < \bar{r}$, and write $\theta = p + su$.
For localized trajectories, whenever tubular coordinates are available, we write $s_k \coloneqq \norm{n_k} > 0$ and $u_k \coloneqq n_k / s_k$.

\subsubsection{Comparison of damping rules}

\begin{lemma}[Comparison of damping rules] \label{lem:damping_comparison}
	Fix $\kappa \geq 0$ and define the $\mathrm{CDP}_{\rm{exp}}$ and $\mathrm{CDP}_{\rm{rat}}$ updates by
	\begin{equation*}
		\begin{aligned}
			T_{\mathrm{exp}}(\theta) & \coloneqq \theta - \frac{4\mathcal{L}(\theta)}{\norm{\nabla \mathcal{L}(\theta)}^2} e^{-\kappa \mathcal{L}(\theta)} \nabla \mathcal{L}(\theta), \\
			T_{\mathrm{rat}}(\theta) & \coloneqq \theta - \frac{4\mathcal{L}(\theta)}{\norm{\nabla \mathcal{L}(\theta)}^2} \frac{1}{1 + \kappa \mathcal{L}(\theta)} \nabla \mathcal{L}(\theta).
		\end{aligned}
	\end{equation*}
	Uniformly for $\theta = p + su$, with $p \in \mathcal{K}$, unit $u \in N_p\mathcal{M}$, and sufficiently small $s > 0$,
	\begin{equation} \label{eq:damping_map_difference}
		\norm{T_{\mathrm{exp}}(\theta) - T_{\mathrm{rat}}(\theta)} = \bigo{s^5}.
	\end{equation}
	The same bound holds after two steps.
	After either one or two updates from the same point, write the two images as $p_{\mathrm{exp}} + n_{\mathrm{exp}}$ and $p_{\mathrm{rat}} + n_{\mathrm{rat}}$ in tubular coordinates. Then
	\begin{equation*}
		\begin{gathered}
			\norm{p_{\mathrm{exp}} - p_{\mathrm{rat}}} + \norm{n_{\mathrm{exp}} - n_{\mathrm{rat}}} = \bigo{s^5}, \\
			\abs{\norm{n_{\mathrm{exp}}}^2 - \norm{n_{\mathrm{rat}}}^2} = \bigo{s^6}, \text{ and } \norm{\frac{n_{\mathrm{exp}}}{\norm{n_{\mathrm{exp}}}} - \frac{n_{\mathrm{rat}}}{\norm{n_{\mathrm{rat}}}}} = \bigo{s^4}.
		\end{gathered}
	\end{equation*}
\end{lemma}
\begin{proof}
	The Morse-Bott assumptions in \Cref{ass:morse_bott_structure}, together with compactness of $\mathcal{K}$ from \Cref{ass:localized_sharpness_spectrum}, give $\mathcal{L} = \bigo{s^2}$ and $\norm{\nabla \mathcal{L}} \asymp s$.

	By a Taylor expansion, for fixed $\kappa$, the damping factors differ by
	\begin{equation*}
		e^{-\kappa \mathcal{L}} - \frac{1}{1 + \kappa \mathcal{L}} = -\frac{1}{2}\kappa^2 \mathcal{L}^2 + \bigo{\mathcal{L}^3} = \bigo{s^4}.
	\end{equation*}
	Multiplying by the undamped displacement gives
	\begin{equation*}
		\norm{T_{\mathrm{exp}}(\theta) - T_{\mathrm{rat}}(\theta)} = \abs{e^{-\kappa \mathcal{L}} - \frac{1}{1 + \kappa \mathcal{L}}} \frac{4\mathcal{L}}{\norm{\nabla \mathcal{L}}} = \bigo{s^4}\bigo{s} = \bigo{s^5},
	\end{equation*}
	which proves \cref{eq:damping_map_difference}.
	The frozen normal map preserves length, so both images have normal length $s + \bigo{s^2}$.
	Since the distance function to $\mathcal{M}$ is nonexpansive ($1$-Lipschitz), their $\bigo{s^5}$ separation keeps the joining segment at distance comparable to $s$ from $\mathcal{M}$.
	There, the updates have bounded derivatives: the stepsizes are $\bigo{1}$, their derivatives are $\bigo{1/s}$, the gradient is $\bigo{s}$, and the Hessian is bounded.
	These estimates hold near $\mathcal{K}$, so, writing $T^2 = T \circ T$, a second update gives
	\begin{equation*}
		\norm{T_{\mathrm{exp}}^2(\theta) - T_{\mathrm{rat}}^2(\theta)} \leq C\norm{T_{\mathrm{exp}}(\theta) - T_{\mathrm{rat}}(\theta)} + \norm{(T_{\mathrm{exp}} - T_{\mathrm{rat}})(T_{\mathrm{rat}}(\theta))} = \bigo{s^5}.
	\end{equation*}
	The tubular coordinate map is locally Lipschitz, giving the same bound for the base points and normal vectors.
	Since the normal lengths are comparable to $s$, squaring multiplies the error by $\bigo{s}$ and normalization divides it by a quantity comparable to $s$, giving the two remaining bounds.
\end{proof}
We compute below for $\mathrm{CDP}_{\rm{rat}}$.
By \Cref{lem:damping_comparison}, changing to $\mathrm{CDP}_{\rm{exp}}$ adds errors of order $s^4$ in the unit direction and $s^5$ in the base point, smaller than the $\bigo{s}$ and $\bigo{s^3}$ remainders used below.
The $\bigo{s^6}$ error in the two-step squared normal length likewise preserves its decrease of order $s^4$ for sufficiently small $s$.
Thus the same proof gives \Cref{thm:morse_bott_cdp_convergence} for both rules with the same $\kappa \geq \bar{\kappa}$, possibly smaller radii, and different constants.

\subsubsection{Tubular expansions}

\begin{lemma}[Tubular expansions] \label{lem:morse_bott_tubular_expansions}
	With the notation fixed above, the following expansions hold uniformly on the compact unit normal bundle over $\mathcal{K}$:
	\begin{align}
		\mathcal{L}(p + s u) & = \frac{1}{2} s^2 u^T H(p) u + s^3 R_{\mathcal{L}, 3}(p, u) + s^4 R_{\mathcal{L}, 4}(p, u) + \bigo{s^5}, \label{eq:morse_bott_loss_expansion} \\
		\Pi_{N_p\mathcal{M}} \nabla \mathcal{L}(p + s u) & = s H(p) u + s^2 R_{\nabla_n, 2}(p, u) + s^3 R_{\nabla_n, 3}(p, u) + \bigo{s^4}, \label{eq:morse_bott_normal_gradient_expansion} \\
		\Pi_{T_p\mathcal{M}} \nabla \mathcal{L}(p + s u) & = \frac{1}{2} s^2 \nabla_{\mathcal{M}} \bigl(u^T H(p) u\bigr) + \bigo{s^3}. \label{eq:morse_bott_tangential_gradient_expansion}
	\end{align}
	Here $\nabla_{\mathcal{M}} \bigl(u^T H(p) u\bigr) \in T_p\mathcal{M}$ is the vector characterized by
	\begin{equation*}
		\nabla_{\mathcal{M}} \bigl(u^T H(p) u\bigr)^T w
		= \eval{\frac{\d}{\d t}}_{t = 0} u(t)^T H(\gamma(t)) u(t),
	\end{equation*}
	for every $w \in T_p\mathcal{M}$, where $\gamma$ is a smooth curve in $\mathcal{M}$ with $\gamma(0) = p$ and $\gamma'(0) = w$, and $u(t)$ is the normal-parallel transport of $u$ along $\gamma$.
	In particular, after shrinking $\bar{r}$ if necessary, there is $c_{\nabla} > 0$ such that
	\begin{equation} \label{eq:morse_bott_gradient_lower_bound}
		\norm{\nabla \mathcal{L}(p + s u)} \geq c_{\nabla} s, \ \forall p \in \mathcal{K}, u \in N_p\mathcal{M}, \norm{u}
		= 1, 0 < s < \bar{r}.
	\end{equation}
	Now fix $\kappa \geq 0$.
	There exists $r_\kappa \in (0, \bar{r})$ such that, whenever $0 < s < r_\kappa$, the one-step image of $\theta$ has tubular coordinates
	\begin{equation*}
		\theta^+ = p^+ + n^+, \ n^+ \in N_{p^+}\mathcal{M}, \ 0 < \norm{n^+} < \bar{r}.
	\end{equation*}
	Define
	\begin{equation*}
		Q_p(u) \coloneqq u - 2 \frac{u^T H(p) u}{u^T H(p)^2 u} H(p) u.
	\end{equation*}
	With $r_\kappa$ possibly made smaller,
	\begin{align}
		\beta_\kappa(p, s u) & = 2 \frac{u^T H(p) u}{u^T H(p)^2 u} + \bigo{s}, \label{eq:damped_polyak_step_expansion}\\
		n^+ & = s Q_p(u) + s^2 R_{n, 2}(p, u) + s^3 \bigl(R^{(1)}_{n, 3}(p, u) + \kappa R^{(2)}_{n, 3}(p, u)\bigr) + \bigo{s^4}, \label{eq:morse_bott_normal_update_expansion}\\
		p^+ & = \operatorname{exp}_p\left(-s^2 \frac{u^T H(p) u}{u^T H(p)^2 u} \nabla_{\mathcal{M}} \bigl(u^T H(p) u\bigr) + \bigo{s^3}\right), \label{eq:morse_bott_tangential_update_expansion}
	\end{align}
	where $R_{n, 2}$ is $C^2$ and $R^{(1)}_{n, 3}$ and $R^{(2)}_{n, 3}$ are $C^1$, all independent of $\kappa$.
	The error term in \cref{eq:morse_bott_tangential_update_expansion} is tangent-valued.
	The explicit coefficient of the $\kappa$-term is
	\begin{equation} \label{eq:morse_bott_normal_damping_coefficient}
		R^{(2)}_{n, 3}(p, u) = \frac{(u^T H(p) u)^2}{u^T H(p)^2 u} H(p) u.
	\end{equation}
	Moreover, a direct computation gives $\norm{Q_p(u)} = 1$.
	Consequently, if $n^+ = s^+ u^+$ with $s^+ > 0$ and $\norm{u^+} = 1$, then
	\begin{equation} \label{eq:morse_bott_radius_direction_update}
		s^+ = s + \bigo{s^2}, \ u^+ = Q_p(u) + \bigo{s}.
	\end{equation}
	Moreover, taking norms in \cref{eq:morse_bott_normal_update_expansion} gives a $C^2$ coefficient field $\rho^{(1)}$ and $C^1$ coefficient fields $\rho^{(2)}$ and $\rho^{(3)}$, independent of $\kappa$, such that
	\begin{equation} \label{eq:morse_bott_normal_radius_expansion}
		s^+
		= s + s^2 \rho^{(1)}(p, u) + s^3 \bigl(\rho^{(2)}(p, u) + \kappa \rho^{(3)}(p, u)\bigr) + \bigo{s^4}.
	\end{equation}
	In particular,
	\begin{equation} \label{eq:morse_bott_radius_damping_coefficient}
		\rho^{(3)}(p, u) = Q_p(u)^T R^{(2)}_{n, 3}(p, u).
	\end{equation}
\end{lemma}
\begin{proof}
	Choose a finite cover of $\mathcal{K}$ by open sets in $\mathcal{M}$ on which the normal bundle has $C^4$ orthonormal normal frames $\nu_1, \ldots, \nu_m$.
	In one such trivialization, write
	\begin{equation*}
		n
		= \sum_{i = 1}^m n_i \nu_i(p), \ \widetilde{\mathcal{L}}(p, n) \coloneqq \mathcal{L}(p + n)
		= \mathcal{L}\left(p + \sum_{i = 1}^m n_i \nu_i(p)\right),
	\end{equation*}
	and set $H_{ij}(p) \coloneqq \nu_i(p)^T H(p)\nu_j(p)$.
	The Morse-Bott assumptions imply
	\begin{equation*}
		\widetilde{\mathcal{L}}(p, 0)
		= 0, \ \nabla_n \widetilde{\mathcal{L}}(p, 0)
		= 0, \ \text{and } \nabla_p \widetilde{\mathcal{L}}(p, 0)
		= 0.
	\end{equation*}
	The required derivatives are uniformly controlled.
	Taylor expansion is taken in the Cartesian normal variable:
	\begin{equation*}
		\widetilde{\mathcal{L}}(p, n)
		= \frac{1}{2} n^T H(p) n + \frac{1}{6} D_n^3 \widetilde{\mathcal{L}}(p, 0)[n,n,n] + \frac{1}{24} D_n^4 \widetilde{\mathcal{L}}(p, 0)[n,n,n,n] + \bigo{\norm{n}^5}.
	\end{equation*}
	Here $D_n^\ell \widetilde{\mathcal{L}}(p, 0)$ denotes the $\ell$-th derivative with respect to the normal variable, with $p$ fixed, viewed as an $\ell$-linear form on $N_p\mathcal{M}$.
	Setting $n = su$ gives $C^1$ coefficient fields $R_{\mathcal{L}, 3}$ and $R_{\mathcal{L}, 4}$ on the compact unit normal bundle over $\mathcal{K}$ such that
	\begin{equation*}
		\widetilde{\mathcal{L}}(p, s u)
		= \frac{1}{2} s^2 u^T H(p)u + s^3 R_{\mathcal{L}, 3}(p, u) + s^4 R_{\mathcal{L}, 4}(p, u) + \bigo{s^5}.
	\end{equation*}
	This is \cref{eq:morse_bott_loss_expansion}. Differentiating the Cartesian expansion in $n$ and then setting $n = su$ gives \cref{eq:morse_bott_normal_gradient_expansion}.
	We now identify the tangential coefficient.
	Let $w \in T_p\mathcal{M}$, let $u \in N_p\mathcal{M}$ with $\norm{u} = 1$, let $\gamma(t)$ be a curve in $\mathcal{M}$ with $\gamma(0) = p$ and $\gamma'(0) = w$, and let $u(t)$ be parallel in the normal bundle along $\gamma(t)$ with $u(0) = u$.
	Define also $v(t)$ as the normal vector field along $\gamma(t)$ obtained by freezing the coordinates of $u$ in the moving frame $\left(\nu_1(\gamma(t)), \dots, \nu_m(\gamma(t))\right)$.
	Then
	\begin{equation*}
		\eval{\frac{\d}{\d t}}_{t = 0} u(t)^T H(\gamma(t)) u(t)
		= \eval{\frac{\d}{\d t}}_{t = 0} v(t)^T H(\gamma(t)) v(t) + 2 (H(p) u)^T \eval{\frac{\d}{\d t}}_{t = 0} \bigl(u(t) - v(t)\bigr).
	\end{equation*}
	On the other hand, differentiating $\mathcal{L}\left(\gamma(t) + s v(t)\right)$ gives
	\begin{equation*}
		\eval{\frac{\d}{\d t}}_{t = 0} \mathcal{L}\left(\gamma(t) + s v(t)\right)
		= \frac{1}{2} s^2 \eval{\frac{\d}{\d t}}_{t = 0} v(t)^T H(\gamma(t)) v(t) + \bigo{s^3}.
	\end{equation*}
	Since
	\begin{equation*}
		\eval{\frac{\d}{\d t}}_{t = 0} \mathcal{L}\left(\gamma(t) + s v(t)\right)
		= \nabla \mathcal{L}(p + s u)^T \left(w + s v'(0)\right),
	\end{equation*}
	and the leading normal gradient is $s H(p) u + \bigo{s^2}$, we get
	\begin{equation*}
		w^T \Pi_{T_p\mathcal{M}} \nabla \mathcal{L}(p + s u)
		= \frac{1}{2} s^2 \eval{\frac{\d}{\d t}}_{t = 0} v(t)^T H(\gamma(t)) v(t) - s^2 (H(p) u)^T \Pi_{N_p\mathcal{M}} v'(0) + \bigo{s^3}.
	\end{equation*}
	Since $u(t)$ is parallel in the normal bundle,
	\begin{equation*}
		\Pi_{N_p\mathcal{M}} u'(0) = 0, \text{ so }
		(H(p) u)^T \eval{\frac{\d}{\d t}}_{t = 0} \bigl(u(t) - v(t)\bigr)
		= - (H(p) u)^T \Pi_{N_p\mathcal{M}} v'(0).
	\end{equation*}
	Using this and the identity above for $\eval{\frac{\d}{\d t}}_{t = 0} u(t)^T H(\gamma(t)) u(t)$, we obtain
	\begin{equation*}
		w^T \Pi_{T_p\mathcal{M}} \nabla \mathcal{L}(p + s u)
		= \frac{1}{2} s^2 \eval{\frac{\d}{\d t}}_{t = 0} u(t)^T H(\gamma(t)) u(t) + \bigo{s^3}.
	\end{equation*}
	This proves \cref{eq:morse_bott_tangential_gradient_expansion}.
	Let
	\begin{equation*}
		\lambda_{\min} \coloneqq \inf_{p \in \mathcal{K}} \lambda_m(p) > 0.
	\end{equation*}
	By uniformity in \cref{eq:morse_bott_normal_gradient_expansion}, after shrinking $\bar{r}$ if necessary there is $c > 0$ such that
	\begin{equation*}
		\norm{\Pi_{N_p\mathcal{M}} \nabla \mathcal{L}(p + s u) - s H(p) u} \leq s^2 c
	\end{equation*}
	for every $p \in \mathcal{K}$, every unit $u \in N_p\mathcal{M}$, and every $0 < s < \bar{r}$.
	Since $\norm{H(p) u} \geq \lambda_m(p) \geq \lambda_{\min}$,
	\begin{equation*}
		\norm{\Pi_{N_p\mathcal{M}} \nabla \mathcal{L}(p + s u)} \geq s \lambda_{\min} - s^2 c.
	\end{equation*}
	After shrinking $\bar{r}$ once more so that $c s \leq \lambda_{\min}/2$ for $0 < s < \bar{r}$, we get
	\begin{equation*}
		\norm{\Pi_{N_p\mathcal{M}} \nabla \mathcal{L}(p + s u)} \geq \frac{\lambda_{\min}}{2} s.
	\end{equation*}
	Because $T_p\mathcal{M}$ and $N_p\mathcal{M}$ are orthogonal complements in the ambient Euclidean space,
	\begin{equation*}
		\norm{\nabla \mathcal{L}(p + s u)} \geq \norm{\Pi_{N_p\mathcal{M}} \nabla \mathcal{L}(p + s u)}.
	\end{equation*}
	This gives \cref{eq:morse_bott_gradient_lower_bound}.
	Fix $\kappa$ and choose $r_\kappa \in (0, \bar{r})$ small enough for the estimates below.
	The expansions of $\mathcal{L}$ and $\nabla \mathcal{L}$ give
	\begin{equation*}
		\frac{4 \mathcal{L}(p + s u)}{\norm{\nabla \mathcal{L}(p + s u)}^2}
		= 2 \frac{u^T H(p) u}{u^T H(p)^2 u} + \bigo{s}, \text{ and }
		\frac{1}{1 + \kappa \mathcal{L}(p + s u)} = 1 - \frac{1}{2} \kappa s^2 u^T H(p) u + \bigo{s^3}.
	\end{equation*}
	This proves \cref{eq:damped_polyak_step_expansion}.
	The $\kappa$-dependent part of the stepsize at order $s^2$ is
	\begin{equation*}
		-\kappa s^2 \frac{(u^T H(p) u)^2}{u^T H(p)^2 u} + \bigo{s^3}.
	\end{equation*}
	Since the frozen normal update is $s u - \beta_\kappa(p, s u)\Pi_{N_p\mathcal{M}}\nabla \mathcal{L}(p + s u)$ and the leading normal gradient is $s H(p) u$, this produces the order-$\kappa s^3$ normal contribution
	\begin{equation*}
		\kappa s^3 \frac{(u^T H(p) u)^2}{u^T H(p)^2 u} H(p) u,
	\end{equation*}
	which is exactly \cref{eq:morse_bott_normal_damping_coefficient}.
	In the frozen splitting $T_p\mathcal{M} \oplus N_p\mathcal{M}$ at the base point $p$, the post-step point satisfies
	\begin{align*}
		\Pi_{T_p\mathcal{M}} (\theta^+ - p)
		& = -s^2 \frac{u^T H(p) u}{u^T H(p)^2 u} \nabla_{\mathcal{M}} \bigl(u^T H(p) u\bigr) + \bigo{s^3}, \\
		\Pi_{N_p\mathcal{M}} (\theta^+ - p)
		& = s Q_p(u) + s^2 R_{n, 2}(p, u) + s^3 \bigl(R^{(1)}_{n, 3}(p, u) + \kappa R^{(2)}_{n, 3}(p, u)\bigr) + \bigo{s^4},
	\end{align*}
	for some $C^1$ coefficient fields $R_{n, 2}$ and $R^{(1)}_{n, 3}$, independent of $\kappa$.
	The frozen estimates show that $\theta^+$ is at distance at most $c_\kappa s$ from $p$, and that its frozen normal component has norm $s + \bigo{s^2}$, because $\norm{Q_p(u)} = 1$.
	After reducing $r_\kappa$ if needed, $\theta^+$ therefore lies in the fixed tubular neighborhood and its normal coordinate is nonzero and smaller than $\bar{r}$.
	Passing from the frozen splitting at $p$ to genuine tubular coordinates at the nearby base point $p^+$ changes the base component by $\bigo{s^3}$ and the normal component by another $C^1$ order-$s^3$ term.
	Indeed, in local coordinates for the tubular map, the derivative of the inverse at $(p,0)$ is the frozen splitting.
	Here the frozen tangent displacement is $\bigo{s^2}$ and the frozen normal displacement is $\bigo{s}$, so the mixed tangent-normal error is $\bigo{s^3}$.
	At this order the coordinate change depends only on the leading frozen normal component and the leading tangent displacement, so it is absorbed into the $\kappa$-independent field $R^{(1)}_{n, 3}$.
	This gives \cref{eq:morse_bott_normal_update_expansion,eq:morse_bott_tangential_update_expansion}.
	Since the leading term in \cref{eq:morse_bott_normal_update_expansion} has norm $s$, \cref{eq:morse_bott_radius_direction_update,eq:morse_bott_normal_radius_expansion} follow by taking norms and normalizing, the $\kappa$-coefficient in the norm is $Q_p(u)^T R^{(2)}_{n,3}(p,u)$, giving \cref{eq:morse_bott_radius_damping_coefficient}.
	The stated coefficient regularity follows from the $C^5$ assumptions and the $C^4$ regularity of the tubular projection.
\end{proof}

\subsubsection{Alignment}

\begin{lemma}[Alignment] \label{lem:morse_bott_spectral_alignment}
	With the notation fixed above, the frozen map $Q_p$ satisfies
	\begin{equation*}
		\norm{Q_p(u)} = 1 \text{ and } \alpha_p(Q_p(u)) \geq \alpha_p(u).
	\end{equation*}
	The inequality is strict whenever $0 < \alpha_p(u) < 1$.
	Consequently, for every $\nu \in (0, \frac{1}{2})$ there exists $c_\nu^{(1)} > 0$ such that
	\begin{equation*}
		\nu \leq \alpha_p(u) \leq 1 - \nu \implies \alpha_p(Q_p(u)) \geq \alpha_p(u) + c_\nu^{(1)}, \text{ uniformly in } p, u.
	\end{equation*}
	For each fixed $\kappa \geq 0$, after shrinking the radius, there exists $c_\nu^{(2)} > 0$ such that the true one-step dynamics satisfy
	\begin{equation} \label{eq:morse_bott_alignment_gain}
		\nu \leq \alpha_p(u) \leq 1 - \nu \implies \alpha_{p^+}(u^+) \geq \alpha_p(u) + c_\nu^{(1)} - c_\nu^{(2)} s.
	\end{equation}
	Moreover, there exist constants $0 < c_1 \leq c_2 < \infty$ such that
	\begin{equation} \label{eq:alignment_transverse_component_bounds}
		c_1 \norm{(I - V_1(p)) u}^2 \leq 1 - \alpha_p(u) \leq c_2 \norm{(I - V_1(p)) u}^2, \text{ uniformly in } p , u.
	\end{equation}
	Finally, there exist $c_3 \in (0, 1)$, $c_4 \in (0, 1)$, and $c_5 > 0$, chosen independently of $\kappa$, such that for each fixed $\kappa$ the radius can be shrunk so that
	\begin{equation} \label{eq:morse_bott_transverse_contraction}
		1 - \alpha_p(u) \leq c_3 \implies \norm{(I - V_1(p^+))u^+} \leq c_4 \norm{(I - V_1(p)) u} + c_5 s.
	\end{equation}
\end{lemma}
\begin{proof}
	Fix $p$ and write $u = \sum_{i = 1}^m u_i v_i(p)$ in an orthonormal eigenbasis of $H(p)$.
	Then
	\begin{equation*}
		Q_p(u)
		= \sum_{i = 1}^m q_i u_i v_i(p), \ q_i \coloneqq 1 - 2 \frac{\lambda_i(p) u^T H(p) u}{u^T H(p)^2 u}.
	\end{equation*}
	The identity $\norm{Q_p(u)} = 1$ follows from
	\begin{equation*}
		\sum_{i = 1}^m q_i^2 u_i^2
		= 1 - 4 \frac{(u^T H(p) u)^2}{u^T H(p)^2 u} + 4 \frac{(u^T H(p) u)^2}{u^T H(p)^2 u}
		= 1.
	\end{equation*}
	Since
	\begin{equation*}
		\lambda_m(p) \leq \frac{u^T H(p)^2 u}{u^T H(p) u} \leq \lambda_1(p) \Rightarrow
		\abs{q_i}
		= \abs{1 - 2 \frac{\lambda_i(p) u^T H(p) u}{u^T H(p)^2 u}} \leq 2 \frac{\lambda_1(p) u^T H(p) u}{u^T H(p)^2 u} - 1
		= \abs{q_1}, \ \forall i \geq 2.
	\end{equation*}
	Therefore
	\begin{equation*}
		\alpha_p(Q_p(u))
		= \frac{\lambda_1(p) q_1^2 u_1^2}{\sum_{i = 1}^m \lambda_i(p) q_i^2 u_i^2} \geq \frac{\lambda_1(p) u_1^2}{\sum_{i = 1}^m \lambda_i(p) u_i^2}
		= \alpha_p(u).
	\end{equation*}
	The inequality is strict if both the top and non-top components are nonzero.
	Compactness of the set $\{(p, u): p \in \mathcal{K}, \norm{u} = 1, \nu \leq \alpha_p(u) \leq 1 - \nu\}$ gives the uniform gain.
	For the true dynamics, \cref{eq:morse_bott_tangential_update_expansion,eq:morse_bott_radius_direction_update} give $d_{\mathcal{M}}(p^+, p) = \bigo{s^2}$ and $u^+ = Q_p(u) + \bigo{s}$.
	The function $(p, u)\mapsto\alpha_p(u)$ has uniform $C^1$ bounds in the finite simple-eigenvalue trivialization family around $\mathcal{K}$, so \cref{eq:morse_bott_alignment_gain} follows.
	The equivalence \cref{eq:alignment_transverse_component_bounds} follows from
	\begin{equation*}
		\norm{(I - V_1(p)) u}^2
		= \sum_{i = 2}^m u_i^2, \ 1 - \alpha_p(u)
		= \frac{\sum_{i = 2}^m \lambda_i(p) u_i^2}{\sum_{i = 1}^m \lambda_i(p) u_i^2}
	\end{equation*}
	and the uniform upper and lower eigenvalue bounds on $\mathcal{K}$.
	It remains to prove \cref{eq:morse_bott_transverse_contraction}.
	If $m = 1$, then $V_1(p) = I$ and the claim is trivial.
	Assume $m \geq 2$.
	Define $q_{\max}(p, u) \coloneqq \max_{2 \leq i \leq m} \abs{q_i}$.
	The map $(p, u) \mapsto q_{\max}(p, u)$ is continuous on the compact unit normal bundle over $\mathcal{K}$.
	On the top line,
	\begin{equation*}
		q_{\max}(p, \pm v_1(p)) = \max_{2 \leq i \leq m} \abs{q_i}.
	\end{equation*}
	By the definition of $q_i$, the spectral gap \cref{eq:morse_bott_spectral_gap}, and the uniform positive lower bound on $\lambda_m$, this quantity is strictly less than $1$ for every $p \in \mathcal{K}$.
	Since it depends continuously on $p$ and $\mathcal{K}$ is compact, the bound is uniform.
	Using \cref{eq:alignment_transverse_component_bounds}, choose $c_3 \in (0, 1)$ and $c_4 \in (0, 1)$ such that
	\begin{equation*}
		1 - \alpha_p(u) \leq c_3 \implies q_{\max}(p, u) \leq c_4.
	\end{equation*}
	Then
	\begin{equation*}
		\norm{(I - V_1(p))Q_p(u)} \leq c_4 \norm{(I - V_1(p)) u}.
	\end{equation*}
	The true update satisfies
	\begin{equation*}
		u^+ = Q_p(u) + \bigo{s} + R_{u_\kappa, 2}(p, u, s),
	\end{equation*}
	where $\norm{R_{u_\kappa, 2}(p, u, s)} \leq s^2 c_\kappa$ for fixed $\kappa$, and also $d_{\mathcal{M}}(p^+, p) = \bigo{s^2}$.
	After fixing $\kappa$ and shrinking the radius, the remainder $R_{u_\kappa, 2}$ is absorbed into $c_5 s$.
	The $C^1$ control of $V_1$ in the finite trivialization family gives \cref{eq:morse_bott_transverse_contraction}, with $c_3$, $c_4$, and $c_5$ independent of $\kappa$.
\end{proof}
Fix from now on a cone slope $L \geq 1$ satisfying
\begin{equation} \label{eq:top_cone_slope}
	L > \frac{8 c_5}{1 - c_4}.
\end{equation}

\subsubsection{Entry into the top cone}

\begin{proposition}[Entry into the top cone] \label{prop:morse_bott_top_cone_entry}
	For every fixed $\kappa \geq 0$, every $\nu \in (0, 1)$, and every target radius $r \in (0, \bar{r})$, there exist $c_{\mathrm{entry}} > 0$, $N_{\mathrm{entry}} \in \natural$, and $r_0 \in (0, r/2)$ with the following property.
	Let $(\theta_k)_{k \geq 0}$ be a localized trajectory as defined in \Cref{ass:localized_sharpness_spectrum}.
	Assume that $\theta_0$ has tubular coordinates $\theta_0 = p_0 + n_0$ with $p_0 \in \mathcal{K}$ and
	\begin{equation*}
		0 < \norm{n_0} \leq r_0, \ u_0 \coloneqq \frac{n_0}{\norm{n_0}}, \ \alpha_{p_0}(u_0) \geq \nu.
	\end{equation*}
	Set
	\begin{equation*}
		k^{\max}_{\mathrm{entry}} \coloneqq N_{\mathrm{entry}} + \ceil{c_{\mathrm{entry}} \log{\frac{1}{\norm{n_0}}}}.
	\end{equation*}
	Then there is $k_0 \leq k^{\max}_{\mathrm{entry}}$ such that all iterates $\theta_k$, $0 \leq k \leq k_0$, have tubular coordinates $\theta_k = p_k + n_k$ with $p_k \in \mathcal{K}$ and $0 < \norm{n_k} < \bar{r}$, and
	\begin{equation} \label{eq:top_cone_entry_bounds}
		0 < \norm{n_{k_0}} \leq r, \ \norm{(I - V_1(p_{k_0}))n_{k_0}} \leq L \norm{n_{k_0}}^2.
	\end{equation}
	Moreover, on the prefix,
	\begin{equation*}
		\frac{1}{2} \norm{n_0} \leq \norm{n_k} \leq 2 \norm{n_0}, \text{ for } 0 \leq k \leq k_0.
	\end{equation*}
\end{proposition}
\begin{proof}
	After fixing $\kappa$, choose $r_{\mathrm{entry}} \in (0, \bar{r})$ small enough that \Cref{lem:morse_bott_tubular_expansions,lem:morse_bott_spectral_alignment} apply for all $0 < s \leq r_{\mathrm{entry}}$.
	In particular, there exists $c_6 > 0$ such that if an iterate has tubular coordinates $\theta_k = p_k + n_k$ with $p_k \in \mathcal{K}$ and $0 < s_k \leq r_{\mathrm{entry}}$, then the next iterate has tubular coordinates and, by localization, its base point lies in $\mathcal{K}$, while
	\begin{equation} \label{eq:entry_normal_radius_increment}
		\abs{s_{k + 1} - s_k} \leq c_6 s_k^2.
	\end{equation}
	Choose $c_{\mathrm{entry}} > 0$ so large that, for all sufficiently small $r_1 > 0$,
	\begin{equation} \label{eq:entry_alignment_time_bound}
		c_4^{\ceil{c_{\mathrm{entry}} \log{\frac{1}{r_1}}}} \leq \frac{L}{4} r_1.
	\end{equation}
	Let
	\begin{equation*}
		\nu_0 \coloneqq \min\left\{\nu, \frac{c_1 c_3}{4 c_2}, \frac{1}{4}\right\}.
	\end{equation*}
	Let $c_{\nu_0}^{(1)}$ and $c_{\nu_0}^{(2)}$ be the corresponding constants from \Cref{lem:morse_bott_spectral_alignment}, and let
	\begin{equation*}
		N_{\mathrm{entry}} \coloneqq \ceil{\frac{2(1 - 2 \nu_0)}{c_{\nu_0}^{(1)}}}.
	\end{equation*}
	Choose $r_0 \in (0, \min\{r/2, r_{\mathrm{entry}}/4\})$ small enough that all fixed-$\kappa$ estimates used below apply and, for every $0 < r_1 \leq r_0$,
	\begin{equation} \label{eq:entry_radius_conditions}
		4 c_6 \left(N_{\mathrm{entry}} + \ceil{c_{\mathrm{entry}} \log{\frac{1}{r_1}}}\right) r_1 \leq \frac{1}{2}, \
		2 c_{\nu_0}^{(2)} r_1 \leq \frac{c_{\nu_0}^{(1)}}{2}, \
		2 c_5 r_1 \leq \frac{1 - c_4}{2} \sqrt{c_3/c_2}.
	\end{equation}
	Let
	\begin{equation*}
		k^{\max}_{\mathrm{entry}} \coloneqq N_{\mathrm{entry}} + \ceil{c_{\mathrm{entry}} \log{\frac{1}{s_0}}}.
	\end{equation*}
	We first prove, simultaneously, that the iterates have tubular coordinates and that $s_k$ stays comparable to $s_0$ for $0 \leq k \leq k^{\max}_{\mathrm{entry}}$.
	Assume inductively that the tubular coordinates are defined with base in $\mathcal{K}$ and $s_i \leq 2 s_0$ for $0 \leq i < k \leq k^{\max}_{\mathrm{entry}}$.
	Since $2s_0 \leq r_{\mathrm{entry}}/2$, the local one-step estimate applies at each preceding step and produces the next tubular coordinate.
	Using \cref{eq:entry_normal_radius_increment,eq:entry_radius_conditions},
	\begin{equation} \label{eq:entry_normal_radius_bounds}
		\abs{s_k - s_0} \leq \sum_{i = 0}^{k - 1} c_6 s_i^2 \leq 4 c_6 k^{\max}_{\mathrm{entry}} s_0^2 \leq \frac{1}{2} s_0 \implies
		\frac{1}{2} s_0 \leq s_k \leq 2 s_0, \text{ for } 0 \leq k \leq k^{\max}_{\mathrm{entry}}.
	\end{equation}
	Let $k_1$ be the first index in $\{0, \ldots, N_{\mathrm{entry}}\}$ with $1 - \alpha_{p_{k_1}}(u_{k_1}) \leq \nu_0$.
	Such an index exists: if not, then for $0 \leq k < N_{\mathrm{entry}}$ one would have $\alpha_{p_k}(u_k) \in [\nu_0, 1 - \nu_0)$, and \cref{eq:morse_bott_alignment_gain,eq:entry_normal_radius_bounds,eq:entry_radius_conditions} would give
	\begin{equation*}
		\alpha_{p_{k + 1}}(u_{k + 1}) \geq \alpha_{p_k}(u_k) + c_{\nu_0}^{(1)} - c_{\nu_0}^{(2)} s_k \geq \alpha_{p_k}(u_k) + \frac{c_{\nu_0}^{(1)}}{2}.
	\end{equation*}
	Hence $\alpha_{p_{N_{\mathrm{entry}}}}(u_{N_{\mathrm{entry}}}) \geq \nu_0 + N_{\mathrm{entry}} c_{\nu_0}^{(1)} / 2 \geq 1 - \nu_0$, a contradiction.
	At this first near-top time, \cref{eq:alignment_transverse_component_bounds} yields
	\begin{equation} \label{eq:entry_transverse_component_bound}
		\norm{(I - V_1(p_{k_1}))u_{k_1}} \leq \sqrt{\nu_0/c_1} \leq \frac{1}{2}\sqrt{c_3/c_2}.
	\end{equation}
	For $k_1 \leq k \leq k^{\max}_{\mathrm{entry}}$, set
	\begin{equation*}
		e_k \coloneqq \norm{(I - V_1(p_k))u_k}.
	\end{equation*}
	We claim that $e_k \leq \sqrt{c_3/c_2}$ throughout this interval.
	This is true at $k_1$ by \cref{eq:entry_transverse_component_bound}.
	If it is true at $k < k^{\max}_{\mathrm{entry}}$, then \cref{eq:alignment_transverse_component_bounds} gives $1 - \alpha_{p_k}(u_k) \leq c_3$, and \cref{eq:morse_bott_transverse_contraction,eq:entry_normal_radius_bounds,eq:entry_radius_conditions} give
	\begin{equation*}
		e_{k + 1} \leq c_4 e_k + c_5 s_k \leq c_4\sqrt{c_3/c_2} + 2 c_5 s_0 \leq \sqrt{c_3/c_2}.
	\end{equation*}
	Therefore the near-top regime persists.
	Iterating the same inequality gives
	\begin{equation*}
		e_k \leq c_4^{k - k_1} e_{k_1} + \frac{2 c_5}{1 - c_4} s_0 \leq c_4^{k - k_1} + \frac{2 c_5}{1 - c_4} s_0.
	\end{equation*}
	Let
	\begin{equation*}
		k_0 \coloneqq k_1 + \ceil{c_{\mathrm{entry}} \log{\frac{1}{s_0}}}.
	\end{equation*}
	Then $k_0 \leq k^{\max}_{\mathrm{entry}}$.
	Using \cref{eq:entry_alignment_time_bound,eq:top_cone_slope,eq:entry_normal_radius_bounds},
	\begin{equation*}
		e_{k_0} \leq \frac{L}{4} s_0 + \frac{2 c_5}{1 - c_4} s_0 \leq \frac{L}{2} s_0 \leq L s_{k_0}.
	\end{equation*}
	Multiplying by $s_{k_0}$ gives the cone inequality in \cref{eq:top_cone_entry_bounds}, and $s_{k_0} \leq 2 s_0 \leq r$ gives the size bound.
	The prefix estimate and the asserted tubular-coordinate validity on the prefix follow from \cref{eq:entry_normal_radius_bounds} and the induction above.
\end{proof}

\subsubsection{Dynamics in the top cone}

\begin{proposition}[Dynamics in the top cone] \label{prop:morse_bott_top_cone_dynamics}
	There exists $\bar{\kappa} > 0$ such that for every fixed $\kappa \geq \bar{\kappa}$ there are $r_{\mathrm{cone}} > 0$ and constants $0 < c_7 \leq c_8 < \infty$ with the following property.
	Consider a localized trajectory segment consisting of an initial point $\theta$ and its next two iterates $\theta^+$ and $\theta^{++}$.
	Assume that the initial point has tubular coordinates $\theta = p + n$ with $p \in \mathcal{K}$ and
	\begin{equation*}
		0 < \norm{n} \leq r_{\mathrm{cone}}, \ \norm{(I - V_1(p))n} \leq L \norm{n}^2.
	\end{equation*}
	Then $\theta^+$ and $\theta^{++}$ have tubular coordinates
	\begin{equation*}
		\theta^+ = p^+ + n^+, \ \theta^{++} = p^{++} + n^{++},
	\end{equation*}
	with base points in $\mathcal{K}$ and $0 < \norm{n^+}, \norm{n^{++}} < \bar{r}$, and the following hold:
	\begin{enumerate}
		\item
			\begin{equation} \label{eq:top_cone_invariance}
				\norm{(I - V_1(p^+))n^+} \leq L \norm{n^+}^2, \text{ and if also } \norm{n} \leq r_{\mathrm{cone}}/2, \text{ then } \norm{n^+} \leq r_{\mathrm{cone}};
			\end{equation}
		\item
			\begin{equation} \label{eq:morse_bott_two_step_normal_contraction}
				\norm{n}^2 - c_8 \norm{n}^4 \leq \norm{n^{++}}^2 \leq \norm{n}^2 - c_7 \norm{n}^4;
			\end{equation}
		\item
			\begin{equation} \label{eq:morse_bott_sharpness_gradient_drift}
				p^+
				= \operatorname{exp}_p\bigl(-\norm{n}^2 \nabla_{\mathcal{M}} \log{\lambda_1}(p) + \bigo{\norm{n}^3}\bigr).
			\end{equation}
	\end{enumerate}
\end{proposition}
\begin{proof}
	Let
	\begin{equation*}
		\lambda_\star \coloneqq \inf_{p \in \mathcal{K}}\lambda_1(p) > 0.
	\end{equation*}
	The main point is the two-step expansion on the exact top line.
	Work first in a trivialization with a local unit top eigenvector field $v_1$.
	Let $S_p$ be the corresponding one-dimensional $\mathrm{CDP}_{\rm{rat}}$ map, i.e.
	\begin{equation*}
		S_p(s) \coloneqq s - 4 \frac{\mathcal{L}(p + s v_1(p))}{\left(\frac{\d}{\d s} \mathcal{L}(p + s v_1(p))\right)^2}
		\frac{1}{1 + \kappa \mathcal{L}(p + s v_1(p))}
		\frac{\d}{\d s} \mathcal{L}(p + s v_1(p)).
	\end{equation*}
	Since
	\begin{equation*}
		\mathcal{L}(p + s v_1(p)) = \frac{1}{2} s^2 \lambda_1(p) + \bigo{s^3},
	\end{equation*}
	one has
	\begin{equation*}
		S_p(s) = -s + s^2 R_{s, 2}(p) + \bigo{s^3}
	\end{equation*}
	for a local continuous coefficient $R_{s, 2}$.
	Therefore, for $s > 0$ small,
	\begin{equation*}
			\abs{S_p(\pm s)} = s \mp s^2 R_{s, 2}(p) + \bigo{s^3}.
	\end{equation*}
	The full dynamics and the one-dimensional dynamics agree in top-line size through order $s^2$:
	\begin{equation} \label{eq:top_direction_scalar_update_comparison}
		n = \pm s v_1(p) \implies \norm{n^+} = \abs{S_p(\pm s)} + \bigo{s^3}.
	\end{equation}
	Differentiating $s \mapsto \mathcal{L}(p + s v_1(p))$ gives
	\begin{equation*}
		\frac{\d}{\d s} \mathcal{L}(p + s v_1(p)) = \nabla \mathcal{L}(p + s v_1(p))^T v_1(p).
	\end{equation*}
	At the points $p \pm s v_1(p)$, \Cref{lem:morse_bott_tubular_expansions} gives that the tangential component and the normal components orthogonal to $v_1(p)$ are $\bigo{s^2}$, hence
	\begin{equation*}
		\norm{\nabla \mathcal{L}(p \pm s v_1(p))}^2
		= \left(\nabla \mathcal{L}(p \pm s v_1(p))^T v_1(p)\right)^2 + \bigo{s^4}.
	\end{equation*}
	It follows that the full $\mathrm{CDP}_{\rm{rat}}$ factor and its one-dimensional counterpart differ by $\bigo{s^2}$.
	This stepsize-factor difference contributes $\bigo{s^3}$ to the post-step point.
	The tangential and transverse-normal gradient components contribute $\bigo{s^2}$ to the point, but only $\bigo{s^3}$ to the normal length: they are orthogonal to the leading top-normal component of size $s$, and the base-point conversion also changes the normal length only by $\bigo{s^3}$.
	This proves \cref{eq:top_direction_scalar_update_comparison}.
	Along the two top branches write
	\begin{equation*}
		\rho^{(1)}_\pm(p) \coloneqq \rho^{(1)}(p, \pm v_1(p)), \ \rho^{(3)}_\pm(p) \coloneqq \rho^{(3)}(p, \pm v_1(p)).
	\end{equation*}
	Since $Q_p(\pm v_1(p)) = \mp v_1(p)$ and, by \cref{eq:morse_bott_normal_damping_coefficient}, $R^{(2)}_{n, 3}(p, \pm v_1(p)) = \lambda_1(p) (\pm v_1(p))$, \cref{eq:morse_bott_radius_damping_coefficient} gives $\rho^{(3)}_\pm(p) = -\lambda_1(p)$.
	Thus each top-branch size update contributes $-\kappa \lambda_1(p) s^3$ at order $s^3$.
	Moreover, \cref{eq:top_direction_scalar_update_comparison} gives $\rho^{(1)}_+(p) = -R_{s, 2}(p)$ and $\rho^{(1)}_-(p) = R_{s, 2}(p)$.
	Hence the order-$s^2$ size changes cancel:
	\begin{equation*}
		\rho^{(1)}_+(p) + \rho^{(1)}_-(p) = 0.
	\end{equation*}
	On the top line, the first step gives $d_{\mathcal{M}}(p^+, p) = \bigo{s^2}$, $s^+ = s + \bigo{s^2}$, and $u^+ = \mp v_1(p^+) + \bigo{s}$.
	Thus the direct $\kappa$-term in the second step is again $-\kappa \lambda_1(p) s^3$ with an error $\bigo{s^4}$ in $s^{++}$.
	After squaring, this error is absorbed into the $\bigo{s^5}$ remainder.
	Applying \cref{eq:morse_bott_normal_radius_expansion} twice, first on one top branch and then on the opposite branch at the shifted base point, gives
	\begin{equation*}
		s^{++} = s + s^3 \bigl(R_{s, 3}(p) - 2 \kappa \lambda_1(p)\bigr) + \bigo{s^4}
	\end{equation*}
	for a local continuous coefficient $R_{s, 3}$.
	The regularity in \Cref{lem:morse_bott_tubular_expansions} justifies the remainder.
	By parity of the Taylor coefficients, $\rho^{(1)}$ is odd in $u$, whereas $\rho^{(2)}$, $\rho^{(3)}$, and the direction correction $R_{n, 2} - \rho^{(1)} Q_p$ are even.
	The cubic composition terms are therefore even, including $(\rho^{(1)})^2$ and the derivative of $\rho^{(1)}$ applied to that correction.
	Base-point changes contribute only at order $s^4$.
	Thus $R_{s, 3}$ is unchanged under $v_1 \mapsto -v_1$ and patches globally over $\mathcal{K}$.
	Squaring gives a globally defined continuous $R_{s^2, 4}: \mathcal{K} \to \real$ such that on the exact top line
	\begin{equation} \label{eq:top_direction_two_step_radius_expansion}
			\norm{n^{++}}^2 = s^2 + s^4 \bigl(R_{s^2, 4}(p) - 4 \kappa \lambda_1(p)\bigr) + \bigo{s^5}.
	\end{equation}

	Now consider a general vector in the cone.
	In a local top-eigenvector trivialization, choose the sign of $v_1(p)$ so that $u^T v_1(p) \geq 0$.
	The cone condition implies
	\begin{equation} \label{eq:top_cone_direction_alignment}
		\norm{u - v_1(p)} \leq \sqrt{2} L s.
	\end{equation}
	Indeed, if $t = u^T v_1(p) \geq 0$, then
	\begin{equation*}
		\norm{u - v_1(p)}^2 = 2(1 - t)
		\leq 2(1 - t^2) = 2 \norm{(I - V_1(p)) u}^2.
	\end{equation*}
	Applying \cref{eq:morse_bott_normal_radius_expansion} first at $(p, u, s)$ and then at $(p^+, u^+, s^+)$, using \cref{eq:morse_bott_radius_direction_update,eq:morse_bott_tangential_update_expansion}, gives locally Lipschitz coefficient fields $\rho_{\mathrm{cone}}^{(1)}$, $\rho_{\mathrm{cone}}^{(2)}$, and $\rho_{\mathrm{cone}}^{(3)}$, independent of $\kappa$, on the unit normal bundle over $\mathcal{K}$ such that for each fixed $\kappa$ there is a remainder $R_{\mathrm{cone}_\kappa, 5}$ with $\abs{R_{\mathrm{cone}_\kappa, 5}(p, u, s)} \leq s^5 c_\kappa$ and
	\begin{equation*}
		\norm{n^{++}}^2
		= s^2 + s^3 \rho_{\mathrm{cone}}^{(1)}(p, u) + s^4 \bigl(\rho_{\mathrm{cone}}^{(2)}(p, u) + \kappa \rho_{\mathrm{cone}}^{(3)}(p, u)\bigr) + R_{\mathrm{cone}_\kappa, 5}(p, u, s).
	\end{equation*}
	Here $u^+ = Q_p(u) + \bigo{s}$, $s^+ = s + \bigo{s^2}$, and $d_{\mathcal{M}}(p^+, p) = \bigo{s^2}$.
	The $\kappa$-dependent correction to $u^+$ is order $s^2$, so changed-argument $\kappa$-effects beyond the displayed coefficient are order $\kappa s^5$ in the squared norm and are absorbed into $R_{\mathrm{cone}_\kappa, 5}$.
	This is why no additional order-$\kappa s^4$ coefficient appears in the squared-norm expansion.
	On the exact top line, \cref{eq:top_direction_two_step_radius_expansion} implies
	\begin{equation*}
		\rho_{\mathrm{cone}}^{(1)}(p, \pm v_1(p))
		= 0, \ \rho_{\mathrm{cone}}^{(2)}(p, \pm v_1(p))
		= R_{s^2, 4}(p), \ \rho_{\mathrm{cone}}^{(3)}(p, \pm v_1(p))
		= -4 \lambda_1(p).
	\end{equation*}
	Using \cref{eq:top_cone_direction_alignment}, local Lipschitz continuity, and the finite top-eigenvector trivialization family, there is $c_{\mathrm{cone}}^{(1)} = c_{\mathrm{cone}}^{(1)}(L)$ such that on the cone
	\begin{equation*}
		\abs{\rho_{\mathrm{cone}}^{(1)}(p, u)} + \abs{\rho_{\mathrm{cone}}^{(2)}(p, u) - R_{s^2, 4}(p)} + \abs{\rho_{\mathrm{cone}}^{(3)}(p, u) + 4 \lambda_1(p)} \leq c_{\mathrm{cone}}^{(1)} s.
	\end{equation*}
	Thus the $s^3 \rho_{\mathrm{cone}}^{(1)}$ term contributes only a bounded $s^4$ coefficient, and after increasing $c_{\mathrm{cone}}^{(1)}$ if necessary,
	\begin{equation} \label{eq:top_cone_two_step_radius_expansion}
		\norm{n^{++}}^2
		= s^2 + s^4 \bigl(R_{s^2, 4}(p) - 4 \kappa \lambda_1(p) + R_{\mathrm{cone}, 4}(p, u, s) + R_{\mathrm{cone}_\kappa, 4}(p, u, s)\bigr),
	\end{equation}
	where $\abs{R_{\mathrm{cone}, 4}(p, u, s)} \leq c_{\mathrm{cone}}^{(1)}$ uniformly and $\abs{R_{\mathrm{cone}_\kappa, 4}(p, u, s)} \leq c_\kappa s$ for fixed $\kappa$.
	Choose $\bar{\kappa} > 0$ so large that
	\begin{equation*}
		\sup_{p \in \mathcal{K}} R_{s^2, 4}(p) + c_{\mathrm{cone}}^{(1)} + 1 - 4 \bar{\kappa} \lambda_\star < 0.
	\end{equation*}
	Fix $\kappa \geq \bar{\kappa}$.
	Let $r_\kappa \in (0, \bar{r})$ be small enough that \Cref{lem:morse_bott_tubular_expansions} and the fixed-$\kappa$ estimates used in \Cref{lem:morse_bott_spectral_alignment} apply.
	Choose $r_{\mathrm{cone}} \in (0, r_\kappa/4)$ small enough that the one-step bound $s^+ = s + \bigo{s^2}$ implies $s^+ \leq 2s$ whenever $s \leq r_{\mathrm{cone}}$.
	Then any localized segment in the statement with $s \leq r_{\mathrm{cone}}$ has tubular coordinates for $\theta^+$ and $\theta^{++}$: indeed $s^+ \leq 2r_{\mathrm{cone}} \leq r_\kappa$, and localization places the newly produced tubular base points in $\mathcal{K}$.
	After shrinking $r_{\mathrm{cone}}$ so that $\abs{R_{\mathrm{cone}_\kappa, 4}(p, u, s)} \leq 1$ in \cref{eq:top_cone_two_step_radius_expansion}, the coefficient in \cref{eq:top_cone_two_step_radius_expansion} is at most a negative constant.
	It is also bounded below by a finite constant depending on the fixed $\kappa$.
	This proves \cref{eq:morse_bott_two_step_normal_contraction} after naming the constants $c_7$ and $c_8$.
	We prove cone invariance next.
	Shrink $r_{\mathrm{cone}}$ further so that the fixed-$\kappa$ version of \cref{eq:morse_bott_transverse_contraction} applies and $c_2 L^2 r_{\mathrm{cone}}^2 \leq c_3$.
	Then the cone condition and \cref{eq:alignment_transverse_component_bounds} imply $1 - \alpha_p(u) \leq c_3$, hence
	\begin{equation*}
		\norm{(I - V_1(p^+))u^+} \leq c_4 \norm{(I - V_1(p)) u} + c_5 s \leq (c_4 L + c_5) s.
	\end{equation*}
	Also \cref{eq:morse_bott_radius_direction_update} gives $\abs{s^+ - s} \leq s^2 c_{\mathrm{cone}}^{(2)}$ on the cone.
	Since $L(1 - c_4) - c_5 > 0$, shrink $r_{\mathrm{cone}}$ once more so that
	\begin{equation*}
		c_{\mathrm{cone}}^{(2)} r_{\mathrm{cone}} \leq \min\left\{\frac{1}{2}, \frac{L(1 - c_4) - c_5}{L}\right\}.
	\end{equation*}
	Then $s \leq s^+/(1 - c_{\mathrm{cone}}^{(2)} r_{\mathrm{cone}})$, and therefore
	\begin{equation*}
		\norm{(I - V_1(p^+))u^+} \leq \frac{c_4 L + c_5}{1 - c_{\mathrm{cone}}^{(2)} r_{\mathrm{cone}}} s^+ \leq L s^+.
	\end{equation*}
	Multiplying by $s^+$ proves \cref{eq:top_cone_invariance}.
	If $s \leq r_{\mathrm{cone}}/2$, then $s^+ \leq s + s^2 c_{\mathrm{cone}}^{(2)} \leq r_{\mathrm{cone}}$.
	Finally, \cref{eq:morse_bott_tangential_update_expansion} gives
	\begin{equation*}
		p^+
		= \operatorname{exp}_p\left(-s^2 \frac{u^T H(p) u}{u^T H(p)^2 u} \nabla_{\mathcal{M}} \bigl(u^T H(p) u\bigr) + \bigo{s^3}\right).
	\end{equation*}
	On the cone, $u = \pm v_1(p) + \bigo{s}$.
	On the exact top line,
	\begin{equation*}
		\frac{(\pm v_1)^T H(p)(\pm v_1)}{(\pm v_1)^T H(p)^2(\pm v_1)} = \frac{1}{\lambda_1(p)},
	\end{equation*}
	and the simple-eigenvalue perturbation formula gives
	\begin{equation*}
		\nabla_{\mathcal{M}} \bigl((\pm v_1(p))^T H(p)(\pm v_1(p))\bigr)
		= \nabla_{\mathcal{M}} \lambda_1(p).
	\end{equation*}
	Thus the top-line coefficient is $\nabla_{\mathcal{M}} \log{\lambda_1}(p)$.
	By $C^1$ regularity, the coefficient differs from this by $\bigo{s}$ on the cone, and \cref{eq:morse_bott_sharpness_gradient_drift} follows.
\end{proof}
\begin{corollary}[Polynomial decay estimates in the cone] \label{cor:morse_bott_normal_decay}
	For every fixed $\kappa \geq \bar{\kappa}$, after shrinking $r_{\mathrm{cone}}$ if necessary, the following holds.
	Suppose that, for some $k_0$ and $T \in \natural\cup\{\infty\}$ with $T > k_0$, the tail $(p_k, n_k)_{k_0 \leq k < T}$ lies in the cone of \Cref{prop:morse_bott_top_cone_dynamics}.
	Then the two parity subsequences are decreasing and there are constants $c_9, c_{10}, c_{11} > 0$ such that
	\begin{align}
		\frac{c_9 s_{k_0}^2}{1 + j s_{k_0}^2} &\leq s_{k_0 + 2j}^2 \leq \frac{c_{10} s_{k_0}^2}{1 + j s_{k_0}^2}, \notag\\
		\frac{c_9 s_{k_0}^2}{1 + j s_{k_0}^2} &\leq s_{k_0 + 1 + 2j}^2 \leq \frac{c_{10} s_{k_0}^2}{1 + j s_{k_0}^2}. \label{eq:morse_bott_normal_decay_bounds}
	\end{align}
	for all indices for which the terms are defined.
	On an infinite tail,
	\begin{equation} \label{eq:morse_bott_normal_power_sums}
		s_k \to 0, \ \sum_{k \geq k_0} s_k^2
		= \infty, \ \sum_{k \geq k_0} s_k^3 \leq c_{11} s_{k_0}, \ \sum_{k \geq k_0} s_k^4 \leq c_{11} s_{k_0}^2.
	\end{equation}
\end{corollary}
\begin{proof}
	By \cref{eq:morse_bott_two_step_normal_contraction},
	\begin{align*}
		s_{k_0 + 2j}^2 - c_8 s_{k_0 + 2j}^4 &\leq s_{k_0 + 2j + 2}^2 \leq s_{k_0 + 2j}^2 - c_7 s_{k_0 + 2j}^4,\\
		s_{k_0 + 1 + 2j}^2 - c_8 s_{k_0 + 1 + 2j}^4 &\leq s_{k_0 + 1 + 2j + 2}^2 \leq s_{k_0 + 1 + 2j}^2 - c_7 s_{k_0 + 1 + 2j}^4.
	\end{align*}
	After shrinking $r_{\mathrm{cone}}$, assume $c_8 r_{\mathrm{cone}}^2 \leq 1/2$.
	Taking reciprocals gives the standard $1/j$ decay estimates for the two parity subsequences.
	The one-step estimate $s_{k_0 + 1} = s_{k_0} + \bigo{s_{k_0}^2}$ gives $s_{k_0 + 1}\asymp s_{k_0}$, so both parity classes satisfy \cref{eq:morse_bott_normal_decay_bounds}.
	In particular, for $\ell \in \{0, 1\}$,
	\begin{equation*}
		s_{k_0 + \ell + 2j}^3 \lesssim \frac{s_{k_0}^3}{(1 + j s_{k_0}^2)^{3/2}}, \
		s_{k_0 + \ell + 2j}^4 \lesssim \frac{s_{k_0}^4}{(1 + j s_{k_0}^2)^2}.
	\end{equation*}
	The sums in \cref{eq:morse_bott_normal_power_sums} follow by summing these bounds over the two parity classes.
\end{proof}
\begin{proof}[Proof of \Cref{thm:localized_cdp_dynamics}] \hypertarget{proof:localized_cdp_dynamics}
	The nearest-point projections of $K \cap U$ have compact closure in the closed manifold $\mathcal{M}$.
	Since $\alpha_{p_0}(n_0) \geq \delta^2 > 0$, the proofs of \Cref{prop:morse_bott_top_cone_entry,prop:morse_bott_top_cone_dynamics} and \Cref{cor:morse_bott_normal_decay} apply on this closure without the critical-point assumption. Use \Cref{lem:damping_comparison} for CDP.

	Take $\kappa_0 = \bar{\kappa}$. After shrinking the fixed-$\kappa$ radii, \cref{eq:top_cone_two_step_radius_expansion} gives a uniform positive lower bound for $c_7$.
	Damping changes the normal and base-point updates by $\bigo{\kappa s^3}$ and $\bigo{\kappa s^4}$, respectively, using the tubular projection derivative $\Pi_{T_p\mathcal{M}} + \bigo{s}$.
	Taking these radii at most $(1 + \kappa)^{-1}$ makes the normal $\bigo{s^2}$ and tangential $\bigo{s^3}$ errors uniform in $\kappa$.

	The entry time is $\bigo{1 + \log{(1 / s_0)}} = \smallo{s_0^{-2}}$, with $s_k \asymp s_0$ before entry.
	The upper bound in \cref{eq:morse_bott_normal_decay_bounds} therefore gives $s_k^2 = \bigo{(s_0^{-2} + k)^{-1}}$ for all $k \geq 0$, uniformly in $\kappa$.
	Before entry, the drift error is $\bigo{s_0^2}$ and hence $\bigo{k^{-3/2}}$ for $k \geq 1$, since $s_0^2(1 + \log{(1 / s_0)})^{3/2}$ is bounded.
	After entry, \cref{eq:morse_bott_sharpness_gradient_drift} gives error $\bigo{s_k^3} = \bigo{k^{-3/2}}$, including the $\bigo{s_k^4}$ correction from replacing the Riemannian exponential by the ambient update.
	The entry estimate and cone invariance give tubular coordinates for every iterate, and we may take $k_0 = 1$.
\end{proof}

\subsubsection{Conclusion of the proof}

\begin{proof}[Proof of \Cref{thm:morse_bott_cdp_convergence}] \hypertarget{proof:morse_bott_cdp_convergence}
	Let $\bar{\kappa}$ be the constant from \Cref{prop:morse_bott_top_cone_dynamics}.
	The radius $\bar{r}$ is fixed before $\kappa$, all smaller radii chosen below may depend on the fixed value of $\kappa$.
	Fix $\kappa \geq \bar{\kappa}$ and $\nu \in (0, 1)$.
	Choose $r_{\mathrm{cone}} \in (0, \bar{r})$ small enough that \Cref{prop:morse_bott_top_cone_dynamics} and \Cref{cor:morse_bott_normal_decay} hold for this $\kappa$.
	After this choice, we choose the initial radius $r_0$.
	Apply \Cref{prop:morse_bott_top_cone_entry} with target radius $r_{\mathrm{cone}}/2$.
	Let $r_0$ be the corresponding radius.
	Let $(\theta_k)_{k \geq 0}$ be a trajectory localized as in \Cref{ass:localized_sharpness_spectrum}, and assume that $\theta_0 = p_0 + n_0$ is in tubular coordinates with $0 < \norm{n_0} \leq r_0$ and $\alpha_{p_0}(n_0) \geq \nu$.
	By \Cref{prop:morse_bott_top_cone_entry}, the trajectory has tubular coordinates on the entry prefix and there exists $k_0$ such that
	\begin{equation} \label{eq:top_cone_tail_initialization}
		0 < s_{k_0} \leq \frac{1}{2} r_{\mathrm{cone}}, \ \norm{(I - V_1(p_{k_0}))n_{k_0}} \leq L s_{k_0}^2.
	\end{equation}
	We now propagate the cone forever.
	From \cref{eq:top_cone_tail_initialization}, \Cref{prop:morse_bott_top_cone_dynamics} gives the cone condition at $k_0 + 1$ and $s_{k_0 + 1} \leq r_{\mathrm{cone}}$.
	The two-step estimate gives $s_{k_0 + 2} \leq s_{k_0} \leq r_{\mathrm{cone}}/2$.
	Repeating this argument on the two parity classes gives tubular coordinates for every $k \geq k_0$ and
	\begin{equation*}
		s_k \leq r_{\mathrm{cone}}, \ \norm{(I - V_1(p_k))n_k} \leq L s_k^2.
	\end{equation*}
	The entry prefix satisfies $s_k \leq 2s_0 < \bar{r}$, while the tail satisfies $s_k \leq r_{\mathrm{cone}} < \bar{r}$.
	Thus localization, together with the local one-step estimates, gives the full tubular validity claimed in the theorem.
	Applying \Cref{cor:morse_bott_normal_decay} to the full tail gives
	\begin{equation*}
		s_k \to 0, \ \sum_{k \geq k_0} s_k^2
		= \infty, \ \sum_{k \geq k_0} s_k^3 < \infty, \ \sum_{k \geq k_0} s_k^4 < \infty.
	\end{equation*}
	In particular, $n_k \to 0$.
	Moreover, \cref{eq:morse_bott_two_step_normal_contraction,eq:morse_bott_normal_decay_bounds} imply, for all sufficiently large $k$,
	\begin{equation} \label{eq:morse_bott_normal_decay_rate}
		s_{k + 2}^2 = s_k^2 - \Theta(s_k^4), \ s_k \asymp k^{-1/2}.
	\end{equation}
	Set $\eta_k \coloneqq s_k^2$. By \cref{eq:morse_bott_sharpness_gradient_drift}, for $k \geq k_0$,
	\begin{equation} \label{eq:localized_sharpness_gradient_descent}
		p_{k + 1}
		= \operatorname{exp}_{p_k}\bigl(-\eta_k \nabla_{\mathcal{M}} \log{\lambda_1}(p_k) + \epsilon_k\bigr), \ \epsilon_k \in T_{p_k}\mathcal{M}, \ \norm{\epsilon_k} \leq c_{12} s_k^3.
	\end{equation}
	Thus
	\begin{equation*}
		\eta_k > 0, \ \eta_k \to 0, \ \sum_k \eta_k
		= \infty, \ \sum_k \eta_k^2 < \infty, \ \sum_k \norm{\epsilon_k} < \infty.
	\end{equation*}
	In particular, $\sum_k \norm{\epsilon_k}^2 < \infty$.
	We now prove $p_k \to p_\star$.
	Because $\mathcal{K}$ is compact and $\log{\lambda_1}$ is $C^2$ in the finite simple-eigenvalue trivialization family around $\mathcal{K}$, there is $c_{13} > 0$ such that, for all $p \in \mathcal{K}$ and all sufficiently small $v \in T_p\mathcal{M}$,
	\begin{equation*}
		\log{\lambda_1}(\operatorname{exp}_p v) \leq \log{\lambda_1}(p) + \nabla_{\mathcal{M}} \log{\lambda_1}(p)^T v + c_{13} \norm{v}^2.
	\end{equation*}
	Applying this to $v_k = -\eta_k \nabla_{\mathcal{M}} \log{\lambda_1}(p_k) + \epsilon_k$ and using the uniform bound of $\nabla_{\mathcal{M}} \log{\lambda_1}$ on $\mathcal{K}$ gives, for all sufficiently large $k$,
	\begin{align*}
		\log{\lambda_1}(p_{k + 1})
		&\leq \log{\lambda_1}(p_k)
		- \eta_k \norm{\nabla_{\mathcal{M}} \log{\lambda_1}(p_k)}^2
		+ c_{13} \eta_k^2 + c_{13} \norm{\epsilon_k} + c_{13} \norm{\epsilon_k}^2.
	\end{align*}
	After increasing $c_{13}$ if necessary, this gives
	\begin{equation} \label{eq:sharpness_descent_with_errors}
		\log{\lambda_1}(p_{k + 1}) \leq \log{\lambda_1}(p_k) - \eta_k \norm{\nabla_{\mathcal{M}} \log{\lambda_1}(p_k)}^2 + c_{13} \eta_k^2 + c_{13} \norm{\epsilon_k}.
	\end{equation}
	Since $\log{\lambda_1}$ is bounded below on $\mathcal{K}$ and the two error series are summable, summing \cref{eq:sharpness_descent_with_errors} yields
	\begin{equation} \label{eq:sharpness_gradient_weighted_sum}
		\sum_k \eta_k \norm{\nabla_{\mathcal{M}} \log{\lambda_1}(p_k)}^2 < \infty.
	\end{equation}
	The critical points of $\log{\lambda_1}$ are exactly the critical points of $\lambda_1$, so $p_\star$ is the unique critical point of $\log{\lambda_1}$ in $\mathcal{K}$.
	Moreover, since $\log: (0, \infty) \to \real$ is strictly increasing, $p_\star$ is also a strict local minimum of $\log{\lambda_1}$ on $\mathcal{K}$.
	Fix $\epsilon_0 > 0$ and define
	\begin{equation*}
		\mathcal{K}_{\epsilon_0} \coloneqq \{p \in \mathcal{K}: d_{\mathcal{M}}(p, p_\star) \geq \epsilon_0\}.
	\end{equation*}
	If $\mathcal{K}_{\epsilon_0}$ is empty, there is nothing to prove for this $\epsilon_0$.
	Otherwise compactness and uniqueness of the critical point give
	\begin{equation*}
		m_{\epsilon_0} \coloneqq \inf_{p \in \mathcal{K}_{\epsilon_0 / 2}} \norm{\nabla_{\mathcal{M}} \log{\lambda_1}(p)} > 0.
	\end{equation*}
	Then \cref{eq:sharpness_gradient_weighted_sum} implies
	\begin{equation} \label{eq:sharpness_away_step_sum}
		\sum_{k: p_k \in \mathcal{K}_{\epsilon_0 / 2}} \eta_k \leq m_{\epsilon_0}^{-2} \sum_k \eta_k \norm{\nabla_{\mathcal{M}} \log{\lambda_1}(p_k)}^2 < \infty.
	\end{equation}
	We claim that $\mathcal{K}_{\epsilon_0}$ is visited only finitely many times.
	If not, then either the tail eventually remains in $\mathcal{K}_{\epsilon_0 / 2}$, contradicting \cref{eq:sharpness_away_step_sum} and $\sum_k \eta_k = \infty$, or there are infinitely many disjoint transitions from $\mathcal{K}\setminus \mathcal{K}_{\epsilon_0 / 2}$ to $\mathcal{K}_{\epsilon_0}$.
	For each sufficiently late such transition, choose its entrance index $k_j^+$ and the preceding index $k_j^- < k_j^+$ so that
	\begin{equation*}
		p_{k_j^-} \notin \mathcal{K}_{\epsilon_0 / 2}, \ p_k \in \mathcal{K}_{\epsilon_0 / 2} \text{ for } k_j^- < k \leq k_j^+, \ p_{k_j^+} \in \mathcal{K}_{\epsilon_0}.
	\end{equation*}
	The intrinsic distance traveled from $p_{k_j^-}$ to $p_{k_j^+}$ is at least $\epsilon_0/2$.
	Let
	\begin{equation*}
		c_{14} \coloneqq \sup_{p \in \mathcal{K}} \norm{\nabla_{\mathcal{M}} \log{\lambda_1}(p)}.
	\end{equation*}
	From \cref{eq:localized_sharpness_gradient_descent}, $d_{\mathcal{M}}(p_{k + 1}, p_k) \leq c_{14} \eta_k + \norm{\epsilon_k}$.
	Since $\eta_k \to 0$ and $\sum_k \norm{\epsilon_k} < \infty$, every sufficiently late transition satisfies
	\begin{equation*}
		c_{14} \eta_{k_j^-} \leq \frac{\epsilon_0}{8}, \ \sum_{k = k_j^-}^{k_j^+ - 1} \norm{\epsilon_k} \leq \frac{\epsilon_0}{8}.
	\end{equation*}
	Consequently,
	\begin{equation*}
		c_{14} \sum_{k = k_j^- + 1}^{k_j^+ - 1} \eta_k \geq \frac{\epsilon_0}{2} - c_{14} \eta_{k_j^-} - \sum_{k = k_j^-}^{k_j^+ - 1} \norm{\epsilon_k} \geq \frac{\epsilon_0}{4}.
	\end{equation*}
	But all indices $k_j^- + 1 \leq k \leq k_j^+ - 1$ lie in $\mathcal{K}_{\epsilon_0 / 2}$, so infinitely many such transitions contradict \cref{eq:sharpness_away_step_sum}.
	Therefore $\mathcal{K}_{\epsilon_0}$ is visited finitely often.
	Since $\epsilon_0 > 0$ is arbitrary, $p_k \to p_\star$.
	The cone condition gives the directional alignment:
	\begin{equation*}
		\norm{(I - V_1(p_k))\frac{n_k}{\norm{n_k}}}
		\leq L s_k \to 0.
	\end{equation*}
	For unit $u \in N_{p_k}\mathcal{M}$,
	\begin{equation*}
		\dist{u}{\{w \in N_{p_k}\mathcal{M}: \norm{w} = 1, V_1(p_k)w = w\}}
		\leq \sqrt{2} \norm{(I - V_1(p_k))u},
	\end{equation*}
	so
	\begin{equation*}
		\dist{\frac{n_k}{\norm{n_k}}}{\left\{u \in N_{p_k}\mathcal{M}\midbar\norm{u} = 1, H(p_k) u = \lambda_1(p_k)u\right\}} \to 0.
	\end{equation*}
	Since $p_k \to p_\star$, any local unit top eigenvector field $v_1$ defined near $p_\star$ is available for all sufficiently large $k$, and the last display becomes
	\begin{equation*}
		\dist{\frac{n_k}{\norm{n_k}}}{\{\pm v_1(p_k)\}} \to 0.
	\end{equation*}
	Finally, \cref{eq:localized_sharpness_gradient_descent,eq:morse_bott_normal_decay_rate} give the claimed base recursion, tangent error bound, and
	\begin{equation*}
		d_{\mathcal{M}}(p_{k + 1}, p_k) = \bigo{k^{-1}}.
	\end{equation*}
	All assertions of \Cref{thm:morse_bott_cdp_convergence} follow.
\end{proof}

\subsection{Sharpness convergence rate} \label{subsec:sharpness_convergence_rate}

\begin{figure}[htb]
	\centering
	\includegraphics[width = \textwidth]{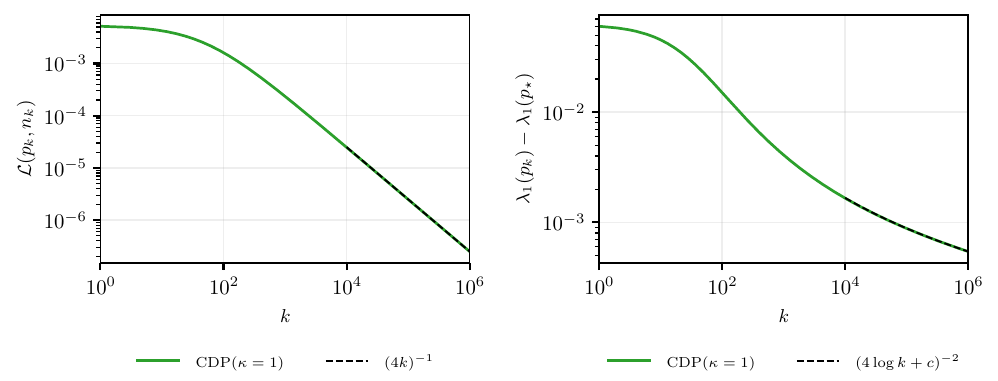}
	\caption{
		CDP dynamics for $\mathcal{L}(p, n) = \frac{1}{2}(1 + p^4)n^2$ over $10^6$ steps, starting from $(0.5, 0.1)$ with $\kappa = 1$.
		Left, loss. Right, sharpness gap $\lambda_1(p_k) - \lambda_1(p_\star) = p_k^4$.
		The constant $c$ is chosen to match the final sharpness gap.
		The loss decays as $k^{-1}$, while the sharpness gap decays only as $(\log{k})^{-2}$.
	}
	\label{fig:quartic_sharpness_dynamics}
\end{figure}

\Cref{thm:morse_bott_cdp_convergence} gives convergence to a local minimum of sharpness without a rate for the sharpness gap.
\begin{corollary}[Polynomial sharpness rate] \label{cor:morse_bott_sharpness_rate}
	Under the hypotheses of \Cref{thm:morse_bott_cdp_convergence}, assume additionally that $\lambda_1$ satisfies a local Polyak-\L{}ojasiewicz (PL) inequality: for some $\mu > 0$ and all $p \in \mathcal{M}$ sufficiently close to $p_\star$,
	\begin{equation*}
		\frac{1}{2}\norm{\nabla_{\mathcal{M}}\lambda_1(p)}^2 \geq \mu\bigl(\lambda_1(p) - \lambda_1(p_\star)\bigr).
	\end{equation*}
	Then, for each fixed admissible $\kappa$ and either damping rule, there exists $\gamma > 0$ such that
	\begin{equation} \label{eq:morse_bott_sharpness_rate}
		\lambda_1(p_k) - \lambda_1(p_\star) = \bigo{k^{-\gamma}}, \text{ and } d_{\mathcal{M}}(p_k, p_\star) = \bigo{k^{-\gamma/2}}.
	\end{equation}
\end{corollary}
\begin{proof} \hypertarget{proof:morse_bott_sharpness_rate}{}
	Set $\eta_k \coloneqq \norm{n_k}^2$.
	By \Cref{thm:morse_bott_cdp_convergence}, $p_k \to p_\star$, $\eta_k \asymp k^{-1}$, and $\norm{\epsilon_k} = \bigo{k^{-3/2}}$.
	Fix $b > 0$ such that $\eta_k \geq b/k$ for all sufficiently large $k$.
	All subsequent estimates are for sufficiently large $k$, so the sharpness gap is nonnegative and the local PL inequality applies.
	Since $p_k \to p_\star$ and $\lambda_1$ is continuous with $\lambda_1(p_\star) > 0$, we can also assume that $\lambda_1(p_k) \leq 2\lambda_1(p_\star)$.

	Using the equation for the update of $p$ in \Cref{thm:morse_bott_cdp_convergence} and $\nabla_{\mathcal{M}}\log{\lambda_1} = \nabla_{\mathcal{M}}\lambda_1 / \lambda_1$, a Taylor expansion along the exponential map gives
	\begin{equation} \label{eq:morse_bott_sharpness_taylor_expansion}
		\lambda_1(p_{k + 1}) - \lambda_1(p_k) = -\frac{\eta_k}{\lambda_1(p_k)}\norm{\nabla_{\mathcal{M}}\lambda_1(p_k)}^2 + \innerproduct{\nabla_{\mathcal{M}}\lambda_1(p_k)}{\epsilon_k} + \bigo{\eta_k^2\norm{\nabla_{\mathcal{M}}\lambda_1(p_k)}^2 + \norm{\epsilon_k}^2}.
	\end{equation}
	Next, Young's inequality with weight $\eta_k/(4\lambda_1(p_k))$ bounds the mixed term:
	\begin{equation} \label{eq:morse_bott_sharpness_young}
		\innerproduct{\nabla_{\mathcal{M}}\lambda_1(p_k)}{\epsilon_k} \leq \frac{\eta_k}{4\lambda_1(p_k)}\norm{\nabla_{\mathcal{M}}\lambda_1(p_k)}^2 + \frac{\lambda_1(p_k)}{\eta_k}\norm{\epsilon_k}^2.
	\end{equation}
	Let $C > 0$ be the constant in the Taylor remainder bound, independent of $k$.
	Since $\eta_k \to 0$ and $\lambda_1(p_k)$ is bounded, $C \eta_k \lambda_1(p_k) \leq 1/4$ for sufficiently large $k$, so
	\begin{equation} \label{eq:morse_bott_sharpness_taylor_remainder}
		C\eta_k^2\norm{\nabla_{\mathcal{M}}\lambda_1(p_k)}^2 \leq \frac{\eta_k}{4\lambda_1(p_k)}\norm{\nabla_{\mathcal{M}}\lambda_1(p_k)}^2.
	\end{equation}
	Combining \cref{eq:morse_bott_sharpness_taylor_expansion,eq:morse_bott_sharpness_young,eq:morse_bott_sharpness_taylor_remainder} and $\norm{\epsilon_k}^2/\eta_k = \bigo{k^{-2}}$, gives
	\begin{equation*}
		\lambda_1(p_{k + 1}) - \lambda_1(p_k) \leq -\frac{\eta_k}{2\lambda_1(p_k)}\norm{\nabla_{\mathcal{M}}\lambda_1(p_k)}^2 + \bigo{k^{-2}}.
	\end{equation*}
	Adding $\lambda_1(p_k) - \lambda_1(p_\star)$ to both sides gives
	\begin{equation*}
		\lambda_1(p_{k + 1}) - \lambda_1(p_\star) \leq \lambda_1(p_k) - \lambda_1(p_\star) - \frac{\eta_k}{2\lambda_1(p_k)}\norm{\nabla_{\mathcal{M}}\lambda_1(p_k)}^2 + \bigo{k^{-2}}.
	\end{equation*}
	Applying the local PL inequality, $\eta_k \geq b/k$, and $\lambda_1(p_k) \leq 2\lambda_1(p_\star)$ gives
	\begin{equation*}
		\lambda_1(p_{k + 1}) - \lambda_1(p_\star) \leq \left(1 - \frac{\mu b}{2\lambda_1(p_\star)k}\right)\bigl(\lambda_1(p_k) - \lambda_1(p_\star)\bigr) + \bigo{k^{-2}}.
	\end{equation*}
	For any $0 < \gamma < \min\{\mu b/(2\lambda_1(p_\star)), 1\}$, comparison with a sufficiently large multiple of $k^{-\gamma}$ gives $\lambda_1(p_k) - \lambda_1(p_\star) = \bigo{k^{-\gamma}}$.

	To obtain the distance rate, fix a sufficiently small $t > 0$.
	Since $\lambda_1$ is $C^2$ near its local minimum $p_\star$, Taylor expansion (for all sufficiently large $k$) gives
	\begin{equation*}
		\lambda_1(p_\star) \leq \lambda_1\bigl(\operatorname{exp}_{p_k}(-t\nabla_{\mathcal{M}}\lambda_1(p_k))\bigr) \leq \lambda_1(p_k) - \frac{t}{2}\norm{\nabla_{\mathcal{M}}\lambda_1(p_k)}^2.
	\end{equation*}
	The gradient step remains near $p_\star$ because $p_k \to p_\star$ and $\nabla_{\mathcal{M}}\lambda_1(p_k) \to 0$.
	Rearranging gives $\norm{\nabla_{\mathcal{M}}\lambda_1(p_k)}^2 \leq (2/t)\bigl(\lambda_1(p_k) - \lambda_1(p_\star)\bigr) = \bigo{k^{-\gamma}}$.
	The distance is at most the length of the update geodesic, which equals the norm of its initial tangent vector.
	Since $\lambda_1$ is bounded away from zero near $p_\star$ and $\eta_k \asymp k^{-1}$, the triangle inequality gives
	\begin{equation*}
		d_{\mathcal{M}}(p_{k + 1}, p_k) \leq \frac{\eta_k}{\lambda_1(p_k)}\norm{\nabla_{\mathcal{M}}\lambda_1(p_k)} + \norm{\epsilon_k} = \bigo{k^{-1 - \gamma/2}} + \bigo{k^{-3/2}}.
	\end{equation*}
	Summing the step lengths and using $p_k \to p_\star$ and $0 < \gamma < 1$ gives
	\begin{equation*}
		d_{\mathcal{M}}(p_k, p_\star) \leq \sum_{j = k}^{\infty} d_{\mathcal{M}}(p_{j + 1}, p_j) = \bigo{k^{-\gamma/2}} + \bigo{k^{-1/2}} = \bigo{k^{-\gamma/2}}.
	\end{equation*}
	This proves the distance estimate in \cref{eq:morse_bott_sharpness_rate}.
\end{proof}
\begin{remark}[Strict minimality is not enough]
	Consider
	\begin{equation*}
		\mathcal{L}(p, n) = \frac{1}{2}(1 + p^4)n^2.
	\end{equation*}
	Then $\mathcal{M} = \{n = 0\}$ and $\lambda_1(p) = 1 + p^4$, with a unique critical point and strict minimum at $p_\star = 0$.
	The loss is Morse-Bott along $\mathcal{M}$, but the sharpness fails the local PL inequality at $0$, since $\abs{\lambda_1'(p)}^2 / (2(\lambda_1(p) - 1)) = 8p^2 \to 0$ as $p \to 0$.

	Fix $\kappa > 0$ and write $\beta_k$ for the $\mathrm{CDP}_{\rm{exp}}$ stepsize from \cref{eq:cdp_step_size} at $(p_k, n_k)$.
	The exact updates are
	\begin{equation*}
		p_{k + 1} = p_k - 2\beta_k p_k^3 n_k^2, \text{ and } n_{k + 1} = \left(1 - \beta_k(1 + p_k^4)\right)n_k.
	\end{equation*}
	For $p_0 > 0$ and $n_0 \neq 0$ sufficiently small, these give $0 < p_{k + 1} < p_k$ and $0 < \abs{n_{k + 1}} < \abs{n_k}$, so the iterates stay near $(0, 0)$.
	Since $e^{-\kappa\mathcal{L}} = 1 - \kappa\mathcal{L} + \bigo{\mathcal{L}^2}$, the normal update gives
	\begin{equation*}
		n_{k + 1}^2 = n_k^2 - \left(2\kappa(1 + p_k^4) + \frac{16p_k^6}{(1 + p_k^4)^2}\right)n_k^4 + \bigo{n_k^6}.
	\end{equation*}
	The coefficient of $n_k^4$ is bounded above and away from zero, so taking reciprocals gives $n_{k + 1}^{-2} - n_k^{-2} \asymp 1$, and therefore $n_k^2 \asymp k^{-1}$.
	Similarly, the tangential update gives
	\begin{equation*}
		p_{k + 1}^{-2} - p_k^{-2} = \frac{8n_k^2}{1 + p_k^4} + \bigo{n_k^4} \asymp k^{-1}.
	\end{equation*}
	Summing yields $p_k^{-2} \asymp \log{k}$, and consequently
	\begin{equation*}
		\mathcal{L}(p_k, n_k) \asymp k^{-1}, \text{ while } \lambda_1(p_k) - 1 = p_k^4 \asymp (\log{k})^{-2}.
	\end{equation*}
	Thus the loss decays polynomially while the sharpness gap decays more slowly than every polynomial rate.
	\Cref{fig:quartic_sharpness_dynamics} illustrates these different rates.
\end{remark}

\section{CDP dynamics on a quadratic Morse-Bott family} \label{sec:quadratic_morse_bott_dynamics}

In \Cref{sec:nonlinear_morse_bott_structure}, we identify the quadratic Morse-Bott family that governs the loss at leading order near a regular interpolation manifold.
In this appendix, we study the corresponding exact quadratic model in flat Euclidean product geometry.
This isolates the CDP dynamics without higher-order loss terms or the geometric corrections of a curved minimizer manifold.
Let $\tau, m \in \natural$ with $\tau > m$ and consider
\begin{equation} \label{eq:quadratic_morse_bott_loss}
	\mathcal{L}(p, n) \coloneqq \frac{1}{2} n^T H(p) n,
	\text{ for } (p, n) \in \real^{\tau - m} \times \real^m,
\end{equation}
where $H(p) \in \real^{m \times m}$ is symmetric positive definite for every $p \in \real^{\tau - m}$.
The manifold of global minima of $\mathcal{L}$ is
\begin{equation*}
	\mathcal{M} \coloneqq \{(p, n) \in \real^{\tau - m} \times \real^m \mid n = 0\},
\end{equation*}
and the normal space at $(p, 0)$ is the $n$-space.
In particular, $\mathcal{L}$ is Morse-Bott along $\mathcal{M}$ in the sense of \Cref{def:morse_bott_loss}.

For each $p \in \real^{\tau - m}$, let $\lambda_1(p) \geq \dots \geq \lambda_m(p) > 0$ denote the eigenvalues of $H(p)$.
Moreover we define $(v_i(p))_{i = 1}^m$ as the corresponding orthonormal eigenbasis, and write $E_1(p) \coloneqq \ker{\left(H(p) - \lambda_1(p) I_m\right)}$ for the top eigenspace.
Throughout this section we assume $H \in C^2(\real^{\tau - m}, \real^{m \times m})$.
By continuity and positive definiteness, for every compact set $\mathcal{K} \subset \real^{\tau - m}$ there exist $0 < \lambda_{\min, \mathcal{K}} \leq \lambda_{\max, \mathcal{K}} < \infty$ such that
\begin{equation*}
	\lambda_{\min, \mathcal{K}} I_m \preceq H(p) \preceq \lambda_{\max, \mathcal{K}} I_m, \text{ for all } p \in \mathcal{K}.
\end{equation*}

\subsection{CDP iteration for quadratic Morse-Bott functions}

We now analyze the CDP iteration with $\kappa = 0$ in \cref{eq:cdp_step_size}, off the minimizer manifold, i.e. for $n_k \neq 0$:
\begin{equation} \label{eq:quadratic_polyak_update}
	\theta_{k + 1} = \theta_k - \beta(p_k, n_k) \nabla \mathcal{L}(\theta_k),
	\text{ where } \theta_k = (p_k, n_k) \in \real^{\tau - m} \times \real^m.
\end{equation}
A direct computation gives the gradients
\begin{equation} \label{eq:quadratic_loss_gradients}
		\nabla_p \mathcal{L}(p, n) = \frac{1}{2}
		\begin{pmatrix}
			n^T (\partial_{p_1} H(p)) n \\
			\vdots \\
			n^T (\partial_{p_{\tau - m}} H(p)) n
		\end{pmatrix},
		\text{ and }
	\nabla_n \mathcal{L}(p, n) = H(p) n,
\end{equation}
and, in particular,
\begin{equation} \label{eq:quadratic_gradient_norm}
	\norm{\nabla \mathcal{L}(p, n)}^2
	= \norm{H(p) n}^2 + \frac{1}{4} \sum_{j = 1}^{\tau - m} \left(n^T (\partial_{p_j} H(p)) n\right)^2.
\end{equation}
For small $\norm{n}$ and $p$ in a fixed compact set, the second term in \cref{eq:quadratic_gradient_norm}
is of order $\bigo{\norm{n}^4}$, while the first one is $\Theta(\norm{n}^2)$.
Writing down the stepsize for our case explicitly, we obtain
\begin{equation} \label{eq:quadratic_polyak_step_size}
	\beta(p, n) \coloneqq 4 \frac{\mathcal{L}(p, n)}{\norm{\nabla \mathcal{L}(p, n)}^2}
	= 4 \frac{\frac{1}{2} n^T H(p) n}{\norm{H(p)n}^2 + \bigo{\norm{n}^4}}
	= 2 \frac{n^T H(p) n}{\norm{H(p)n}^2} + \bigo{\norm{n}^2},
\end{equation}
as $n \to 0$ (assuming $n \neq 0$) uniformly for $p$ in compact subsets of $\real^{\tau - m}$, where we used the $C^2$ regularity of $H$ and the compact spectral bounds above.

\subsection{Spectral alignment for fixed \texorpdfstring{$p$}{p}}

We first analyze the purely quadratic map in $n$ with $p$ frozen, i.e., assuming that the curvature of our quadratic model is fixed.
This already captures the key spectral effect.
Here and throughout, a superscript ${}^+$ denotes a one-step update.
\begin{lemma}[Norm preservation and alignment at fixed $p$] \label{lem:fixed_hessian_norm_and_alignment}
	Fix $p \in \real^{\tau - m}$, $n \in \real^m \setminus \{0\}$, and consider the CDP update with $\kappa = 0$ applied to $n \mapsto \mathcal{L}(p, n)$ with $p$ fixed:
	\begin{equation*}
		n^+ = n - \beta_0(p, n) H(p) n, \text{ with }
		\beta_0(p, n) \coloneqq 2 \frac{n^T H(p) n}{\norm{H(p) n}^2}.
	\end{equation*}
	Let $H(p) v_i(p) = \lambda_i(p) v_i(p)$ be the eigen-decomposition and write $n = \sum_{i = 1}^m \xi_i v_i(p)$.
	For $n \neq 0$, define the Rayleigh quotient by
	\begin{equation*}
		R_p(n) \coloneqq \frac{n^T H(p)n}{\norm{n}^2}.
	\end{equation*}
	The following results hold.
	\begin{enumerate}
		\item \textit{Norm preservation.} $\norm{n^+} = \norm{n}$.
		\item \textit{Alignment sharpening.}
			\begin{equation*}
				R_p(n^+) \geq R_p(n).
			\end{equation*}
			Moreover, the inequality is strict exactly when the spectral support\footnote{The spectral support of $n$ is the set of eigenvalues $\lambda_i(p)$ such that $\xi_i \neq 0$.} of $n$ contains at least two distinct values.
		\item \textit{Top-eigenspace convergence.} If $n_0$ has nonzero projection onto $E_1(p)$, then the distance of $n_k / \norm{n_k}$ to $E_1(p) \cap \mathbb{S}^{m - 1}$ converges to zero geometrically.
	\end{enumerate}
\end{lemma}
\begin{proof}
	See \hyperlink{proof:fixed_hessian_norm_and_alignment}{proof of \Cref{lem:fixed_hessian_norm_and_alignment}} in \Cref{subsec:quadratic_morse_bott_proofs}.
\end{proof}

\begin{remark}[Rayleigh quotient and alignment] \label{rem:rayleigh_quotient_alignment}
	Following \Cref{lem:fixed_hessian_norm_and_alignment}, for fixed $p \in \mathcal{K}$, the \textit{Rayleigh defect} $\lambda_1(p) - u^T H(p)u$ is the scalar quantity sharpened by the CDP map.
	It connects the two quantities used to measure alignment: the \textit{top eigenspace distance} $\dist{u}{E_1(p) \cap \mathbb{S}^{m - 1}}$ and the \textit{residual} $\norm{\left(H(p) - \lambda_1(p) I_m\right) u}$.

	Indeed, assuming fixed $p \in \mathcal{K}$, $n = su \in \real^m$, $s > 0$, and $u \in \mathbb{S}^{m - 1}$, there exist $c_1, c_2 > 0$ such that
	\begin{equation*}
		\norm{\left(H(p) - \lambda_1(p)I_m\right) u}^2 \leq c_1 \left(\lambda_1(p) - u^T H(p)u\right), \
		\lambda_1(p) - u^T H(p)u \leq c_2 \dist{u}{E_1(p) \cap \mathbb{S}^{m - 1}}^2.
	\end{equation*}
	Thus top eigenspace distance $\bigo{s}$ implies Rayleigh defect $\bigo{s^2}$, which implies residual $\bigo{s}$.
	If the top eigenspace is separated from the lower spectrum by a gap bounded below independently of $s$, then residual $\bigo{s}$ also implies top eigenspace distance $\bigo{s}$.
	Indeed, if $r \coloneqq \dim{E_1(p)} < m$ and $\lambda_1(p) - \lambda_{r + 1}(p) \geq \delta > 0$, then
	\begin{equation*}
		\dist{u}{E_1(p) \cap \mathbb{S}^{m - 1}} \leq c_3 \norm{\left(H(p) - \lambda_1(p)I_m\right)u}
		= \bigo{s}.
	\end{equation*}
	When the top eigenvectors are not separated from the lower spectrum, this last implication may fail, so the residual is the condition to preserve, while the top eigenspace distance is recovered only once the top branch is resolved.
\end{remark}

\subsection{Scale separation for the full \texorpdfstring{$(p, n)$}{(p, n)}-dynamics}

We now return to the full dynamics in \cref{eq:quadratic_polyak_update}, where both $p_k$ and $n_k$ move.
The key point is a separation of scales: the normal component moves on the order of $\norm{n_k}$ per step, whereas the tangential component moves on the order of $\norm{n_k}^2$.
\begin{theorem}[Scale separation near the manifold] \label{thm:quadratic_scale_separation}
	For every compact set $\mathcal{K} \subset \real^{\tau - m}$, there exists a radius $\rho > 0$ and constants $c_1, c_2, c_3 \in (0, \infty)$ with $c_1 \leq c_2$ such that the following holds.
	For any $(p, n)$ with $p \in \mathcal{K}$ and $0 < \norm{n} \leq \rho$, let $(p^+, n^+)$ be one CDP update as in \cref{eq:quadratic_polyak_update}.
	Then
	\begin{equation} \label{eq:quadratic_scale_separation}
		c_1 \norm{n} \leq \norm{n^+ - n} \leq c_2 \norm{n}, \text{ and }
		\norm{p^+ - p} \leq c_3 \norm{n}^2.
	\end{equation}
	In particular, for $\norm{n}$ small, the normal motion is of order $\Theta(\norm{n})$ while the tangential motion is of order $\bigo{\norm{n}^2}$, so the $n$-component evolves on a faster scale than the $p$-component.
\end{theorem}
\begin{proof}
	See \hyperlink{proof:quadratic_scale_separation}{proof of \Cref{thm:quadratic_scale_separation}} in \Cref{subsec:quadratic_morse_bott_proofs}.
\end{proof}

\begin{remark}[Normal length] \label{rem:quadratic_normal_length}
	The full update also satisfies the exact identity
	\begin{equation*}
		\norm{n^+}^2 = \norm{n}^2 - \norm{p^+ - p}^2,
	\end{equation*}
	obtained by substituting the stepsize in \cref{eq:quadratic_polyak_step_size} into $\norm{n - \beta H(p)n}^2$.
	Thus the normal length is nonincreasing, with its decrease determined by the tangential displacement.
\end{remark}

\begin{figure}[htbp!]
	\centering
	\includegraphics[width = \textwidth]{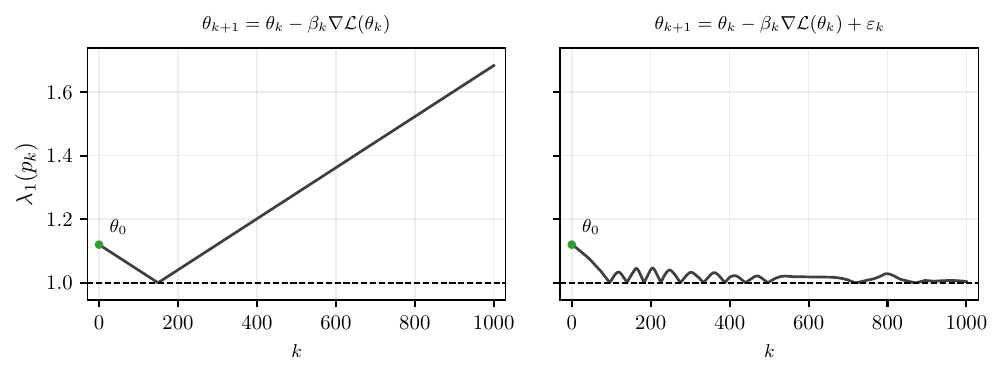}
	\caption{
		Sharpness dynamics over $1000$ CDP steps with $\kappa = 0$ starting from $(p_0, n_{1,0}, n_{2,0}) = (-0.12, 0, 0.03)$, without noise (left) and with independent centered Gaussian noise of standard deviation $10^{-3}$ in $n_1$ and $n_2$ (right).
		After the eigenvalue crossing, the noiseless trajectory remains on the same branch and sharpness increases, whereas noise allows switching and keeps sharpness near its minimum.
	}
	\label{fig:quadratic_eigenvalue_switching}
\end{figure}

\begin{theorem}[Tangential drift along $\partial_p \log{\lambda_1}$] \label{thm:quadratic_tangential_drift}
	For every $\mathcal{K} \subset \real^{\tau - m}$ compact, there exists a radius $\rho > 0$ such that the following holds.
	Let $k \in \natural$, $(p_k, n_k) \in \mathcal{K} \times \real^m$, and define
	\begin{equation*}
		s_k \coloneqq \norm{n_k}, \
		u_k \coloneqq \frac{n_k}{\norm{n_k}}, \
		r_k \coloneqq \dim{E_1(p_k)}.
	\end{equation*}
	Assume
	\begin{equation} \label{eq:quadratic_residual_alignment}
		0 < s_k \leq \rho, \
		\norm{\left(H(p_k) - \lambda_1(p_k)I_m\right) u_k} = \bigo{s_k}.
	\end{equation}
	Then the CDP update in \cref{eq:quadratic_polyak_update} propagates the residual condition\footnote{This one-step estimate does not ensure a uniform residual bound across repeated eigenvalue crossings.}
	\begin{equation} \label{eq:quadratic_residual_alignment_update}
		\norm{\left(H(p_{k + 1}) - \lambda_1(p_{k + 1})I_m\right) u_{k + 1}} = \bigo{s_{k + 1}},
	\end{equation}
	and satisfies the branch drift\footnote{Here $\partial_p$ denotes the limiting subdifferential \cite[Definition 8.3(b)]{variational_rockafellar_1998}, and the vector $u$ is held fixed in $\nabla_p\left(u^T H(p)u\right)$.}
	\begin{equation} \label{eq:quadratic_rayleigh_branch_drift}
		p_{k + 1} = p_k - s_k^2 \nabla_p\log{\left(u_k^T H(p_k)u_k\right)} + \bigo{s_k^4},
	\end{equation}
	with constants uniform on $\mathcal{K}$ for a fixed input residual bound.
	Moreover, if either $r_k = m$, or $r_k < m$ and $\lambda_1(p_k) - \lambda_{r_k + 1}(p_k) \geq \delta$ for some $\delta > 0$ independent of $s_k$, then
	\begin{equation} \label{eq:quadratic_sharpness_subgradient_drift}
		p_{k + 1} = p_k - s_k^2 \zeta_k + \bigo{s_k^3}, \
		\zeta_k \in \partial_p \log{\lambda_1(p_k)},
	\end{equation}
	with the remainder constant depending on $\delta$.
	If, additionally, $r_k = 1$, then $\lambda_1$ is $C^2$ near $p_k$ and
	\begin{equation} \label{eq:quadratic_sharpness_gradient_drift}
		p_{k + 1} = p_k - s_k^2 \nabla_p\log{\lambda_1(p_k)} + \bigo{s_k^3}.
	\end{equation}
\end{theorem}
\begin{proof}
	See \hyperlink{proof:quadratic_tangential_drift}{proof of \Cref{thm:quadratic_tangential_drift}} in \Cref{subsec:quadratic_morse_bott_proofs}.
\end{proof}
\begin{remark}[Eigenvector switching and branch selection] \label{rem:quadratic_eigenvalue_switching}
	Given a configuration $(p, n) \in \mathcal{K} \times \real^m$, suppose that $\lambda_1(p)$ has multiplicity and that $n \in E_1(p)$, top eigenspace.
	Then assuming $n = s u$, $s > 0$, $u \in \mathbb{S}^{m - 1}$, the selected subgradient is given by
	\begin{equation*}
		\zeta \coloneqq \frac{\nabla_p\left(u^T H(p)u\right)}{\lambda_1(p)} \in \partial_p \log{\lambda_1(p)},
	\end{equation*}
	and the corresponding infinitesimal motion along the manifold is $-\zeta$.

	For a fixed $p \in \mathcal{K}$ and a direction $d \in \real^{\tau - m}$, define the linear map
	\begin{equation*}
		Z_p(d) \coloneqq \Pi_{E_1(p)} \circ \left(D_d H(p)\big|_{E_1(p)}\right): E_1(p) \to E_1(p),
	\end{equation*}
	where $D_d H(p) = \sum_{j = 1}^{\tau - m} d_j \partial_{p_j} H(p)$ is the derivative of the matrix $H$ in the direction $d$.
	The quadratic form induced by $Z_p(d)$ should be viewed as the first-order score for branch selection after an infinitesimal motion along $d$.

	At $p$, every $w \in E_1(p) \cap \mathbb{S}^{m - 1}$ has the same Rayleigh quotient, namely $w^T H(p)w = \lambda_1(p)$.
	After a step of size $\eta$ in direction $d$, we have
	\begin{equation*}
		w^T H(p + \eta d)w = \lambda_1(p) + \eta w^T Z_p(d)w + \smallo{\eta}.
	\end{equation*}
	The $0^{\text{th}}$-order term $\lambda_1(p)$ is the same for every $w \in E_1(p) \cap \mathbb{S}^{m - 1}$, so the branch that becomes top (to first order) after the motion $d$ is determined by the first-order score $w^T Z_p(d)w$.
	Thus the directions which become top (to first order) are exactly the maximizers of $w^T Z_p(d)w$ on $E_1(p) \cap \mathbb{S}^{m - 1}$, equivalently the top eigenspace of $Z_p(d)$.

	Applying this with $d = -\zeta$, we call nonzero projection of $u$ onto the top eigenspace of $Z_p(-\zeta)$ the \textit{branch selection condition}.

	If it fails, the dynamics can stay locked on the branch selected by $u$ while $\lambda_1$ is controlled by another branch.
	This is a real switching obstruction.
\end{remark}
\begin{theorem}[Finite entrance into residual alignment] \label{thm:quadratic_residual_alignment_entry}
	Let $\mathcal{K} \subset \real^{\tau - m}$ be compact.
	Fix $p_0 \in \mathcal{K}$ and a unit vector $u_0$ with nonzero projection onto $E_1(p_0)$.
	For sufficiently small $s_0 > 0$, consider a CDP trajectory as in \cref{eq:quadratic_polyak_update}, initialized at $(p_0, s_0 u_0)$ and satisfying $p_k \in \mathcal{K}$.
	Then for some $k_0 = \bigo{\log{(1 / s_0)}}$,
	\begin{equation} \label{eq:quadratic_residual_alignment_entry}
		\norm{\left(H(p_{k_0}) - \lambda_1(p_{k_0})I_m\right)n_{k_0}} = \bigo{s_{k_0}^2}.
	\end{equation}
\end{theorem}
\begin{proof}
	See \hyperlink{proof:quadratic_residual_alignment_entry}{proof of \Cref{thm:quadratic_residual_alignment_entry}} in \Cref{subsec:quadratic_morse_bott_proofs}.
\end{proof}
\begin{remark}[Exact quadratic summary] \label{rem:quadratic_morse_bott_dynamics_summary}
	Near $\mathcal{M} = \{n = 0\}$, the exact quadratic model isolates the mechanism cleanly.
	For frozen $p$, \Cref{lem:fixed_hessian_norm_and_alignment} shows that the normal CDP map preserves $\norm{n}$ and aligns $n$ with the top eigenspace of $H(p)$ (alignment measured with Rayleigh quotient), provided $n_0$ has nonzero projection onto it.
	In the full $(p, n)$-dynamics, \Cref{thm:quadratic_scale_separation} shows that the normal component moves on the fast scale $\norm{n}$, while the tangential component moves on the slower scale $\norm{n}^2$.
	\Cref{thm:quadratic_residual_alignment_entry} gives finite entry into residual alignment for fixed $p_0$ and $u_0$ with nonzero initial top projection, provided $s_0$ is sufficiently small depending on these initial data.
	\Cref{rem:quadratic_eigenvalue_switching} explains how the dynamics can remain locked on another branch at switches.
	Finally, \Cref{thm:quadratic_tangential_drift} shows that the residual condition gives the branch drift, and that this drift becomes a subgradient step for $\log{\lambda_1}$ once the top branch is resolved.
	If the resolved top eigenspace is one-dimensional, the same formula becomes a simple gradient step for $\log{\lambda_1}$.
	\Cref{sec:morse_bott_dynamics} gives the corresponding local convergence theorem for general Morse-Bott losses with CDP steps.
\end{remark}
\begin{remark}[A switching example] \label{rem:quadratic_switching_example}
	Let $p \in (-1, 1)$ and
	\begin{equation*}
		H(p) \coloneqq
		\begin{pmatrix}
			1 + p & 0 \\
			0 & 1 - p
		\end{pmatrix}, \
		\mathcal{L}(p, n) = \frac{1}{2}n^T H(p)n.
	\end{equation*}
	Then $\lambda_1(p) = 1 + \abs{p}$ and $\log{\lambda_1(p)} = \log{(1 + \abs{p})}$.
	At the switch $p = 0$, the top eigenspace is all of $\real^2$ and $\partial_p \log{\lambda_1(0)} = [-1, 1]$.
	If $u = u_1 e_1 + u_2 e_2 \in \mathbb{S}^1$, then
	\begin{equation*}
		u^T H(0)u = 1, \
		\left.\nabla_p\log{\left(u^T H(p)u\right)}\right|_{p = 0} = u_1^2 - u_2^2 \in [-1, 1].
	\end{equation*}
	Thus \cref{eq:quadratic_rayleigh_branch_drift} selects one subgradient at the switch.
	For $u = e_2$, the branch selection condition fails: the selected motion makes $e_1$ top.
	The noiseless dynamics preserve $n_1 = 0$, so after the crossing the normal component is exactly orthogonal to the new top eigendirection $e_1$.
	The trajectory therefore continues along the $e_2$ branch, increasing $p$ and hence $\lambda_1(p) = 1 + p$ for $p > 0$.
	The same example shows why a pointwise positive gap is not enough to recover the smooth gradient formula.
	Take $p_k = s_k$ and $u_k = e_2$.
	Then the top eigenspace is $\operatorname{span}\{e_1\}$ and the gap is positive, but it equals $2s_k$ and therefore vanishes at the normal scale.
	Moreover
	\begin{equation*}
		\norm{\left(H(p_k) - \lambda_1(p_k)I_2\right)n_k} = 2s_k^2, \
		\dist{u_k}{E_1(p_k) \cap \mathbb{S}^1} = \sqrt{2}.
	\end{equation*}
	The branch drift follows $1 - p$ and gives
	\begin{equation*}
		p_{k + 1} = p_k + \frac{s_k^2}{1 - p_k} + \bigo{s_k^4},
	\end{equation*}
	whereas the smooth top-gradient formula would give
	\begin{equation*}
		p_{k + 1} = p_k - \frac{s_k^2}{1 + p_k} + \bigo{s_k^3}.
	\end{equation*}
	The two signs are different.
	This is why the gap in \Cref{thm:quadratic_tangential_drift} must be bounded below independently of $s_k$.

	Small random perturbations in $n_1$ can move the dynamics out of the plane $n_1 = 0$.
	In \Cref{fig:quadratic_eigenvalue_switching}, this makes the trajectory turn back toward $p = 0$ after the switch.
	This does not guarantee immediate alignment, and our drift formulas apply to the dynamics without noise.
\end{remark}

\subsection{Numerical example}

\begin{figure}[htb]
	\centering
	\includegraphics[width = 0.49\textwidth]{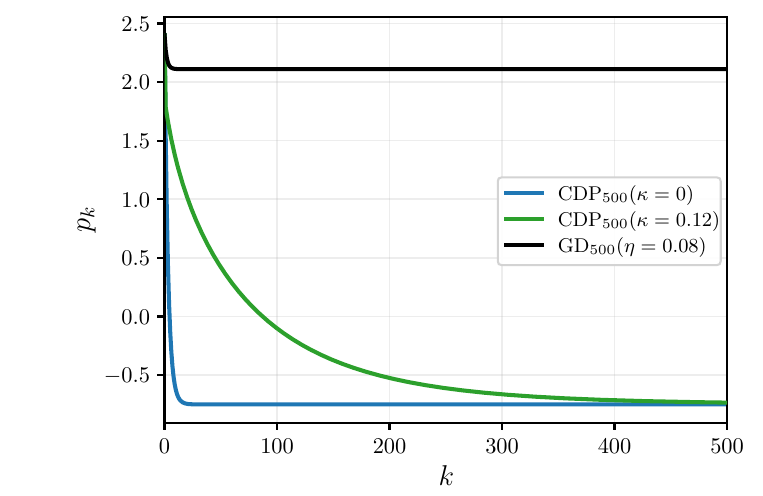}
	\includegraphics[width = 0.49\textwidth]{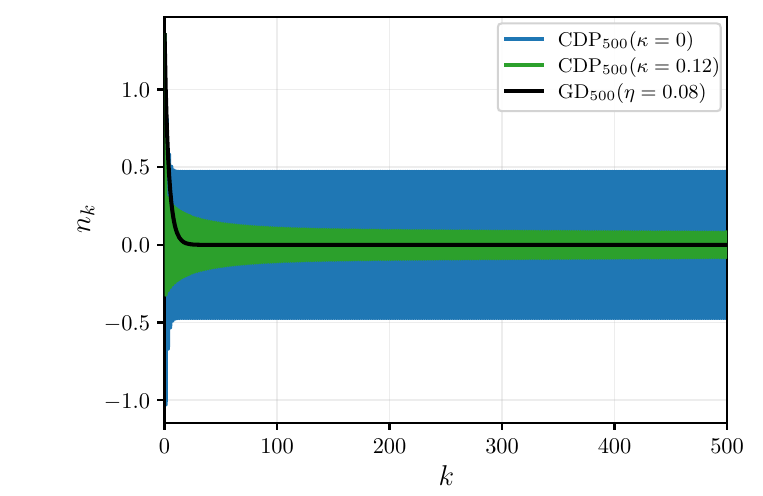}
	\caption{
		Trajectories of the tangential and normal coordinates $p_k$ and $n_k$.
		CDP with $\kappa = 0$ reaches the flat region quickly but keeps a visible period-$2$ oscillation in $n_k$, while a mild damping $\kappa = 0.12$ contracts the oscillation and retains the same tangential drift.
		By contrast, GD with $\eta = 0.08$ reaches interpolation with much less tangential motion and therefore remains near a sharper point.
	}
	\label{fig:quadratic_morse_bott_coordinates}
\end{figure}

\begin{figure}[htb]
	\centering
	\raisebox{-0.5\height}{\includegraphics[width = 0.49\textwidth]{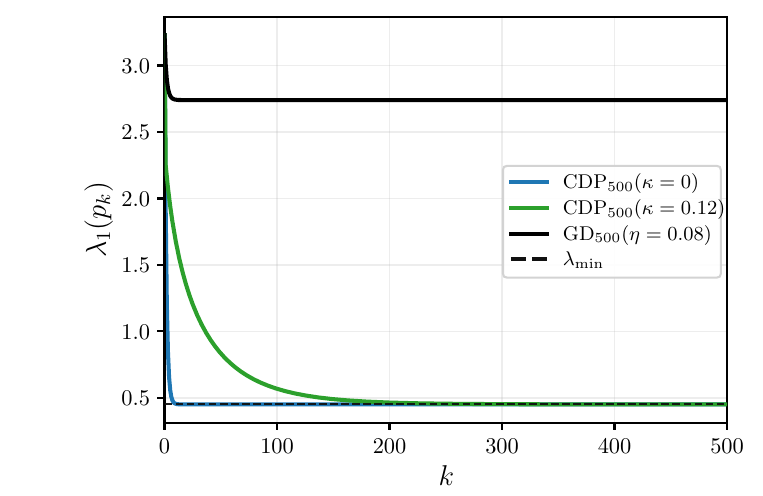}}
	\raisebox{-0.5\height}{\includegraphics[width = 0.49\textwidth]{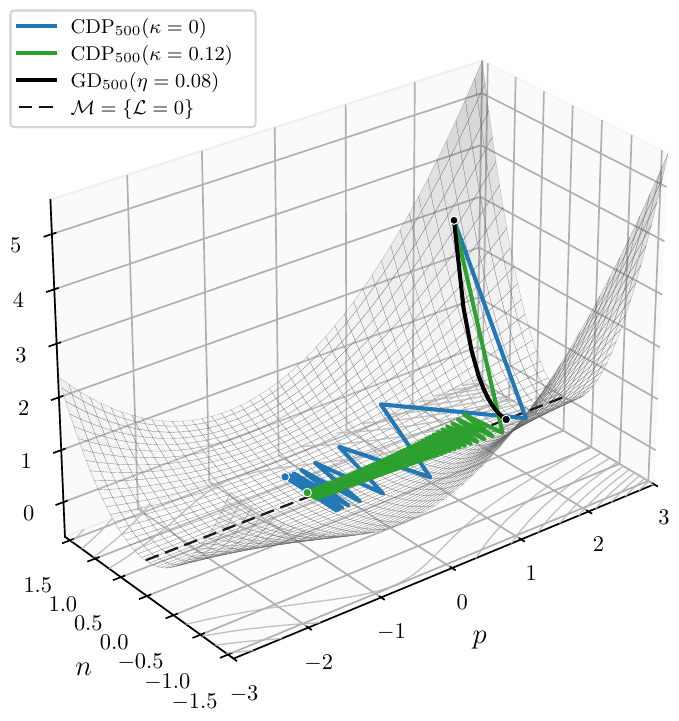}}
	\caption{
		Trajectories of the curvature $\lambda_1(p_k)$ together with the quadratic Morse-Bott surface and the interpolation manifold $\mathcal{M} = \{\mathcal{L} = 0\}$.
		The CDP trajectories first approach the manifold and then drift along it toward flatter interpolating points, with $\kappa = 0$ showing the expected period-$2$ normal oscillations and $\kappa = 0.12$ smoothing it out.
	}
	\label{fig:quadratic_morse_bott_sharpness}
\end{figure}

The reduced quadratic model captures cleanly the alignment-and-drift mechanism for the CDP dynamics.
In \Cref{fig:quadratic_morse_bott_coordinates,fig:quadratic_morse_bott_sharpness}, we use a deterministic Morse-Bott model with one tangential and one normal coordinate ($\tau = 2$, $m = 1$), so that $H(p)$ reduces to a scalar curvature,
\begin{equation*}
	\lambda(p) = \lambda_{\min} + \lambda_{\mathrm{quadratic}}(p - p_\star)^2, \
	\mathcal{L}(p, n) = \frac{1}{2} \lambda(p) n^2.
\end{equation*}
For the run shown in the figures, we take $p_\star = -0.75$, $\lambda_{\min} = 0.45$, and $\lambda_{\mathrm{quadratic}} = 0.28$, so that
\begin{equation*}
	\mathcal{L}(p, n) = \frac{1}{2}\bigl(0.45 + 0.28 (p + 0.75)^2\bigr)n^2.
\end{equation*}
Hence the manifold of minima is $\mathcal{M} = \{n = 0\}$, and the flattest interpolating point lies at $p = -0.75$.
CDP with $\kappa = 0$ quickly reaches the minimum curvature level along $\mathcal{M}$, maintaining a leading-order period-$2$ oscillation in the normal coordinate.
Adding a small damping (see \cref{eq:cdp_step_size}) preserves the drift toward the same flat region while shrinking the oscillation in $n_k$, whereas fixed-step GD reaches interpolation near its initial tangential location and therefore remains at a sharper point.


\subsection{Proofs for the quadratic Morse-Bott model} \label{subsec:quadratic_morse_bott_proofs}

In this section, we collect the proofs deferred from \Cref{sec:quadratic_morse_bott_dynamics}.
Throughout, we use the setup of \Cref{sec:quadratic_morse_bott_dynamics}, in particular, the quadratic model $\mathcal{L}$ in \cref{eq:quadratic_morse_bott_loss} is Morse-Bott along $\mathcal{M} = \{n = 0\}$ in the sense of \Cref{def:morse_bott_loss}.

\subsubsection{Alignment with fixed \texorpdfstring{$p$}{p}} \label{subsubsec:fixed_hessian_alignment_proof}

\begin{proof}[Proof of \Cref{lem:fixed_hessian_norm_and_alignment}] \hypertarget{proof:fixed_hessian_norm_and_alignment}
	Since $p$ is fixed throughout the proof, we write $H$, $\lambda_i$, $v_i$, and $\beta_0(n)$ in place of $H(p)$, $\lambda_i(p)$, $v_i(p)$, and $\beta_0(p, n)$.
	\begin{enumerate}
		\item
			In the eigenbasis $n = \sum_{i = 1}^m \xi_i v_i \neq 0$, we have
			\begin{equation*}
				H n = \sum_{i = 1}^m \lambda_i \xi_i v_i, \
				\norm{H n}^2 = \sum_{i = 1}^m \lambda_i^2 \xi_i^2, \text{ and }
				n^T H n = \sum_{i = 1}^m \lambda_i \xi_i^2 \implies
				\beta_0(n) = 2 \frac{\sum_{i = 1}^m \lambda_i \xi_i^2}{\sum_{i = 1}^m \lambda_i^2 \xi_i^2}.
			\end{equation*}
			The full update is
			\begin{equation*}
				n^+ = n - \beta_0(n) H n, \text{ which in coordinates is, }
				\xi_i^+ = (1 - \beta_0(n) \lambda_i) \xi_i, \forall i \in [m],
			\end{equation*}
			therefore
			\begin{equation*}
				\begin{aligned}
					\norm{n^+}^2 & = \sum_{i = 1}^m (\xi_i^+)^2 = \sum_{i = 1}^m (1 - \beta_0(n) \lambda_i)^2 \xi_i^2 = \\
					& = \sum_{i = 1}^m \left(1 - 2 \beta_0(n) \lambda_i + \beta_0(n)^2 \lambda_i^2\right) \xi_i^2 = \\
					& = \norm{n}^2 - 2 \beta_0(n) \sum_{i = 1}^m \lambda_i \xi_i^2 + \beta_0(n)^2 \sum_{i = 1}^m \lambda_i^2 \xi_i^2.
				\end{aligned}
			\end{equation*}
			By definition of $\beta_0(n)$,
			\begin{equation*}
				\beta_0(n)^2 \sum_{i = 1}^m \lambda_i^2 \xi_i^2
				= 4 \frac{\left(\sum_{i = 1}^m \lambda_i \xi_i^2\right)^2}{\sum_{i = 1}^m \lambda_i^2 \xi_i^2}, \
				2 \beta_0(n) \sum_{i = 1}^m \lambda_i \xi_i^2
				= 4 \frac{\left(\sum_{i = 1}^m \lambda_i \xi_i^2\right)^2}{\sum_{i = 1}^m \lambda_i^2 \xi_i^2},
			\end{equation*}
			so these two terms cancel and $\norm{n^+}^2 = \norm{n}^2$.
		\item
			Let
			\begin{equation*}
				\mu_1 \coloneqq \sum_{i = 1}^m \lambda_i \xi_i^2, \
				\mu_2 \coloneqq \sum_{i = 1}^m \lambda_i^2 \xi_i^2, \
				\mu_3 \coloneqq \sum_{i = 1}^m \lambda_i^3 \xi_i^2.
			\end{equation*}
			Since $\norm{n^+} = \norm{n}$, it is enough to show that $(n^+)^T H n^+ \geq n^T H n$. Using $\beta_0(n) = 2\mu_1/\mu_2$, we compute
			\begin{equation*}
				\begin{aligned}
					(n^+)^T H n^+ - n^T H n
					& = \sum_{i = 1}^m \lambda_i \left((1 - \beta_0(n) \lambda_i)^2 - 1\right)\xi_i^2 = \\
					& = \beta_0(n)^2 \mu_3 -2 \beta_0(n) \mu_2 = \\
					& = 4 \frac{\mu_1}{\mu_2^2} \left(\mu_1 \mu_3 - \mu_2^2\right).
				\end{aligned}
			\end{equation*}
			By Cauchy-Schwarz, since $\lambda_i > 0$ for all $i$,
			\begin{equation*}
				\mu_2^2
				= \left(\sum_{i = 1}^m \lambda_i^2 \xi_i^2\right)^2
				= \left(\sum_{i = 1}^m (\lambda_i^{1/2} \xi_i)(\lambda_i^{3/2} \xi_i)\right)^2 \leq \mu_1\mu_3.
			\end{equation*}
			Hence $(n^+)^THn^+ \geq n^THn$, and since the norm is preserved this gives $R_p(n^+)\geq R_p(n)$.
			Equality in Cauchy-Schwarz inequality holds exactly when the values of $\lambda_i$ are all equal on the spectral support of $n$, i.e., when $n$ is supported on a single eigenspace of $H$.
			This proves the strictness statement.
		\item
			If $E_1(p) = \real^m$, the claim is immediate.
			Otherwise, $\beta_0(n) \in [2 / \lambda_1, 2 / \lambda_m]$ and, for every $\lambda_i < \lambda_1$,
			\begin{equation*}
				\frac{\abs{1 - \beta_0(n)\lambda_i}}{\abs{1 - \beta_0(n)\lambda_1}} \leq q < 1,
			\end{equation*}
			with a common $q$ by compactness of the stepsize interval.
			The top projection remains nonzero, and the ratio of each lower component to its norm contracts by at most $q$ per step.
			Together with norm preservation, this proves geometric convergence to the top eigenspace.
	\end{enumerate}
\end{proof}

\subsubsection{Scale separation near the manifold} \label{subsubsec:quadratic_scale_separation_proof}

\begin{proof}[Proof of \Cref{thm:quadratic_scale_separation}] \hypertarget{proof:quadratic_scale_separation}
	By compactness and positive definiteness,
	\begin{equation*}
		\lambda_{\min, \mathcal{K}} \norm{n} \leq \norm{H(p)n} \leq \lambda_{\max, \mathcal{K}} \norm{n}.
	\end{equation*}
	Since $H$ is $C^2(\real^{\tau - m}, \real^{m \times m})$ and we restrict $p$ to a compact set $\mathcal{K} \subset \real^{\tau - m}$, there exists $c_H > 0$ such that
	\begin{equation*}
		\norm{\partial_{p_j} H(p)}_2 \leq c_H, \text{ for all } p \in \mathcal{K}, \ j
		= 1, \dots, \tau - m.
	\end{equation*}
	Hence
	\begin{equation} \label{eq:quadratic_tangential_gradient_bound}
		\abs{n^T (\partial_{p_j} H(p)) n} \leq c_H \norm{n}^2 \implies
		4 \norm{\nabla_p \mathcal{L}(p, n)}^2
		= \sum_{j = 1}^{\tau - m} \left(n^T (\partial_{p_j} H(p)) n\right)^2
		\leq (\tau - m) c_H^2 \norm{n}^4.
	\end{equation}
	From \cref{eq:quadratic_gradient_norm},
	\begin{equation*}
		\begin{gathered}
			\norm{\nabla \mathcal{L}(p, n)}^2 = \norm{\nabla_n \mathcal{L}(p, n)}^2 + \norm{\nabla_p \mathcal{L}(p, n)}^2 = \norm{H(p)n}^2 + R(p, n), \text{ with} \\
			0 \leq R(p, n) \leq\frac{\tau - m}{4} c_H^2 \norm{n}^4.
		\end{gathered}
	\end{equation*}
	Using $\norm{H(p)n}^2 \geq \lambda_{\min, \mathcal{K}}^2 \norm{n}^2$, for $0 < \norm{n} \leq \rho$ we get
	\begin{equation*}
		R(p, n) \leq \frac{\tau - m}{4} c_H^2 \norm{n}^4 \leq \frac{(\tau - m) c_H^2 \rho^2}{4 \lambda_{\min, \mathcal{K}}^2} \norm{H(p)n}^2.
	\end{equation*}
	Choosing $\rho > 0$ small enough so that $(\tau - m) c_H^2 \rho^2 / (4 \lambda_{\min, \mathcal{K}}^2) \leq 1/2$ yields
	\begin{equation} \label{eq:quadratic_gradient_bound}
		\norm{H(p)n}^2 \leq \norm{\nabla \mathcal{L}(p, n)}^2 \leq \frac{3}{2} \norm{H(p)n}^2.
	\end{equation}
	Furthermore, for any unit vector $u$ we have
	\begin{equation} \label{eq:quadratic_hessian_bound}
		\lambda_{\min, \mathcal{K}} \leq u^T H(p)u \leq \lambda_{\max, \mathcal{K}}, \text{ and so }
		\lambda_{\min, \mathcal{K}} \norm{n}^2 \leq n^T H(p)n \leq \lambda_{\max, \mathcal{K}} \norm{n}^2.
	\end{equation}
	Using the bounds in \cref{eq:quadratic_gradient_bound,eq:quadratic_hessian_bound}, we get
	\begin{equation*}
		\beta(p, n)
		= 2 \frac{n^T H(p)n}{\norm{\nabla \mathcal{L}(p, n)}^2} \in \left[\frac{4\lambda_{\min, \mathcal{K}}}{3 \lambda_{\max, \mathcal{K}}^2}, \frac{2\lambda_{\max, \mathcal{K}}}{\lambda_{\min, \mathcal{K}}^2}\right].
	\end{equation*}
	Since each eigenvalue $\lambda_i(p) \in [\lambda_{\min, \mathcal{K}}, \lambda_{\max, \mathcal{K}}]$, there exists $c > 0$ such that, for all $i$,
	\begin{equation} \label{eq:quadratic_polyak_normal_step_bound}
		c \leq \beta(p, n)\lambda_i(p) \leq c^{-1},
	\end{equation}
	whenever $0 < \norm{n} \leq \rho$ and $p \in \mathcal{K}$.

	Now, we prove the two inequalities in \cref{eq:quadratic_scale_separation} separately.
	\begin{itemize}[label = {--}, leftmargin = 20pt]
		\item[$\bm{n}$.]
			From the update rule, $n^+ - n = -\beta(p, n) H(p)n$, so
			\begin{equation*}
				\norm{n^+ - n} = \beta(p, n) \norm{H(p) n}.
			\end{equation*}
			Using \cref{eq:quadratic_polyak_normal_step_bound} and $\lambda_{\min, \mathcal{K}} \norm{n} \leq \norm{H(p)n} \leq \lambda_{\max, \mathcal{K}} \norm{n}$, we obtain
			\begin{equation*}
				c_1 \norm{n} \leq \norm{n^+ - n} \leq c_2 \norm{n}
			\end{equation*}
			for suitable $0 < c_1 \leq c_2 < \infty$, proving the first two inequalities in \cref{eq:quadratic_scale_separation}.
		\item[$\bm{p}$.]
			By \cref{eq:quadratic_tangential_gradient_bound}
			\begin{equation*}
				\norm{\nabla_p \mathcal{L}(p, n)} \leq \frac{\sqrt{\tau - m} c_H}{2} \norm{n}^2.
			\end{equation*}
			Since $p^+ - p = -\beta(p, n)\nabla_p \mathcal{L}(p, n)$ and $\beta(p, n)$ is uniformly bounded above on $\mathcal{K} \times \{0 < \norm{n} \leq \rho\}$, we get
			\begin{equation*}
				\norm{p^+ - p} \leq c_3 \norm{n}^2
			\end{equation*}
			for some $c_3 > 0$, completing the proof of \cref{eq:quadratic_scale_separation}.
	\end{itemize}
\end{proof}

\subsubsection{Tangential drift along \texorpdfstring{$\partial_p \log{\lambda_1}$}{subgradient log lambda1}} \label{subsubsec:quadratic_tangential_drift_proof}

\begin{proof}[Proof of \Cref{thm:quadratic_tangential_drift}] \hypertarget{proof:quadratic_tangential_drift}
	By \cref{eq:quadratic_polyak_update} and \cref{eq:quadratic_loss_gradients},
	\begin{equation*}
		p_{k + 1} - p_k = -\beta(p_k, n_k) \nabla_p \mathcal{L}(p_k, n_k), \text{ with }
		\nabla_p \mathcal{L}(p_k, n_k) = \frac{1}{2}s_k^2 \nabla_p \left(u_k^T H(p_k) u_k\right).
	\end{equation*}
	Moreover, by \cref{eq:quadratic_polyak_step_size},
	\begin{equation*}
		\beta(p_k, n_k) = 2\frac{u_k^T H(p_k)u_k}{u_k^T H(p_k)^2u_k} + \bigo{s_k^2}.
	\end{equation*}
	Therefore
	\begin{equation} \label{eq:quadratic_tangential_update_expansion}
		p_{k + 1}
		= p_k - s_k^2 \frac{u_k^T H(p_k) u_k}{u_k^T H(p_k)^2 u_k} \nabla_p \left(u_k^T H(p_k) u_k\right) + \bigo{s_k^4}.
	\end{equation}
	The residual assumption is $\norm{\left(H(p_k) - \lambda_1(p_k)I_m\right)u_k} = \bigo{s_k}$, and by Cauchy-Schwarz it directly implies that the Rayleigh defect is of order $\bigo{s_k}$.
	For the same reason, since $H(p_k) + \lambda_1(p_k)I_m$ is uniformly bounded on $\mathcal{K}$, we can write
	\begin{equation*}
		u_k^T H(p_k)u_k - \lambda_1(p_k)
		= \bigo{s_k}, \text{ and } u_k^T H(p_k)^2u_k
		= \lambda_1(p_k)^2 + \bigo{s_k}.
	\end{equation*}
	Moreover,
	\begin{equation*}
		\norm{\left(H(p_k) - \left(u_k^T H(p_k) u_k\right)I_m\right) u_k} \leq
		\norm{\left(H(p_k) - \lambda_1(p_k) I_m\right) u_k} + \abs{\lambda_1(p_k) - u_k^T H(p_k) u_k}
		= \bigo{s_k},
	\end{equation*}
	which implies
	\begin{equation*}
		u_k^T H(p_k)^2u_k - \left(u_k^T H(p_k)u_k\right)^2
		= u_k^T \left(H(p_k) - \left(u_k^T H(p_k) u_k\right) I_m\right)^2 u_k
		= \bigo{s_k^2}.
	\end{equation*}
	The previous results, together with the fact that $u_k^T H(p_k)u_k$ and $u_k^T H(p_k)^2u_k$ are uniformly bounded away from zero on $\mathcal{K}$, allow us to conclude
	\begin{equation*}
		\frac{u_k^T H(p_k)u_k}{u_k^T H(p_k)^2 u_k} = \frac{1}{u_k^T H(p_k)u_k} + \bigo{s_k^2},
	\end{equation*}
	and, substituting this into \cref{eq:quadratic_tangential_update_expansion} gives the result in \cref{eq:quadratic_rayleigh_branch_drift}:
	\begin{equation*}
		p_{k + 1} = p_k - s_k^2 \nabla_p \log{\left(u_k^T H(p_k) u_k\right)} + \bigo{s_k^4}.
	\end{equation*}

	We now prove propagation of the residual condition.
	The full normal update satisfies
	\begin{equation*}
		n_{k + 1}
		= s_k \left(u_k - 2\frac{u_k^T H(p_k) u_k}{u_k^T H(p_k)^2 u_k}H(p_k)u_k\right) + \bigo{s_k^3}.
	\end{equation*}
	Analogously as in \Cref{lem:fixed_hessian_norm_and_alignment}, the vector in parentheses has norm $1$, hence by reverse triangle inequality $s_{k + 1} = s_k + \bigo{s_k^3}$.
	Using again $\norm{\left(H(p_k) - \lambda_1(p_k) I_m\right) u_k} = \bigo{s_k}$, by the same strategy used above to control the Rayleigh defect with Cauchy-Schwarz, we can show that
	\begin{equation*}
		1 - \frac{\left(u_k^T H(p_k)u_k\right)\lambda_1(p_k)}{u_k^T H(p_k)^2u_k} = \bigo{s_k},
	\end{equation*}
	and as a consequence
	\begin{equation*}
		\begin{aligned}
			\frac{n_{k + 1}}{s_k} + u_k & = 2\left(1 - \frac{\left(u_k^T H(p_k)u_k\right)\lambda_1(p_k)}{u_k^T H(p_k)^2u_k}\right)u_k +\\
			&\blankeq - 2\frac{u_k^T H(p_k)u_k}{u_k^T H(p_k)^2u_k}\left(H(p_k) - \lambda_1(p_k) I_m\right) u_k + \bigo{s_k^2} = \bigo{s_k}.
		\end{aligned}
	\end{equation*}
	Since $s_{k + 1} = s_k + \bigo{s_k^3}$, this gives $u_{k + 1} = -u_k + \bigo{s_k}$.
	Therefore
	\begin{equation*}
		\begin{aligned}
			\norm{\left(H(p_k) - \lambda_1(p_k) I_m\right) u_{k + 1}} &\leq \norm{\left(H(p_k) - \lambda_1(p_k) I_m\right) u_k} + \\
			& \blankeq + \norm{H(p_k) - \lambda_1(p_k) I_m} \norm{u_{k + 1} + u_k} = \bigo{s_k}.
		\end{aligned}
	\end{equation*}
	By \Cref{thm:quadratic_scale_separation}, $p_{k + 1} - p_k = \bigo{s_k^2}$.
	Since $H$ and $\lambda_1$ are Lipschitz on a fixed compact neighborhood of $\mathcal{K}$ containing $p_{k + 1}$ for sufficiently small $s_k$,
	\begin{equation*}
		H(p_{k + 1}) - H(p_k) = \bigo{s_k^2}, \
		\lambda_1(p_{k + 1}) - \lambda_1(p_k) = \bigo{s_k^2}.
	\end{equation*}
	Combining the previous estimates gives \cref{eq:quadratic_residual_alignment_update}:
	\begin{equation*}
		\norm{\left(H(p_{k + 1}) - \lambda_1(p_{k + 1}) I_m\right) u_{k + 1}}
		= \bigo{s_k}
		= \bigo{s_{k + 1}}.
	\end{equation*}

	It remains to prove the subgradient interpretation when the top branch is resolved.
	If $r_k = m$, set $w_k \coloneqq u_k$.
	If $r_k < m$ and $\lambda_1(p_k) - \lambda_{r_k + 1}(p_k) \geq \delta > 0$ independently of $s_k$, then
	\begin{equation*}
		\dist{u_k}{E_1(p_k) \cap \mathbb{S}^{m - 1}} \leq c_{\delta}\norm{\left(H(p_k) - \lambda_1(p_k) I_m\right) u_k}
		= \bigo{s_k}.
	\end{equation*}
	In this case, we choose $w_k \in E_1(p_k) \cap \mathbb{S}^{m - 1}$ such that $\norm{u_k - w_k} = \bigo{s_k}$.
	In both cases, we have $\norm{u_k - w_k} = \bigo{s_k}$, with $w_k \in E_1(p_k) \cap \mathbb{S}^{m - 1}$.

	Now, since $u_k$ and $w_k$ have unit norm, $w_k^T (u_k - w_k) = \bigo{s_k^2}$.
	Hence, using $H(p_k)w_k = \lambda_1(p_k)w_k$ and the uniform bounds on $H$,
	\begin{equation*}
		u_k^T H(p_k)u_k = \lambda_1(p_k) + \bigo{s_k^2}, \
		\nabla_p\left(u_k^T H(p_k)u_k\right) = \nabla_p \left(w_k^T H(p_k)w_k\right) + \bigo{s_k}.
	\end{equation*}
	Since $\lambda_1(p_k)$ is uniformly bounded away from zero on $\mathcal{K}$, this gives
	\begin{equation} \label{eq:quadratic_top_branch_gradient}
		\nabla_p \log{\left(u_k^T H(p_k) u_k\right)}
		= \frac{\nabla_p \left(w_k^T H(p_k) w_k\right)}{\lambda_1(p_k)} + \bigo{s_k}.
	\end{equation}
	Since $w_k \in E_1(p_k) \cap \mathbb{S}^{m - 1}$, we have $\lambda_1(p_k) = w_k^T H(p_k) w_k$.
	Since $H(p)$ is symmetric positive definite, by definition of the operator norm, we have that, $\forall p \in \real^{\tau - m}$, by Cauchy-Schwarz,
	\begin{equation*}
		\lambda_1(p)
		= \max_{w \in \mathbb{S}^{m - 1}} \norm{H(p) w} \geq \norm{H(p) w_k} \geq w_k^T H(p) w_k.
	\end{equation*}
	Thus the $C^2$ function $p \mapsto w_k^T H(p) w_k$ supports $p \mapsto \lambda_1(p)$ from below at $p = p_k$.
	Therefore, for any $h \in \real^{\tau - m}$ with $\norm{h} \ll 1$,
	\begin{equation*}
		\lambda_1(p_k + h) \geq \lambda_1(p_k) + \nabla_p \left(w_k^T H(p_k)w_k\right)^T h + \smallo{\norm{h}},
	\end{equation*}
	which, by the definition of subgradient \cite[Definition 8.3]{variational_rockafellar_1998}, gives $\nabla_p \left(w_k^T H(p_k)w_k\right) \in \partial_p \lambda_1(p_k)$.
	Since $\lambda_1(p_k) > 0$ on $\mathcal{K}$, the chain rule for the logarithm gives
	\begin{equation*}
		\partial_p \log{\lambda_1(p_k)} = \frac{\partial_p \lambda_1(p_k)}{\lambda_1(p_k)} \implies
		\zeta_k \coloneqq \frac{\nabla_p \left(w_k^T H(p_k)w_k\right)}{\lambda_1(p_k)} \in \partial_p \log{\lambda_1(p_k)}.
	\end{equation*}
	Substituting \cref{eq:quadratic_top_branch_gradient} into \cref{eq:quadratic_rayleigh_branch_drift} gives the thesis:
	\begin{equation*}
		p_{k + 1}
		= p_k - s_k^2 \left(\zeta_k + \bigo{s_k}\right) + \bigo{s_k^4}
		= p_k - s_k^2 \zeta_k + \bigo{s_k^3}, \text{ with } \zeta_k \in \partial_p \log{\lambda_1(p_k)}.
	\end{equation*}
	If $r_k = 1$, then $\lambda_1$ is $C^2$ near $p_k$ and the subgradient above is the usual gradient, giving \cref{eq:quadratic_sharpness_gradient_drift}.
	With the input bound and $\delta$ fixed, all constants are uniform for $p_k \in \mathcal{K}$ by compactness, the spectral bounds, and bounds on the first derivatives of $H$ in a fixed compact neighborhood of $\mathcal{K}$.
\end{proof}

\subsubsection{Finite entrance into residual alignment} \label{subsubsec:quadratic_residual_alignment_entry_proof}

\begin{proof}[Proof of \Cref{thm:quadratic_residual_alignment_entry}] \hypertarget{proof:quadratic_residual_alignment_entry}
	Fix $p_0, u_0$ and set $P_0 \coloneqq \Pi_{E_1(p_0)}$.
	All constants below may depend on these fixed initial data, and the admissible radius is chosen after $p_0$ and $u_0$.
	If $P_0 = I_m$, the residual is already zero at $k_0 = 0$.
	Hence we may assume $P_0 \neq I_m$.
	For a unit norm vector $u$, define the frozen map from \Cref{lem:fixed_hessian_norm_and_alignment} by
	\begin{equation*}
		Q_{p_0}(u) \coloneqq u - 2\frac{u^T H(p_0)u}{u^T H(p_0)^2u}H(p_0)u.
	\end{equation*}
	The spectral bounds give
	\begin{equation*}
		\lambda_m(p_0) u^T H(p_0)u \leq u^T H(p_0)^2u \leq \lambda_1(p_0) u^T H(p_0)u.
	\end{equation*}
	Hence its stepsize $\beta_0$ belongs to $[2 / \lambda_1(p_0), 2 / \lambda_m(p_0)]$.
	To compare the components inside and outside the top eigenspace, we take the inner product of the update with $v_i(p_0)$, which isolates the $i$-th component:
	\begin{equation*}
		v_i(p_0)^T Q_{p_0}(u) = (1 - \beta_0\lambda_i(p_0))v_i(p_0)^T u.
	\end{equation*}
	For every $\beta_0$ in this interval and $\lambda_i(p_0) < \lambda_1(p_0)$,
	\begin{equation*}
		\frac{\abs{1 - \beta_0\lambda_i(p_0)}}{\abs{1 - \beta_0\lambda_1(p_0)}} < 1, \text{ where } \abs{1 - \beta_0\lambda_1(p_0)} = \beta_0 \lambda_1(p_0) - 1 \geq 1.
	\end{equation*}
	Since $P_0 Q_{p_0}(u) = (1 - \beta_0\lambda_1(p_0))P_0 u$, the top projection cannot decrease in norm.
	When $P_0 u \neq 0$, each component outside the top eigenspace contracts relative to the top projection:
	\begin{equation*}
		\frac{\abs{v_i(p_0)^T Q_{p_0}(u)}}{\norm{P_0 Q_{p_0}(u)}} = \frac{\abs{1 - \beta_0\lambda_i(p_0)}}{\abs{1 - \beta_0\lambda_1(p_0)}}\frac{\abs{v_i(p_0)^T u}}{\norm{P_0 u}}.
	\end{equation*}
	By continuity and compactness of the interval, there exists a common $q \in (0, 1)$ such that
	\begin{equation*}
		\frac{\norm{(I_m - P_0)Q_{p_0}(u)}}{\norm{P_0 Q_{p_0}(u)}} \leq q \frac{\norm{(I_m - P_0)u}}{\norm{P_0 u}}, \text{ and } \norm{P_0 Q_{p_0}(u)} \geq \norm{P_0 u}.
	\end{equation*}
	Set $k_0 \coloneqq \ceil{\log{(1 / s_0)} / \abs{\log{q}}}$.
	The full quadratic update gives
	\begin{equation*}
		\begin{aligned}
			s_{j + 1}^2 & = s_j^2 - 2\beta(p_j, n_j)n_j^T H(p_j)n_j + \beta(p_j, n_j)^2\norm{H(p_j)n_j}^2 =  \\
			& = s_j^2 - \beta(p_j, n_j)^2\norm{\nabla_p \mathcal{L}(p_j, n_j)}^2 = s_j^2 - \norm{p_{j + 1} - p_j}^2.
		\end{aligned}
		\implies s_j \leq s_0.
	\end{equation*}
	On a fixed compact neighborhood of $p_0$, \Cref{thm:quadratic_scale_separation} gives $\norm{p_{i + 1} - p_i} \leq C s_i^2$, hence
	\begin{equation*}
		\begin{aligned}
			\norm{p_j - p_0} & \leq \sum_{i = 0}^{j - 1}\norm{p_{i + 1} - p_i} \leq C j s_0^2, \ 0 \leq s_0^2 - s_j^2 & = \sum_{i = 0}^{j - 1}\norm{p_{i + 1} - p_i}^2 \leq C^2 j s_0^4.
		\end{aligned}
	\end{equation*}
	Since $k_0 s_0^2 \to 0$, these bounds keep $p_j$ in that neighborhood and give $s_j = s_0\left(1 + \bigo{j s_0^2}\right)$, uniformly for $j \leq k_0$.
	The one-step identity also gives $s_j / s_{j + 1} = 1 + \bigo{s_0^2}$.
	Dividing the normal update by $s_{j + 1}$ and using \cref{eq:quadratic_polyak_step_size} and local Lipschitz continuity of $H$, we obtain
	\begin{equation*}
		u_{j + 1} = Q_{p_j}(u_j) + \bigo{s_0^2} = Q_{p_0}(u_j) + \bigo{(j + 1)s_0^2}.
	\end{equation*}
	The accumulated projection error is $\bigo{k_0^2 s_0^2} = \smallo{1}$, so $\norm{P_0 u_j} \geq \frac{1}{2}\norm{P_0 u_0} > 0$ for sufficiently small $s_0$.
	Hence the contraction estimate gives
	\begin{equation*}
		\frac{\norm{(I_m - P_0)u_{j + 1}}}{\norm{P_0 u_{j + 1}}} \leq q \frac{\norm{(I_m - P_0)u_j}}{\norm{P_0 u_j}} + C(j + 1)s_0^2,
	\end{equation*}
	and, by iteration,
	\begin{equation*}
		\frac{\norm{(I_m - P_0)u_{k_0}}}{\norm{P_0 u_{k_0}}} \leq q^{k_0} \frac{\norm{(I_m - P_0)u_0}}{\norm{P_0 u_0}} + C k_0 s_0^2 = \bigo{s_0}.
	\end{equation*}
	By Lipschitz continuity of $H$ and $\lambda_1$ near $p_0$, the claim follows:
	\begin{equation*}
		\norm{\left(H(p_{k_0}) - \lambda_1(p_{k_0})I_m\right)n_{k_0}} \leq C s_{k_0}\left(\norm{(I_m - P_0)u_{k_0}} + k_0 s_0^2\right) = \bigo{s_{k_0}^2}. \qedhere
	\end{equation*}
\end{proof}

\section{Interpolation geometry of neural networks} \label{sec:nonlinear_morse_bott_structure}

In this appendix, we show that, near a regular manifold of interpolating parameters, a smooth neural network loss admits a quadratic Morse-Bott expansion in suitable \textit{tubular coordinates}.
This local reduction sets the geometric foundation for the exact and non-linear dynamics studied in \Cref{sec:quadratic_morse_bott_dynamics,sec:morse_bott_dynamics}.

\subsection{Network notation} \label{subsec:neural_network_setup}

We begin by introducing the notation used to describe the networks and their training losses.
\begin{definition}[Fully connected neural network] \label{def:fully_connected_network}
	Let $L \in \integer_{> 0}$, $\bm{n} \coloneqq (n_0, \dots, n_{L + 1}) \in \integer_{> 0}^{L + 2}$ and $\sigma: \real \to \real$.
	A depth-$(L + 1)$ neural network ($L$ is the number of hidden layers) with layer widths $\bm{n}$ and activation function $\sigma$, is a map $z^{(L + 1)}$ defined as follows: $z^{(l)}: \real^{n_0} \to \real^{n_l}, \forall l \in [L + 1]$, and $\forall x \in \real^{n_0}$,
	\begin{equation} \label{eq:fully_connected_network_layers}
		z^{(1)}(x)
		= W^{(1)} x + b^{(1)}, \ z^{(l)}(x)
		= W^{(l)} \sigma(z^{(l - 1)}(x)) + b^{(l)}, \text{ for } l
		= 2, \dots, L + 1,
	\end{equation}
	where $\sigma$ is applied component-wise, and for any $l \in [L + 1]$, $W^{(l)} \in \real^{n_l \times n_{l - 1}}$, $b^{(l)} \in \real^{n_l}$ denote weight matrices and bias vectors.
	Let $\theta \in \real^{\tau}$, with $\tau = \sum_{l = 1}^{L + 1} n_l (n_{l - 1} + 1)$, be the flattened concatenation of all the network parameters.
	The network can be seen as a parametric map from $\real^{\tau} \times \real^{n_0}$ to $\real^{n_{L + 1}}$, i.e., $f_{\theta}: \real^{n_0} \to \real^{n_{L + 1}}$, $f_{\theta}(x) \coloneqq z^{(L + 1)}(x)$.
\end{definition}
We focus our attention on the optimization of neural networks through gradient descent (GD) with adaptive stepsize, in a regression setting with training set $\train \coloneqq \{(x_{i}, y_{i})\}_{i = 1}^N$.
For the mean squared loss, at points where the gradient exists, the update at iteration $k$ is
\begin{equation} \label{eq:mean_squared_loss_gradient_descent}
	\theta_{k + 1}
	= \theta_k - \eta_k \nabla_{\theta} \mathcal{L}(\theta_k), \text{ with } \mathcal{L}: \real^{\tau} \to \real, \ \mathcal{L}(\theta)
	= \frac{1}{2N} \sum_{i = 1}^N \norm{f_{\theta}(x_i) - y_i}^2.
\end{equation}

\subsection{Neural network loss as a squared residual}

Recall the training loss in \cref{eq:mean_squared_loss_gradient_descent}.
Throughout this section we assume $\sigma \in C^3(\real, \real)$.
Let us define $m \coloneqq N n_{L + 1}$, where $n_{L + 1}$ is the output dimension of the network and $N$ is the number of training samples.
By stacking the residuals, we can write
\begin{equation} \label{eq:network_residual_vector}
	r(\theta) \coloneqq \frac{1}{\sqrt{N}}
	\begin{pmatrix}
		f_{\theta}(x_1) - y_1 \\
		\vdots \\
		f_{\theta}(x_N) - y_N
	\end{pmatrix} \in \real^m, \
	\mathcal{L}(\theta) = \frac{1}{2} \norm{r(\theta)}^2.
\end{equation}
If $\sigma \in C^3(\real, \real)$, then $r \in C^3(\real^{\tau}, \real^m)$ and hence $\mathcal{L}\in C^3(\real^{\tau}, \real)$.
Differentiating,
\begin{equation} \label{eq:residual_loss_gradient_hessian}
	\nabla_{\theta} \mathcal{L}(\theta) = J_r(\theta)^{T} r(\theta), \
	\nabla_{\theta}^2 \mathcal{L}(\theta)
	= J_r(\theta)^{T} J_r(\theta) + \sum_{j = 1}^m r_j(\theta) \nabla_{\theta}^2 r_j(\theta),
\end{equation}
where $J_r(\theta) \in \real^{m \times \tau}$ is the Jacobian of $r$.

\subsection{Interpolation manifold and tubular coordinates} \label{subsec:interpolation_tubular_coordinates}

Throughout this appendix we focus on the \textit{interpolation regime}, where the training loss can reach zero.
Define the set of interpolating parameters
\begin{equation} \label{eq:interpolation_manifold}
	\mathcal{M}_0 \coloneqq \{\theta \in \real^{\tau} \mid \mathcal{L}(\theta)
	= 0\}
	= \{\theta \in \real^{\tau} \mid r(\theta)
	= \bm{0}_m\} \subset \{\theta \in \real^{\tau} \mid \nabla \mathcal{L}(\theta)
	= \bm{0}_\tau\}.
\end{equation}
Since the analysis below is local along $\mathcal{M}_0$, we fix a compact set $K \subset \real^{\tau}$, thought of as a region of parameter space that a given trajectory visits.
\begin{assumption}[Interpolation and nondegeneracy] \label{ass:interpolation_nondegeneracy}
$\blankeq$
	\begin{enumerate}
		\item \textit{Interpolation.} $K \cap \mathcal{M}_0 \neq \emptyset$, i.e. $\exists \theta_\star \in K$ such that $\mathcal{L}(\theta_\star) = 0$.
		\item \textit{Local regularity.} For every $\theta_\star \in K \cap \mathcal{M}_0$ the Jacobian $J_r(\theta_\star)$ has full row rank $m$.
	\end{enumerate}
\end{assumption}

The local regularity condition in \Cref{ass:interpolation_nondegeneracy}
is equivalent to positive definiteness of the empirical tangent kernel
$J_r(\theta_\star)J_r(\theta_\star)^\top$
at every $\theta_\star\in K\cap\mathcal{M}_0$.

\begin{remark}
	The local regularity condition in \Cref{ass:interpolation_nondegeneracy} is not trivial.
	Full row rank can hold generically for suitable architectures and nondegenerate datasets, but is not automatic. We assume it locally near the interpolating parameters under consideration.
	For example, consider a depth-$2$ network, so that $L = 1$ and $\bm{n} = (n_0, n_1, 1)$, i.e.,
	\begin{equation*}
		f_\theta(x) = W^{(2)} \sigma(W^{(1)}x + b^{(1)}) + b^{(2)}.
	\end{equation*}
	Suppose $y_i = c$ for every $i \in [N]$, and $N > 1$, then $\theta_\star = (0, 0, 0, c)$ is interpolating.
	At $\theta_\star$, the derivatives of the residual map with respect to $W^{(1)}$ and $b^{(1)}$ are $0$, while each column corresponding to $W^{(2)}$ is $\sigma(0) N^{-1/2} \mathbf{1}_N$ and the column corresponding to $b^{(2)}$ is $N^{-1/2} \mathbf{1}_N \in \real^N$.
	Hence $\rank{J_r(\theta_\star)} = 1 < N = m$.

	By continuity and compactness, $J_r$ has full row rank in a sufficiently small neighborhood of $K \cap \mathcal{M}_0$.
\end{remark}
Under \Cref{ass:interpolation_nondegeneracy}, there exists an open neighborhood $A \subset \real^\tau$ of $K \cap \mathcal{M}_0$ such that $0$ is a regular value of $r|_A$.
By the Regular Level Set Theorem \cite[Corollary 5.14]{smooth_lee_2013}, $A \cap \mathcal{M}_0 = (r|_A)^{-1}(0)$ is a $C^3$ submanifold of $A$ of codimension $m$.
Without loss of generality (shrinking $K$ if needed) we assume that $K \cap \mathcal{M}_0$ is contained in a single connected component of $A \cap \mathcal{M}_0$, which we denote by $\mathcal{M}$.
Thus, $\mathcal{M}$ is only the local connected component of the regular zero set relevant to $K$, and all the constructions below are local on this manifold piece.
We write $\mathcal{K} \coloneqq K \cap \mathcal{M}$.
For every $\theta_\star \in \mathcal{M}$, we then have
\begin{equation} \label{eq:interpolation_tangent_normal_spaces}
	T_{\theta_\star} \mathcal{M} = \ker{J_r(\theta_\star)},
	N_{\theta_\star} \mathcal{M} = \ker{J_r(\theta_\star)}^{\perp}.
\end{equation}
Moreover, since $r(\theta_\star) = 0$, \cref{eq:residual_loss_gradient_hessian} gives
\begin{equation} \label{eq:interpolation_loss_hessian}
	\nabla^2 \mathcal{L}(\theta_\star) = J_r(\theta_\star)^T J_r(\theta_\star).
\end{equation}
\begin{definition}[Morse-Bott along a manifold] \label{def:morse_bott_loss}
	Let $\mathcal{L} \in C^2(\real^{\tau}, \real)$ and let $\mathcal{M} \subset \real^{\tau}$ be a connected $C^2$ submanifold.
	We say that $\mathcal{L}$ is Morse-Bott along $\mathcal{M}$ if $\mathcal{M} \subset \{\theta \mid \nabla \mathcal{L}(\theta) = 0\}$ and
	\begin{equation} \label{eq:morse_bott_hessian_kernel}
		\ker{\nabla^2 \mathcal{L}(\theta_\star)}
		= T_{\theta_\star}\mathcal{M}, \ \forall \theta_\star \in \mathcal{M}.
	\end{equation}
	Equivalently, the restriction of $\nabla^2 \mathcal{L}(\theta_\star)$ to $N_{\theta_\star}\mathcal{M}$ is nondegenerate.
	If moreover $\mathcal{M}$ is a manifold of (local) minima of $\mathcal{L}$, then the restriction of $\nabla^2 \mathcal{L}(\theta_\star)$ to $N_{\theta_\star}\mathcal{M}$ is positive definite.
\end{definition}
\begin{remark}
	The definition above is a local version of the standard notion (see \cite[Definition 2.41]{morse_nicolaescu_2011}): compactness is not imposed since the analysis is localized on the compact set $\mathcal{K}$, and we focus only on the connected component of global minima visited by the trajectory, rather than on the entire critical set of $\mathcal{L}$.
\end{remark}
For $\theta_\star \in \mathcal{M}$, one has $\nabla \mathcal{L}(\theta_\star) = 0$, and \cref{eq:interpolation_loss_hessian,eq:interpolation_tangent_normal_spaces} give
\begin{equation*}
	\ker{\nabla^2 \mathcal{L}(\theta_\star)} = \ker{J_r(\theta_\star)} = T_{\theta_\star}\mathcal{M}.
\end{equation*}
Moreover, $\nabla^2 \mathcal{L}(\theta_\star) = J_r(\theta_\star)^T J_r(\theta_\star)$ is positive definite on $N_{\theta_\star}\mathcal{M}$.
Thus $\mathcal{L}$ is Morse-Bott along $\mathcal{M}$.
At this point the specific squared-residual structure has served only to identify the local zero-loss manifold and to verify the Morse-Bott property along it.
The next proposition is a local consequence of this Morse-Bott structure.
\begin{proposition}[Quadratic Morse-Bott expansion in tubular coordinates] \label{prop:morse_bott_quadratic_expansion}
	Under \Cref{ass:interpolation_nondegeneracy}, there exists a radius $\rho > 0$ and an open tubular neighborhood
	\begin{equation*}
		U \subset \left\{\theta \in \real^\tau \mid \dist{\theta}{\mathcal{M}} < \rho\right\}
	\end{equation*}
	of $\mathcal{K}$ in $\real^\tau$ such that
	\begin{itemize}[label = {--}]
		\item every $\theta \in U$ can be written uniquely as
			\begin{equation} \label{eq:interpolation_tubular_coordinates}
				\theta = p + n, \ p \in \mathcal{M}, \ n \in N_p\mathcal{M}, \text{ with } \norm{n} < \rho,
			\end{equation}
		\item for every $p \in \mathcal{K}$ and every $n \in N_p\mathcal{M}$ with $\norm{n} < \rho$, one has $p + n \in U$.
	\end{itemize}
	For every $p \in \mathcal{K}$, let $H(p): N_p\mathcal{M} \to N_p\mathcal{M}$ denote the restriction of $\nabla^2 \mathcal{L}(p)$ to the normal space.
	Then $H(p)$ is symmetric positive definite, and writing
	\begin{equation*}
		\mathcal{L}(p, n) \coloneqq \mathcal{L}(p + n), \ \nabla \mathcal{L}(p, n) \coloneqq \nabla \mathcal{L}(p + n), \ \forall p \in \mathcal{K}, \ n \in N_p\mathcal{M}, \text{ with } \norm{n} < \rho,
	\end{equation*}
	one has
	\begin{equation} \label{eq:morse_bott_quadratic_expansion}
		\begin{aligned}
			\mathcal{L}(p, n) & = \frac{1}{2} \left\langle n, H(p) n \right\rangle + \bigo{\norm{n}^3}, \\
			\Pi_{N_p\mathcal{M}} \nabla \mathcal{L}(p, n) & = H(p) n + \bigo{\norm{n}^2}, \\
			\Pi_{T_p\mathcal{M}} \nabla \mathcal{L}(p, n) & = \bigo{\norm{n}^2},
		\end{aligned}
	\end{equation}
	uniformly along $\mathcal{K}$, where $\Pi_{N_p\mathcal{M}}$ and $\Pi_{T_p\mathcal{M}}$ denote the orthogonal projections.
	Moreover, for every $p \in \mathcal{K}$ the nonzero spectrum of $\nabla^2 \mathcal{L}(p)$ coincides with the spectrum of $H(p)$.
\end{proposition}
\begin{proof}
	See \hyperlink{proof:morse_bott_quadratic_expansion}{proof of \Cref{prop:morse_bott_quadratic_expansion}} in \Cref{subsec:morse_bott_expansion_proof}.
\end{proof}
\begin{remark}
	For every $p \in \mathcal{M}$, let
	\begin{equation*}
		\lambda_1(p) \geq \dots \geq \lambda_m(p) > 0
	\end{equation*}
	denote the eigenvalues of $H(p)$ on $N_p\mathcal{M}$.
	Then
	\begin{equation*}
		\lambda_1(p) = \norm{H(p)}_{2} = \norm{\nabla^2 \mathcal{L}(p)}_{2} = \norm{J_r(p)}_{2}^2.
	\end{equation*}
	In the case of $r(\theta^{(1)}, \theta^{(2)}) = x_1 \theta^{(1)} \theta^{(2)} - y_1$, the normal Hessian at $p = (\theta^{(1)}, \theta^{(2)}) \in \mathcal{M}$ is multiplication by $x_1^2 ((\theta^{(1)})^2 + (\theta^{(2)})^2)$, and therefore $\lambda_1(p) = x_1^2 \norm{p}^2$.
\end{remark}
\Cref{prop:morse_bott_quadratic_expansion} identifies the local geometry near a regular interpolation manifold.
Indeed, in tubular coordinates $\theta = p + n$, with $p \in \mathcal{M}$ and $n \in N_p\mathcal{M}$, the leading-order behavior is governed by the intrinsic quadratic Morse-Bott family $\mathcal{L}(p, n) = \frac{1}{2} \langle H(p) n, n \rangle$.
Moreover, by the same result we have that, in a tubular neighborhood of the manifold of minima, the CDP stepsize in \cref{eq:cdp_step_size} with $\kappa = 0$ simplifies to
\begin{equation*}
	\frac{4 \mathcal{L}(p, n)}{\norm{\nabla \mathcal{L}(p, n)}^2}
	= 2 \frac{\langle H(p) n, n \rangle}{\norm{H(p) n}^2} + \bigo{\norm{n}},
\end{equation*}
so that assuming an alignment of the normal component $n$ with the top eigendirection of $H(p)$, the stepsize is approximately $2 / \lambda_1(p)$, and therefore we retrieve once again the EoS scaling of \cite{eos_cohen_2021}.

This is the local bridge from MSE neural networks to the quadratic Morse-Bott framework.
The corresponding exact quadratic model in flat Euclidean product geometry is studied in \Cref{sec:quadratic_morse_bott_dynamics}.
The full non-linear setting, including the curved tubular geometry, is treated in \Cref{sec:morse_bott_dynamics}, with the damping and extra regularity needed for convergence\footnote{To apply the general convergence result of \Cref{sec:morse_bott_dynamics} (\Cref{thm:morse_bott_cdp_convergence}) to neural-network losses, one also needs $\sigma \in C^5$ and the spectral, critical-point, and trajectory-localization conditions of \Cref{ass:localized_sharpness_spectrum}.}.

\subsection{Proof of quadratic Morse-Bott expansion} \label{subsec:morse_bott_expansion_proof}

We report below the proof of the quadratic Morse-Bott expansion stated in \Cref{prop:morse_bott_quadratic_expansion}.
The argument combines the tubular neighborhood theorem with a Taylor expansion of the loss in the normal directions, and identifies the leading-order tangential and normal components of the gradient near the interpolation manifold.
\begin{proof}[Proof of \Cref{prop:morse_bott_quadratic_expansion}] \hypertarget{proof:morse_bott_quadratic_expansion}
	Choose an open set $\mathcal{A}$ with $\mathcal{K} \subset \mathcal{A} \subset \mathcal{M}$ and compact closure in $\mathcal{M}$, together with an open set $\mathcal{V}$ such that $\overline{\mathcal{A}} \subset \mathcal{V} \subset \mathcal{M}$.
	Since $\mathcal{M}$ is $C^3$, the normal bundle $N\mathcal{M}$ is a $C^2$ vector bundle over $\mathcal{M}$, by the construction in \cite[Corollary 10.36]{smooth_lee_2013}.
	Consider the restricted normal bundle
	\begin{equation*}
		N\mathcal{M}\big|_{\mathcal{V}} \coloneqq \bigsqcup_{p \in \mathcal{V}} N_p\mathcal{M}
		= \{(p,n) \in \mathcal{V} \times \real^\tau \mid n \in N_p\mathcal{M}\}.
	\end{equation*}
	By the Tubular Neighborhood Theorem \cite[Theorem 6.24]{smooth_lee_2013}, there exists an open neighborhood $\mathcal{U}$ of the zero section $\{(p,0) \mid p \in \mathcal{V}\}$ in $N\mathcal{M}\big|_{\mathcal{V}}$ such that
	\begin{equation*}
		\Phi(p, n) \coloneqq p + n
	\end{equation*}
	is a $C^2$ diffeomorphism from $\mathcal{U}$ onto an open neighborhood of $\mathcal{V}$ in $\real^\tau$.
	Since $\overline{\mathcal{A}}$ is compact and $\mathcal{U}$ is open around the zero section, choose $\rho > 0$ such that every normal vector of length less than $\rho$ based within distance $2 \rho$ of $\overline{\mathcal{A}}$ belongs to $\mathcal{U}$.
	In particular,
	\begin{equation*}
		\mathcal{U}_\rho
		= \left\{(p, n) \in N\mathcal{M}\big|_{\mathcal{A}} \midbar \norm{n} < \rho \right\}
		\subset \mathcal{U}.
	\end{equation*}
	Set $U \coloneqq \Phi(\mathcal{U}_\rho)$. Then $U$ is an open neighborhood of $\mathcal{A}$ in $\real^\tau$, every $\theta \in U$ admits a unique decomposition as in \cref{eq:interpolation_tubular_coordinates}, and for every $p \in \mathcal{K}$ and every $n \in N_p\mathcal{M}$ with $\norm{n} < \rho$, one has $p + n \in U$ because $\mathcal{K} \subset \mathcal{A}$.
	Indeed, if $\theta = p + n \in U$ also has a representation $\theta = q + v$ with $q \in \mathcal{M}$, $v \in N_q\mathcal{M}$, and $\norm{v} < \rho$, then $\norm{q - p} \leq \norm{n} + \norm{v} < 2\rho$ implies $(q, v) \in \mathcal{U}$.
	Both representations therefore belong to $\mathcal{U}$, where injectivity of $\Phi$ gives uniqueness.
	For $p \in \mathcal{A}$ and $n \in N_p\mathcal{M}$ with $\norm{n} < \rho$, define
	\begin{equation*}
		\widetilde{\mathcal{L}}(p, n) \coloneqq \mathcal{L}(\Phi(p, n)) = \mathcal{L}(p + n).
	\end{equation*}
	By \cref{eq:interpolation_loss_hessian,eq:interpolation_tangent_normal_spaces}, $\ker{\nabla^2 \mathcal{L}(p)} = T_p\mathcal{M}$ for every $p \in \mathcal{A}$.
	Since $\nabla^2 \mathcal{L}(p)$ is symmetric (self-adjoint), it leaves $N_p\mathcal{M} = (T_p\mathcal{M})^\perp$ invariant, and its restriction to $N_p\mathcal{M}$ is positive definite.
	This is precisely the operator $H(p)$.
	The spectral claim follows because $\ker{\nabla^2 \mathcal{L}(p)} = T_p\mathcal{M}$ and $\real^\tau = T_p\mathcal{M} \oplus N_p\mathcal{M}$.
	Since $\mathcal{L}$ is $C^3$ and $\overline{\mathcal{A}}$ is compact, Taylor's theorem in the ambient variable yields
	\begin{equation*}
		\widetilde{\mathcal{L}}(p, n)
		= \mathcal{L}(p) + \left\langle \nabla \mathcal{L}(p), n \right\rangle + \frac{1}{2} \left\langle n, \nabla^2 \mathcal{L}(p) n \right\rangle + \bigo{\norm{n}^3},
	\end{equation*}
	uniformly along $\mathcal{K}$. Since $\mathcal{L}(p) = 0$, $\nabla \mathcal{L}(p) = 0$, and $n \in N_p\mathcal{M}$, this gives
	\begin{equation*}
		\widetilde{\mathcal{L}}(p, n)
		= \frac{1}{2} \left\langle n, H(p) n \right\rangle + \bigo{\norm{n}^3}.
	\end{equation*}
	Applying Taylor's theorem to the ambient gradient gives
	\begin{equation*}
		\nabla \mathcal{L}(p + n)
		= \nabla \mathcal{L}(p) + \nabla^2 \mathcal{L}(p) n + \bigo{\norm{n}^2}
		= H(p) n + \bigo{\norm{n}^2},
	\end{equation*}
	uniformly along $\mathcal{K}$. Projecting onto $N_p\mathcal{M}$ and $T_p\mathcal{M}$ gives the second and third lines of \cref{eq:morse_bott_quadratic_expansion}.
\end{proof}

\section{Experimental details and additional results} \label{sec:experiments}

In \Cref{subsec:optimization_examples}, we illustrated the flattening effect of CDP stepsizes.
Here we discuss those experiments in more detail, along with additional experiments.
For the neural network experiments, the goal is not to reproduce the assumptions of \Cref{sec:morse_bott_dynamics} exactly, but to test whether the same qualitative phenomenon of sharpness reduction persists in practice.

\subsection{Student-teacher models} \label{subsec:student_teacher_details}

\begin{figure}[htb!]
	\centering
	\begin{subfigure}[t]{0.49\linewidth}
		\centering
		\includegraphics[width = \linewidth]{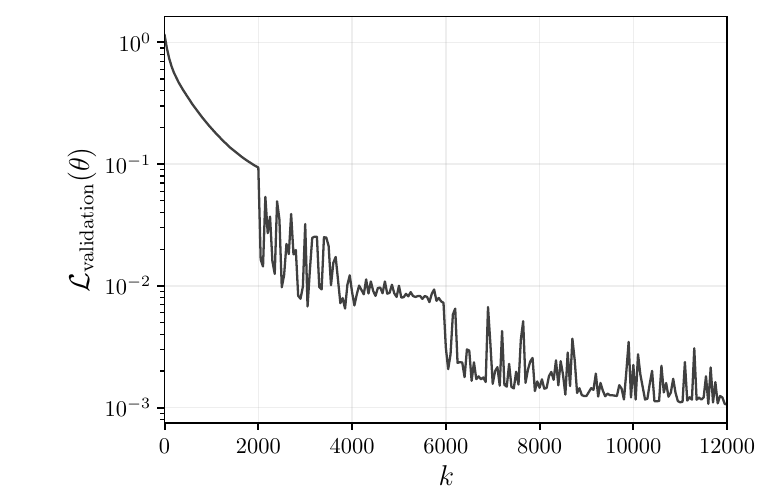}
		\phantomsubcaption
		\label{fig:residual_student_teacher_loss}
	\end{subfigure}
	\hfill
	\begin{subfigure}[t]{0.49\linewidth}
		\centering
		\includegraphics[width = \linewidth]{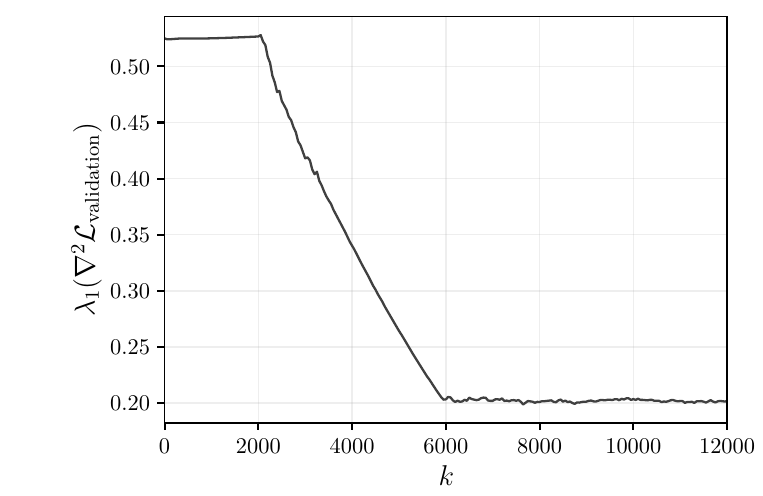}
		\phantomsubcaption
		\label{fig:residual_student_teacher_sharpness}
	\end{subfigure}
	\caption{Analog of \Cref{fig:student_teacher_loss_and_sharpness} for a residual ReLU student-teacher network with $4$ hidden layers, $N_\ell^{\mathrm{teacher}} = 32$ and $N_\ell^{\mathrm{student}} = 128$ for $\ell = 1, \dots, 4$, and residual branches scaled by $0.3$.
	The SGD learning rate is $\eta = 0.012$, and CDP stepsizes use $\kappa = 30$.
	As in the fully connected architecture, validation loss decreases in both phases, whereas sharpness decreases only after switching to CDP.}
	\label{fig:residual_student_teacher_loss_and_sharpness}
\end{figure}

For the fully connected models, we fix a depth $L \geq 1$, an input dimension $N_0 = N_0^{\mathrm{teacher}}$, an output dimension $N_{L + 1} = N_{L + 1}^{\mathrm{teacher}}$, and a sequence of teacher hidden widths $N_1^{\mathrm{teacher}}, \ldots, N_L^{\mathrm{teacher}}$, to produce labels from the teacher architecture
\begin{equation*}
	\begin{aligned}
		y^{\mathrm{teacher}}(x) & = W^{\mathrm{teacher}}_{L + 1}\varphi\left(W^{\mathrm{teacher}}_L \cdots \varphi\left(W^{\mathrm{teacher}}_1x\right)\cdots\right), & W^{\mathrm{teacher}}_\ell &\in \mathbb{R}^{N^{\mathrm{teacher}}_\ell \times N^{\mathrm{teacher}}_{\ell - 1}},
	\end{aligned}
\end{equation*}
where the non-linearity $\varphi: \mathbb{R}\to \mathbb{R}$ is applied coordinate-wise.
We initialize the teacher hidden weights in a standard way
\begin{equation*}
	W^{\mathrm{teacher}}_{\ell;ij} \sim \mathcal{N}(0, 1 / N^{\mathrm{teacher}}_{\ell - 1}), \text{ for } \ell = 1, \dots, L,
\end{equation*}
and generate training labels
\begin{equation*}
	\begin{aligned}
		x_i &\sim \mathcal{N}(0, I_{N_0}), & y_i & = y^{\mathrm{teacher}}(x_i), & i & = 1, \ldots, n.
	\end{aligned}
\end{equation*}
We train a student network $y^{\mathrm{student}}(x;\theta)$ with the same values of $N_0$, $N_{L + 1}$, and non-linearity $\varphi$, but with larger widths $N_\ell^{\mathrm{student}}\geq N_\ell^{\mathrm{teacher}}$ to fit the training data by minimizing the MSE
\begin{equation*}
	\mathcal{L}(\theta) = \frac{1}{2n}\sum_{i = 1}^n \norm{y_i - y^{\mathrm{student}}(x_i; \theta)}^2.
\end{equation*}
The minimum of the objective $\mathcal{L}$ is guaranteed to be $0$.
The ReLU architectures and optimizer settings are given in \Cref{fig:student_teacher_loss_and_sharpness,fig:residual_student_teacher_loss_and_sharpness}.

\begin{figure}[htb!]
	\centering
	\includegraphics[width = \textwidth]{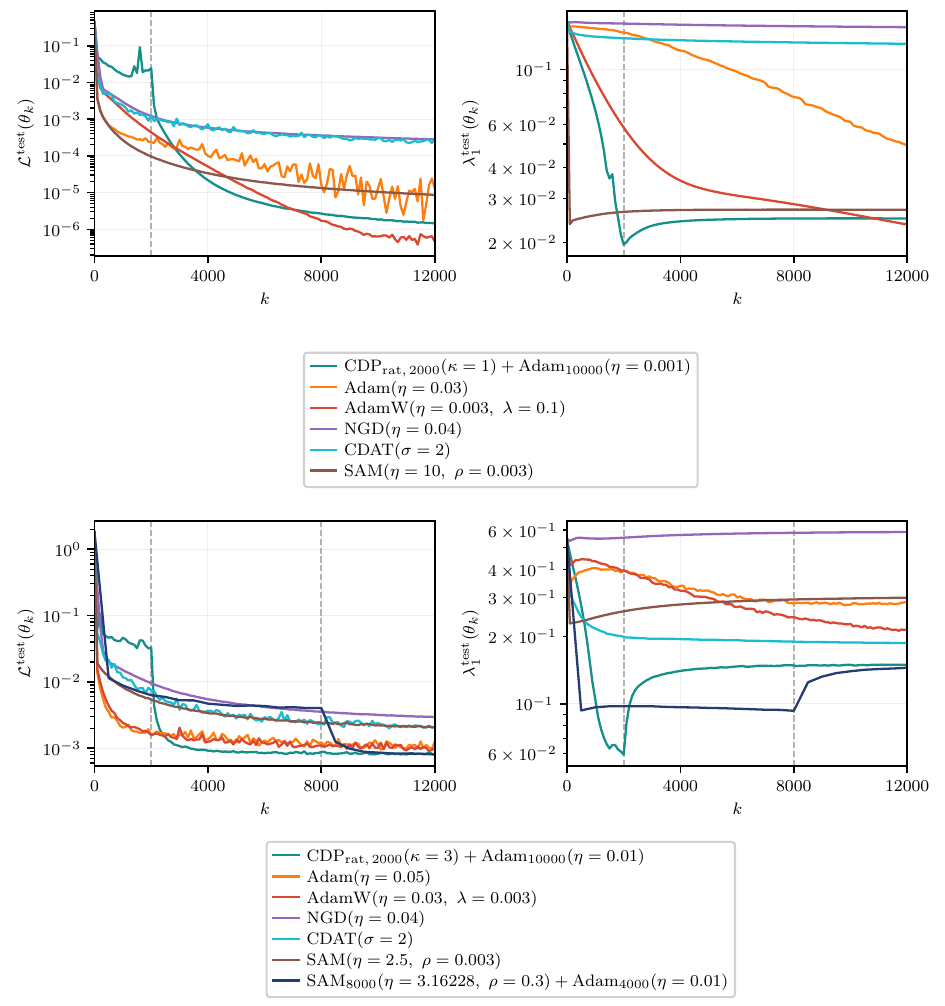}
	\caption{
		Optimizer comparison for the ReLU models in \Cref{fig:student_teacher_loss_and_sharpness,fig:residual_student_teacher_loss_and_sharpness}.
		Top, fully connected. Bottom, residual.
		Curves show mean test loss (left) and sharpness (right).
		CDP followed by Adam is competitive with state-of-the-art sharpness-aware methods, reducing test loss and sharpness in both models.
	}
	\label{fig:relu_optimizer_comparison}
\end{figure}

\textit{Optimizer comparison.}
All methods in \Cref{fig:relu_optimizer_comparison} use $4096$ training and $4096$ held-out samples, batch size $256$, and $12000$ updates (all hyperparameters are selected by mean validation loss).

\begin{figure}[htb!]
	\centering
	\includegraphics[width = 0.49\textwidth]{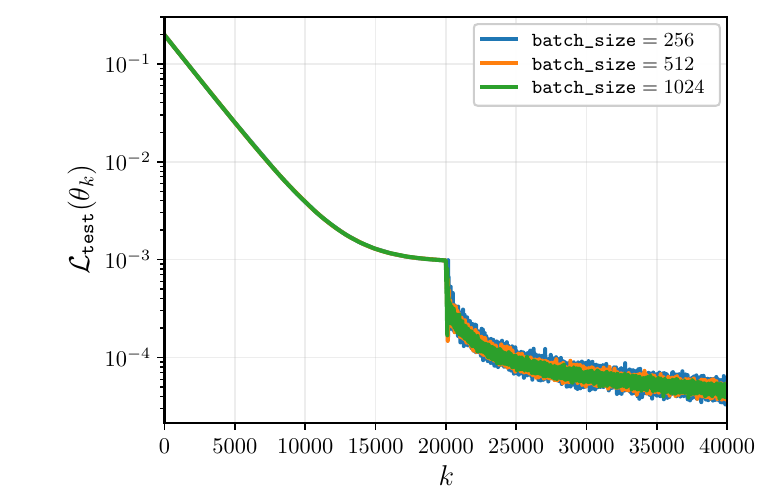}
	\includegraphics[width = 0.49\textwidth]{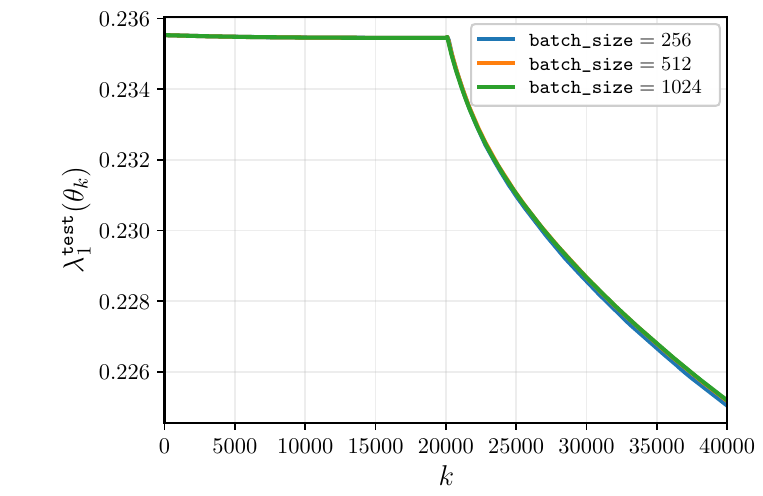}
	\caption{
		Test loss and test sharpness for the batch-size experiment.
		After a GD warmup, the CDP phase decreases $\lambda_1^{\rm{test}}(\theta_k)$ while keeping the test loss small, with similar behavior for batch sizes $256$, $512$, and $1024$.
	}
	\label{fig:deep_network_batch_size_comparison}
\end{figure}

\textit{Smooth non-linear networks.}
The experiments in \Cref{fig:deep_network_polyak_threshold_comparison,fig:deep_network_batch_size_comparison} use the same teacher-student construction with smooth $\tanh$ activations.
The data have $d_{\rm{in}} = 4$, with $4096$ training samples and $512$ test samples.
The teacher has width $4$, while the student is overparameterized with width $1024$, both in the same NTK parameterization.
Both networks have one hidden layer, no biases, and scalar output.
All runs use a GD warmup with stepsize $10^{-3}$, followed by a CDP phase.
The exponential damping coefficient is the reported $\kappa$ divided by the full training loss at the start of the CDP phase.
We monitor test loss and sharpness.

\textit{Switching threshold and damping.}
In \Cref{fig:deep_network_polyak_threshold_comparison}, we vary the training loss threshold $\tau$ at which we switch into the CDP regime, together with the damping parameter.
The three runs use the following values of $(\tau, \kappa)$: $(10^{-3}, 0.75)$, $(5 \times 10^{-3}, 0.5)$, and $(10^{-2}, 0.1)$.
This tests whether the flattening effect is already visible when CDP is activated at different distances from interpolation.
The same qualitative mechanism persists: the test loss keeps decreasing after the switch, and the runs combining earlier switching with weaker damping show a larger decrease in curvature.

\textit{Batch size.}
Using the same network and dataset, we vary the batch size in \Cref{fig:deep_network_batch_size_comparison}, keeping $\kappa = 0.75$.
Changing the batch size has only a mild effect on the final loss and sharpness, while the CDP phase consistently moves the solution toward smaller curvature.
These experiments suggest that sharpness reduction persists with minibatch gradients.

\subsection{LLM fine-tuning} \label{subsec:llm_fine_tuning_details}

We give details for the Qwen and GPT-2 comparisons in \Cref{subsubsec:pretrained_llm_experiments}.

\begin{figure}[htb!]
	\centering
	\includegraphics[width = 0.49\textwidth]{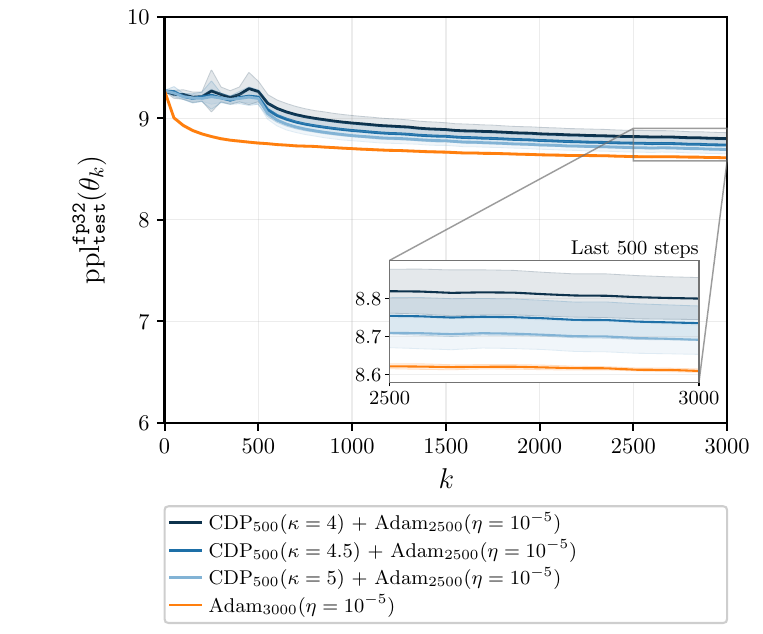}
	\includegraphics[width = 0.49\textwidth]{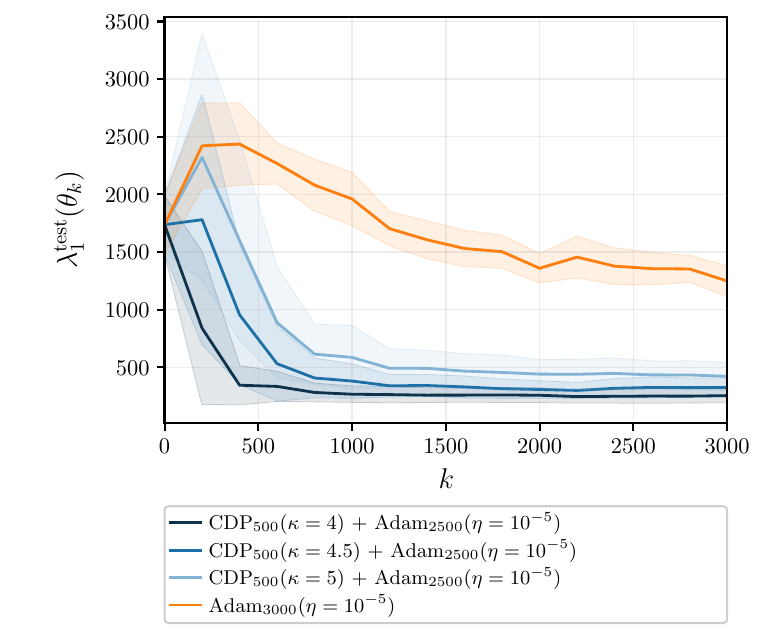}
	\caption{
		Analogous to \Cref{fig:qwen_sharpness_perplexity_sequence_512}, with training sequence length $128$ and test evaluation without a sliding window ($10$ seeds).
		The optimizer settings and the $500$-step CDP phase followed by $2500$ Adam steps are unchanged.
		Left: test perplexity, with a zoom on the final $500$ steps. Right: the top Hessian eigenvalue.
		The lower sharpness persists after the switch to Adam.
	}
	\label{fig:qwen_sharpness_perplexity_sequence_128}
\end{figure}

\textit{Qwen2.5-1.5B.}
The runs in \Cref{fig:qwen_sharpness_perplexity_sequence_512,fig:qwen_sharpness_perplexity_sequence_128} fine-tune \texttt{Qwen2.5-1.5B} on \texttt{WikiText-2}, with training sequence lengths $512$ and $128$, respectively.
We compare Adam with learning rate $10^{-5}$ and no weight decay against a two-phase procedure that first applies $500$ CDP steps and then switches to Adam for the remaining $2500$ steps.
The CDP damping parameter is varied over $\kappa \in \{4, 4.5, 5\}$.
For these LLM warmup experiments, the exponential damping coefficient is the reported $\kappa$ divided by the first training minibatch loss.
We report full-precision test perplexity and the top Hessian eigenvalue.
\Cref{fig:qwen_sharpness_perplexity_sequence_512} shows that the CDP-initialized runs consistently have much lower sharpness than Adam-only, while Adam-only reaches (slightly) lower perplexity.
The lower sharpness persists after the switch to Adam.

\textit{GPT-2 and the step cap.}
\Cref{tab:llm_sharpness_comparison} compares AdamW, CDP, and NGD on \texttt{GPT-2}.
The sensitivity to the step cap is consistent with the local nature of \Cref{thm:cdp_sharpness_dynamics_informal}: with the per-step movement capped at $0.1$ all three seeds show decreasing sharpness, whereas with a looser cap of $0.3$ or $1$ one of three seeds leaves the neighborhood of the pre-trained weights and $\lambda_1$ grows.

\subsection{Post-training quantization} \label{subsec:qwen_small_quantization}

\begin{figure}[tb!]
	\centering
	\includegraphics[width = 0.49\textwidth]{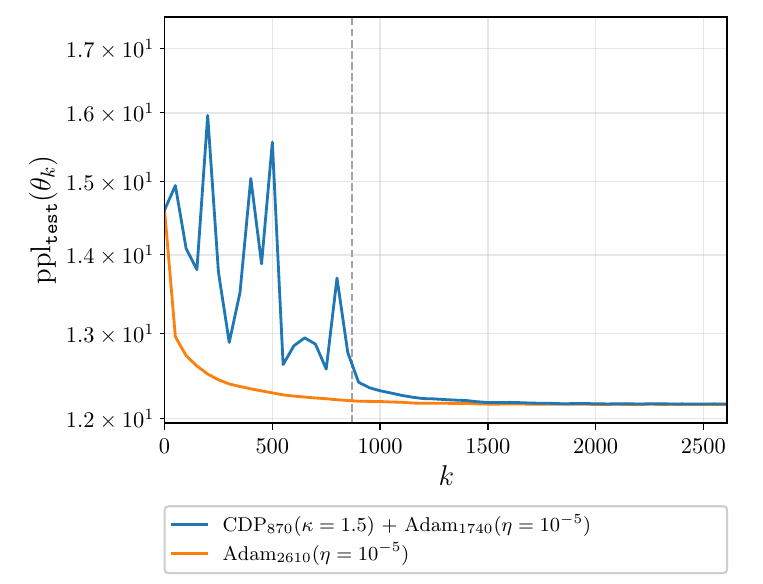}
	\includegraphics[width = 0.49\textwidth]{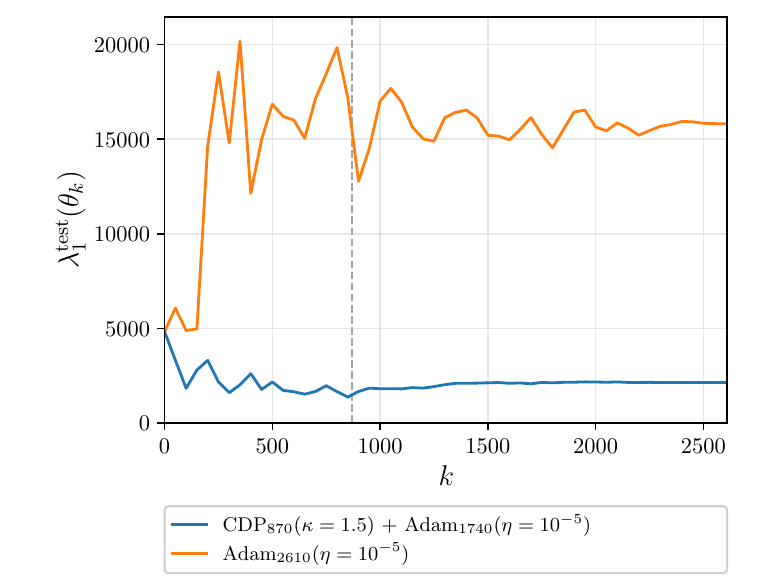}
	\caption{
		\texttt{Qwen2.5-0.5B} fine-tuned on \texttt{WikiText-2}.
		Left: test perplexity.
		Right: mean of the top Hessian eigenvalues on $16$ fixed test sequences, each estimated with $20$ Lanczos iterations in \texttt{fp32}.
		CDP followed by Adam reaches nearly the same final perplexity as Adam alone, with about $7$ times smaller final sharpness.
	}
	\label{fig:qwen_small_dynamics}
\end{figure}

Using \texttt{Qwen2.5-0.5B}, we test whether lower sharpness after fine-tuning accompanies improved robustness to weight quantization.
We compare $2610$ Adam steps with $870$ CDP steps followed by $1740$ Adam steps on \texttt{WikiText-2}.
Both runs use training batch size $8$ and \texttt{bf16}, with sequence length $1024$ for training and nonoverlapping test evaluation.
CDP uses exponential damping with $\kappa = 1.5$, and each Adam phase starts at learning rate $10^{-5}$ with linear decay.
\Cref{fig:qwen_small_dynamics} shows nearly identical final perplexities and a final mean batch-wise sharpness about $7$ times smaller for CDP + Adam.

\begin{table}[tbp!]
	\centering\footnotesize
	\caption{
		Final test perplexity and sharpness with the same initialization, data, precision, and $2610$-update budget (SAM is implemented using Adam as the base optimizer).
		Sharpness averages the top Hessian eigenvalues on $16$ fixed test sequences of length $1024$, each estimated with $20$ Lanczos iterations over all parameters in \texttt{fp32}.
	}
	\label{tab:qwen_small_sharpness}
	\begin{tabular}{l cc}
		\toprule
		\texttt{optimizer} & $\mathrm{ppl}_{\mathtt{test}}$ & $\lambda_1^{\mathrm{test}}$ \\
		\midrule
		Adam($\eta = 10^{-5}$) & $12.159$ & $15819.432$ \\
		NGD($\eta = 0.03$) & $12.323$ & $\mathbf{1693.407}$ \\
		SAM($\eta = 1.5 \times 10^{-5}$, $\rho = 0.01$) & $12.172$ & $3237.732$ \\
		CDAT($\sigma = 1$, $\beta = 0$) & $12.474$ & $1717.565$ \\
		CDP($\kappa = 1.5$) + Adam($\eta = 10^{-5}$) & $12.165$ & $2137.903$ \\
		\bottomrule
	\end{tabular}
\end{table}

\begin{table}[tbp!]
	\centering\footnotesize
	\setlength{\tabcolsep}{1.5pt}
	\captionsetup{skip = 8pt}
	\caption{
		Test perplexity before and after $4$-bit quantization of the final checkpoints compared in \Cref{tab:qwen_small_sharpness}, lower is better (values within $0.02$ of the minimum in each column are highlighted).
		GPTQ and AWQ average two calibration draws.
	}
	\label{tab:qwen_small_quantization}
	\begin{tabular}{l c cc cc cc c c c}
		\toprule
		\texttt{quantization} & / & \multicolumn{2}{c}{RTN} & \multicolumn{2}{c}{GPTQ} & \multicolumn{2}{c}{AWQ} & NF4 & SINQ & A-SINQ \\
		\cmidrule(lr){3-4}\cmidrule(lr){5-6}\cmidrule(lr){7-8}
		\texttt{group\_size} & / & $64$ & $128$ & $64$ & $128$ & $64$ & $128$ & / & $64$ & $64$ \\
		\midrule
		Adam($\eta = 10^{-5}$) & $\mathbf{12.159}$ & $16.222$ & $18.535$ & $13.837$ & $14.340$ & $13.980$ & $14.858$ & $14.172$ & $14.416$ & $14.360$ \\
		NGD($\eta = 0.03$) & $12.323$ & $15.199$ & $\mathbf{16.479}$ & $13.651$ & $14.094$ & $13.561$ & $14.141$ & $13.769$ & $13.734$ & $13.762$ \\
		SAM($\eta = 1.5 \times 10^{-5}$, $\rho = 0.01$) & $\mathbf{12.172}$ & $\mathbf{15.111}$ & $17.154$ & $13.614$ & $14.080$ & $\mathbf{13.458}$ & $\mathbf{14.020}$ & $13.636$ & $\mathbf{13.691}$ & $\mathbf{13.718}$ \\
		CDAT($\sigma = 1$, $\beta = 0$) & $12.474$ & $16.254$ & $17.912$ & $13.823$ & $14.241$ & $13.761$ & $14.358$ & $14.150$ & $14.032$ & $14.242$ \\
		CDP($\kappa = 1.5$) + Adam($\eta = 10^{-5}$) & $\mathbf{12.165}$ & $15.181$ & $16.567$ & $\mathbf{13.518}$ & $\mathbf{13.971}$ & $\mathbf{13.473}$ & $14.108$ & $\mathbf{13.603}$ & $\mathbf{13.705}$ & $\mathbf{13.699}$ \\
		\bottomrule
	\end{tabular}
\end{table}

We quantize the weights of the final checkpoints to $4$ bits with round-to-nearest (RTN), GPTQ \cite{gptq_frantar_2023}, AWQ \cite{awq_lin_2024}, NF4 with double quantization \cite{qlora_dettmers_2023}, and SINQ and A-SINQ \cite{sinq_muller_2026}.
GPTQ and AWQ use $128$ calibration sequences from \texttt{pile-10k}, of length $2048$ and $512$ tokens, respectively, while A-SINQ uses the authors' synthetic calibration prompts.
\begin{remark}[Quantization robustness]
	This comparison is exploratory and does not fully establish that reducing sharpness generally improves robustness to quantization.
	Further experiments across training seeds, models, and datasets would be needed to assess whether this benefit persists, but are beyond the scope of this work.
\end{remark}

\end{document}